\documentclass{article}

\PassOptionsToPackage{numbers, compress}{natbib}
 \usepackage[preprint]{neurips_2026}

\usepackage[utf8]{inputenc} 
\usepackage[T1]{fontenc}    
\usepackage{hyperref}       
\usepackage{url}            
\usepackage{booktabs}       
\usepackage{amsfonts}       
\usepackage{nicefrac}       
\usepackage{microtype}      
\usepackage{xcolor}         
\usepackage{graphicx}
\usepackage{subcaption}
\usepackage{hyperref}
\usepackage{caption}
\usepackage{hyphenat}
\usepackage{setspace}

\usepackage{amsmath}
\usepackage{amssymb}
\usepackage{mathtools}
\usepackage{amsthm}
\usepackage{longtable}
\usepackage{multirow}
\usepackage[table]{xcolor}
\usepackage{float}
\usepackage{tabularx}
\usepackage{tcolorbox}
\usepackage{enumitem}
\usepackage{threeparttable}
\usepackage{algorithm}
\usepackage{algorithmic}
\usepackage{titletoc}
\tcbuselibrary{skins, breakable, listings}
\usepackage[capitalize,noabbrev]{cleveref}

\definecolor{wblue}{RGB}{52,204,204}
\definecolor{dblue}{RGB}{54,160,205}
\definecolor{SPIRAL_1}{RGB}{57,95,207}
\definecolor{SPIRAL_2}{RGB}{56,117,206}
\definecolor{SPIRAL_3}{RGB}{55,139,206}
\definecolor{SPIRAL_4}{RGB}{54,160,205}
\definecolor{SPIRAL_5}{RGB}{53,182,205}
\definecolor{SPIRAL_6}{RGB}{52,204,204}
\definecolor{metablue}{HTML}{0064E0}
\definecolor{metafg}{HTML}{1C2B33}
\definecolor{metabg}{HTML}{F1F4F7}
\definecolor{spiralboxbg}{HTML}{F1F4F7}
\definecolor{spiralboxfg}{HTML}{1C2B33}

\definecolor{tabcolor}{rgb}{0.64, 0.87, 0.93}
\hypersetup{
  colorlinks=true,
  linkcolor=wblue,
  citecolor=dblue,
  urlcolor=metablue
}

\def\ourMethod{SPIRAL}
\def\ourMethodstr{{SPIRAL}}
\def\ourDataset{{ActVideoGen-Dataset}}
\def\ourBenchmark{{ActVideoGen-Bench}}

\newcommand{\ourMethodColored}{\textcolor{SPIRAL_1}{S}\textcolor{SPIRAL_2}{P}\textcolor{SPIRAL_3}{I}\textcolor{SPIRAL_4}{R}\textcolor{SPIRAL_5}{A}\textcolor{SPIRAL_6}{L}}

\newcommand{\hi}[1]{\textbf{\textcolor{wblue}{\scriptsize{#1}}}}
\newcommand{\nohi}[1]{\textbf{\textcolor{gray}{\scriptsize{#1}}}}

\DeclareFontFamily{T1}{optimistic}{}
\DeclareFontShape{T1}{optimistic}{m}{n}{<-> s * [0.88] assets/optimistic}{}
\DeclareFontShape{T1}{optimistic}{b}{n}{<-> s * [0.88] assets/optimistic}{}
\pdfmapline{+optimistic < assets/Optimistic.ttf <T1-WGL4.enc}

\newcommand{\spiraltitleblock}{%
  {\huge\sffamily\bfseries \ourMethodColored: Self-Evolving Action-Conditioned Video Generation via Reflective Planning Agents}%
}

\title{\spiraltitleblock}

\newcommand{\spiralauthorblock}{%
  {\sffamily\bfseries Yu Yang$^{*,1,2,3}$}, {\sffamily\bfseries Yue Liao$^{*,3}$}, {\sffamily\bfseries Jianbiao Mei$^{*,1,2}$}, {\sffamily\bfseries Baisen Wang$^{*,4}$}, {\sffamily\bfseries Xuemeng Yang$^{2}$}, {\sffamily\bfseries Licheng Wen$^{2}$}, {\sffamily\bfseries Jiangning Zhang$^{1,5}$}, {\sffamily\bfseries Xiangtai Li$^{6}$}, {\sffamily\bfseries Liang Lv$^{7}$}, {\sffamily\bfseries Hanlin Chen$^{3}$}, {\sffamily\bfseries Botian Shi$^{2}$}, {\sffamily\bfseries Yong Liu$^{1,\dag}$}, {\sffamily\bfseries Shuicheng Yan$^{3}$}, {\sffamily\bfseries Gim Hee Lee$^{3}$}%
}

\newcommand{\spiralaffiliationblock}{%
  {\normalsize $^1$Zhejiang University}, {\normalsize $^2$KnowledgeXLab at Shanghai AI Lab}, {\normalsize $^3$National University of Singapore}, {\normalsize $^4$Chinese Academy of Sciences}, {\normalsize $^5$Tencent Youtu Lab}, {\normalsize $^6$Nanyang Technological University}, {\normalsize $^7$Wuhan University}%
}

\newcommand{\spiralcontributionblock}{%
  {\small $^*$Equal Contribution}, {\small $^\dag$Corresponding Author}%
}

\newcommand{\spiralprojectpage}{\url{https://yuyang-cloud.github.io/spiral}}
\newcommand{\spiraldate}{May 21, 2026}
\newcommand{\spiralmetadatablock}{%
  {\small {\sffamily\bfseries \raisebox{-0.2em}{\includegraphics[width=0.025\linewidth]{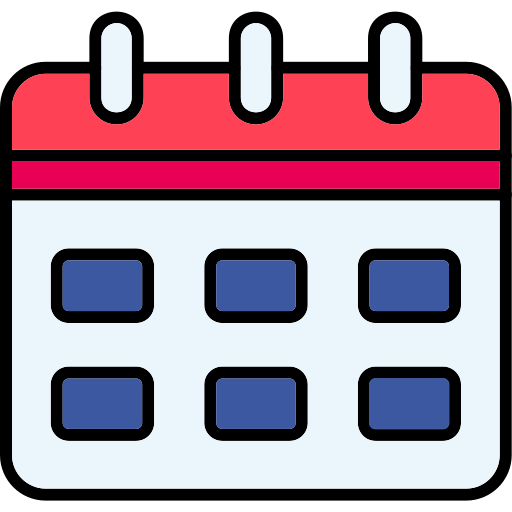}}~~Date:} \spiraldate}\par%
  \vskip 0.08cm%
  {\small {\sffamily\bfseries \raisebox{-0.2em}{\includegraphics[width=0.025\linewidth]{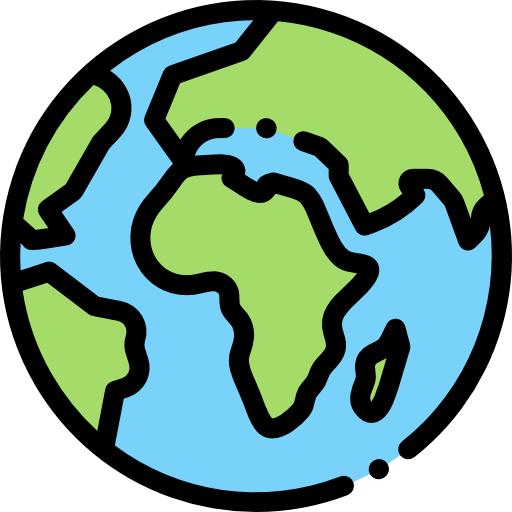}}~~Project Page:} \spiralprojectpage}%
}
\newcommand{\spiralfrontlogoh}{0.9cm}

\author{\spiralauthorblock}

\newcommand{\spiralabstract}{%
Long-horizon action-conditioned video generation aims to synthesize temporally coherent videos that follow complex action instructions over extended horizons, requiring procedural ordering, persistent action execution, and scene consistency beyond conventional TI2V's short-term fidelity.
Existing single-shot video generation models typically operate in an open-loop manner, leading to incomplete action execution, hallucinated motions, and temporal drift.
To address this, we propose \ourMethodColored, a closed-loop framework that performs \underline{\textit{S}}equential \underline{\textit{P}}lanning and \underline{\textit{I}}terative \underline{\textit{R}}eflection for \underline{\textit{A}}ction-conditioned \underline{\textit{L}}ong-horizon video generation.
Specifically, \ourMethod{} instantiates a think-act-reflect process: a PlanAgent decomposes high-level goals into sub-actions, which condition a VideoGenerator to synthesize each segment alongside a memory context, while a CriticAgent evaluates intermediate video segments to provide corrective feedback for iterative refinement.
This closed-loop design further supports self-evolution by utilizing PlanAgent-proposed actions and CriticAgent-derived rewards for GRPO-based post-training to enhance the video generator's long-horizon consistency.
Moreover, we introduce \ourDataset{} for task-specific training, and establish \ourBenchmark{} as a dedicated evaluation suite for measuring action quality and temporal coherence.
Experiments across multiple TI2V backbones alongside the self-evolving strategy show consistent gains on \ourBenchmark{} and VBench, demonstrating the effectiveness of \ourMethod{}.
}

\newcommand{\spiralfrontbox}{%
  \tcbset{enhanced,frame hidden}%
  \tcbset{left=0.5cm}%
  \tcbset{right=0.5cm}%
  \tcbset{top=0.5cm}%
  \tcbset{bottom=0.5cm}%
  \tcbset{arc=16pt}%
  \tcbset{before skip=0pt}%
  \tcbset{grow to left by=1.5pt}%
  \tcbset{grow to right by=1.5pt}%
  \tcbset{overlay={
    \node[
      anchor=north west,
      at= (frame.north west),
      xshift= 0.5cm,
      yshift= -0.5cm] {\includegraphics[height=\spiralfrontlogoh]{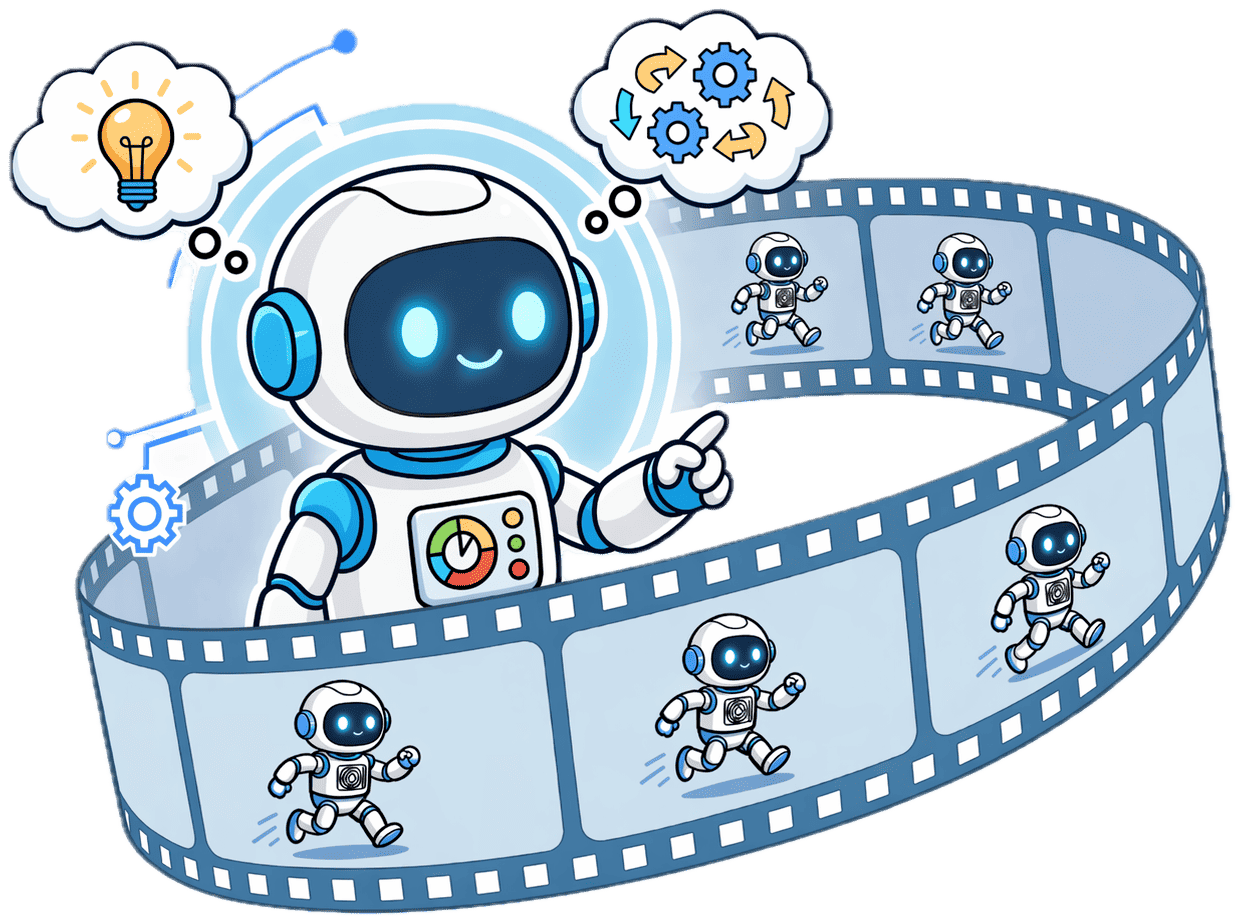}};
    \node[
      anchor=north east,
      at= (frame.north east),
      xshift= -0.5cm,
      yshift= -0.5cm] {%
        \includegraphics[height=\spiralfrontlogoh]{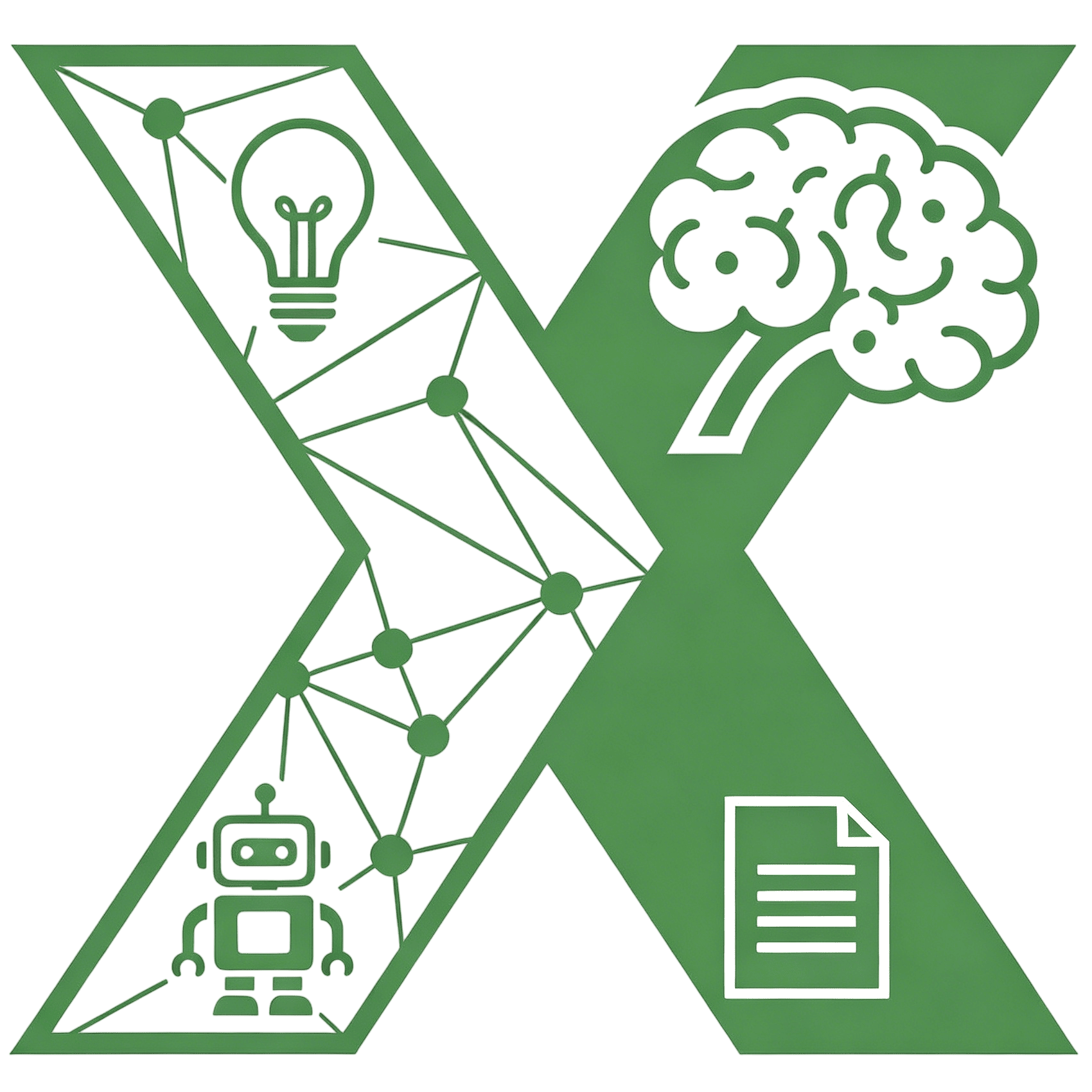}\hspace{0.3cm}%
        \includegraphics[height=\spiralfrontlogoh]{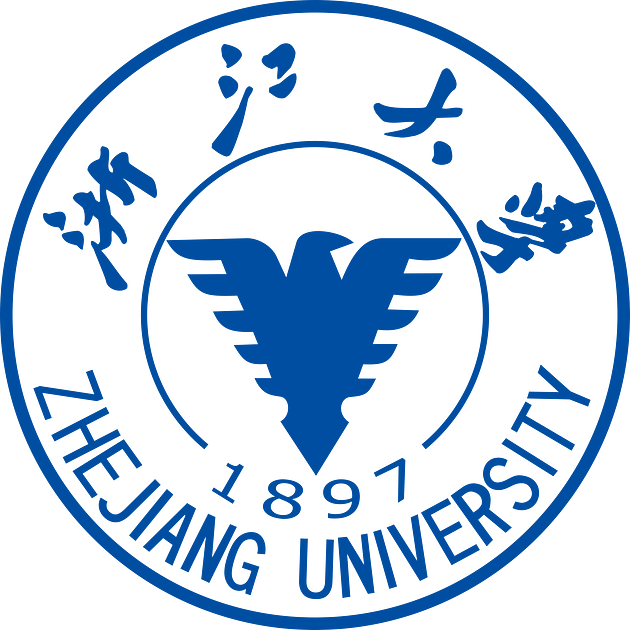}\hspace{0.3cm}%
        \includegraphics[height=\spiralfrontlogoh]{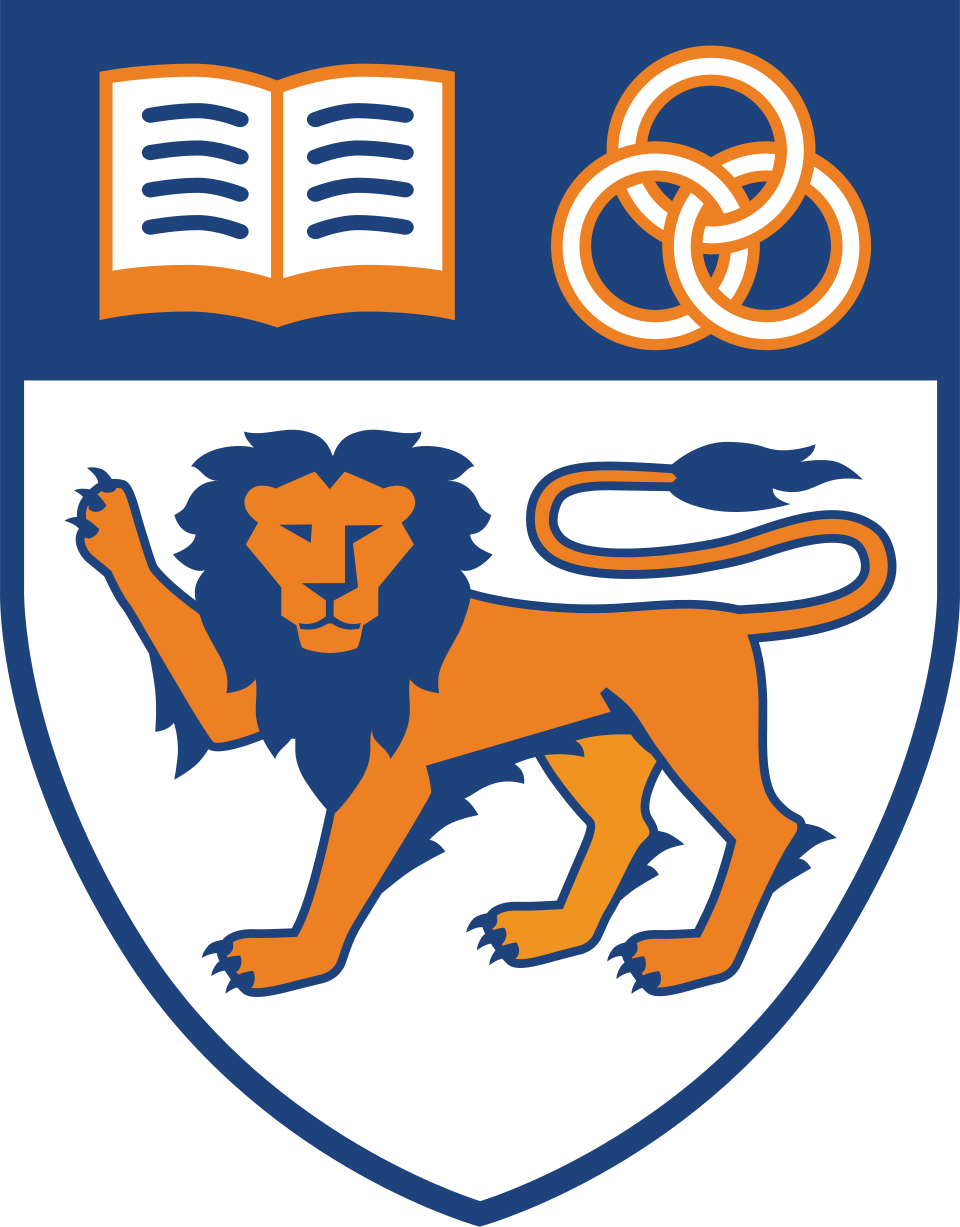}%
      };
  }}%
  \begin{tcolorbox}
    \setlength{\parindent}{0cm}
    \setlength{\parskip}{0.5cm}
    {
      \setlength{\parskip}{0cm}
      \raggedright
      \nohyphens
      {
        \vskip 1.25cm
        \setstretch{1.618}
        \spiraltitleblock\par
      }
      \vskip 0.25cm
      \spiralauthorblock\par
      \vskip 0.35cm
      \spiralaffiliationblock\par
      \vskip 0.08cm
      \spiralcontributionblock\par
    }
    {\color{metafg}\spiralabstract\par}
    \vskip 0.5cm
    {
      \setlength{\parskip}{0cm}
      \spiralmetadatablock\par
    }
  \end{tcolorbox}
  \tcbset{reset}
}

\begin{document}

\newgeometry{top=0.75in,bottom=0.75in,textwidth=6.3in}
\spiralfrontbox
\begin{center}
    \begin{minipage}{1.0\textwidth}
        \centering
        \includegraphics[width=\textwidth]{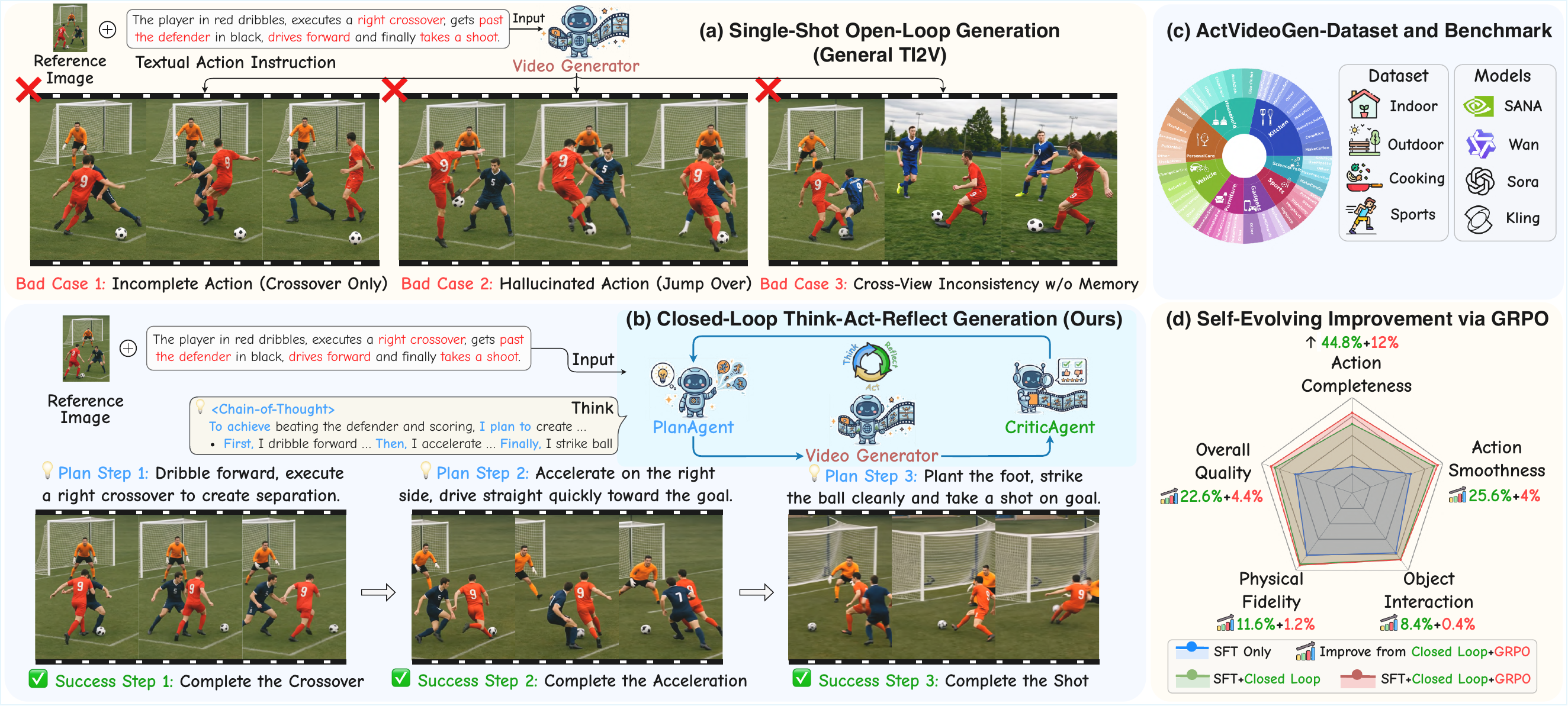}
        \vspace{-0.5cm}
        \captionof{figure}{\textbf{Long-Horizon Action-Conditioned Video Generation}: \emph{Challenges and Solutions.}
        (a) Genral TI2V follows a single-shot and open-loop paradigm, often causing incomplete action execution and hallucinated motions.
        (b) We propose a \emph{closed-loop think-act-reflect} framework for iterative planning, step-wise generation, and critical verification.
        (c) We introduce the ActVideoGen-Dataset and Benchmark for task-specific training and evaluation.
        (d) Our closed-loop design enables GRPO-based self-evolving, continually improving video generation quality.}
        \label{fig:teaser}
    \end{minipage}
\end{center}
\clearpage
\newgeometry{
  textheight=9in,
  textwidth=6.3in,
  top=1in,
  headheight=12pt,
  headsep=25pt,
  footskip=30pt
}
\section{Introduction}
Recent advances in video generation~\citep{Wan,Sora,chen2025sana,seedance2026seedance} have enabled increasingly realistic and coherent visual synthesis across diverse scenes and motions, paving the way for controllable motion simulation~\cite{agarwal2025cosmos,chi2025empowering,akkerman2025interdyn,liao2025genie,chen2025learning} and dynamic modeling~\cite{geng2025motion,li2025wonderplay,zhan2026perpetualwonder,liu2026realwonder}.
Beyond short-clip generation, a pivotal frontier is \emph{long-horizon action-conditioned video generation}. Unlike conventional text-and-image-to-video (TI2V) tasks that focus on globally plausible content and short-term motion, this setting demands sustained action execution, motion causality, and consistent object-tool interaction over extended temporal horizons, making it a formidable challenge.

Despite their success in short-term synthesis, current TI2V models struggle to sustain such complex long-term execution. As illustrated in Fig.~\ref{fig:teaser}(a), this paradigm shift exposes several critical challenges:
\emph{incomplete action execution}, where multi-stage actions are truncated or only partially completed; \emph{hallucinated motions}, where generated motions deviate from the intended instructions without planning and correction; and \emph{long-horizon temporal incoherence}, where the absence of memory or state tracking causes object drift and scene inconsistency.
These failures ultimately stem from the \textbf{\emph{single-shot open-loop}} nature of general TI2V generation, where the lack of task decomposition, intermediate verification, and corrective feedback allows errors to compound over time.

To address these challenges, we propose \ourMethodColored, an agentic framework for long-horizon action-conditioned video generation as shown in Fig.~\ref{fig:teaser}(b). Rather than relying solely on a single-shot generator, SPIRAL serves as a closed-loop harness over the base video generator, augmenting it with explicit planning, verification, memory, and feedback. Specifically, a \emph{PlanAgent} decomposes a high-level goal into step-wise sub-actions, a \emph{VideoGenerator} synthesizes each segment conditioned on the current sub-action and historical context, a \emph{CriticAgent} evaluates intermediate videos and provides corrective feedback, and a \emph{Long-Horizon Memory} module preserves visual and semantic context. Together, these components instantiate a \textbf{\emph{closed-loop think-act-reflect}} mechanism that improves action completeness through sequential planning, reduces hallucinations through iterative reflection, and mitigates long-horizon drift through memory-aware generation.

Furthermore, \ourMethod{} extends its utility beyond an inference-time harness to a \textbf{\emph{trainable self-evolving mechanism}}.
Its closed-loop design naturally yields structured action plans and critic-derived rewards, enabling GRPO-based post-training to enhance the video generator. Concretely, the PlanAgent provides progressively challenging action instructions, and the video generator synthesizes candidate video segments. The CriticAgent then evaluates the action-video pairs in terms of action fidelity and temporal consistency, providing critic-based rewards that directly optimize the generator for improvement. By integrating a \emph{curriculum learning} strategy that incrementally scales action complexity and temporal horizons, it drives the generator to seamlessly evolve from executing short atomic actions to mastering long procedural behaviors, leading to persistent performance gains.

Beyond the framework design, long-horizon action-conditioned video generation also requires task-specific supervision and evaluation protocols. We further introduce \textbf{\emph{\ourDataset}} (Fig.~\ref{fig:teaser}(c)), comprising 24,616 tasks and 118,156 step-level annotations. Each sample is organized around a goal-level description, a structured CoT rationale, step-wise actions, and aligned video segments, providing supervised adaptation for both the PlanAgent and the video generator. Furthermore, we present \textbf{\emph{\ourBenchmark}}, an evaluation suite tailored to this setting with hierarchical difficulty levels and multi-dimensional metrics, emphasizing action quality and long-term consistency.

Extensive experiments show that integrating \ourMethod{} with diverse TI2V backbones consistently improves long-horizon action quality and temporal consistency. As depicted in Fig.~\ref{fig:teaser}(d), compared to the open-loop baseline~\cite{LongLive}, \ourMethod{}'s closed-loop framework improves overall action quality by 22.6\%, while GRPO-based evolving yields an additional 4.4\% gain. These results show that \ourMethod{} is not only an effective inference-time harness, but also a practical self-improving framework for long-horizon action-conditioned video generation.

Our key contributions are summarized as follows:
\begin{itemize}[leftmargin=*, itemsep=0pt, topsep=0pt]
    \item We propose \ourMethod{}, a closed-loop, agentic framework for long-horizon action-conditioned video generation that augments a video generator with explicit planning, reflective feedback, and context memory.
    \item We introduce a GRPO-based evolving strategy, leveraging planning and verification signals to continuously optimize the video generator for self-improvement beyond inference-time correction.
    \item We present \ourDataset{} and \ourBenchmark{}, a task-specific dataset and benchmark designed for training and rigorously evaluating long-horizon action-conditioned video generation.
\end{itemize}
\section{Related Work}
Recent visual and video generation research is evolving from single-shot synthesis into agentic, closed-loop systems driven by chain-of-thought reasoning and iterative planning~\cite{wu2026visual, guo2025can, long20262}. Existing paradigms leverage structured workflows~\cite{he2026gems, feng2026gen}, tool execution~\cite{lin2025jarvisart}, and ``thinking while generating'' approaches~\cite{guo2025thinking, soni2025videoagent, zeng2025coagent} to enhance controllability, grounded verification, and self-correction. While these works address complex visual trajectories, our approach explicitly integrates reflective planning agents with an explicit memory module to ensure sustained action grounding over long temporal horizons. Additional related works are provided in the Appendix~\ref{app:related_work}.
\vspace{-2mm}
\section{Method}
\label{sec:method}

\subsection{Overview}
\noindent \textbf{Problem Formulation.}
\label{sec:problem_formulation}
We formulate long-horizon, action-conditioned video generation as a Markov Decision Process (MDP) defined by the tuple $\mathcal{T} = \langle g, \mathcal{S}, \mathcal{V} \rangle$.
Given a global goal $g$, the system determines structured action steps $\mathcal{S} = \{s_1, \dots, s_T\}$ that guide the synthesis of a coherent video trajectory $\mathcal{V} = \{v_1, \dots, v_T\}$ to accomplish the specified goal.
At each timestep $t$, the framework jointly generates the plan $s_t$ and the video segment $v_t$ conditioned on the historical context, maximizing the cumulative rewards reflecting action completion and goal achievement.

\begin{figure*}[t]
    \centering
    \includegraphics[width=0.99\linewidth]{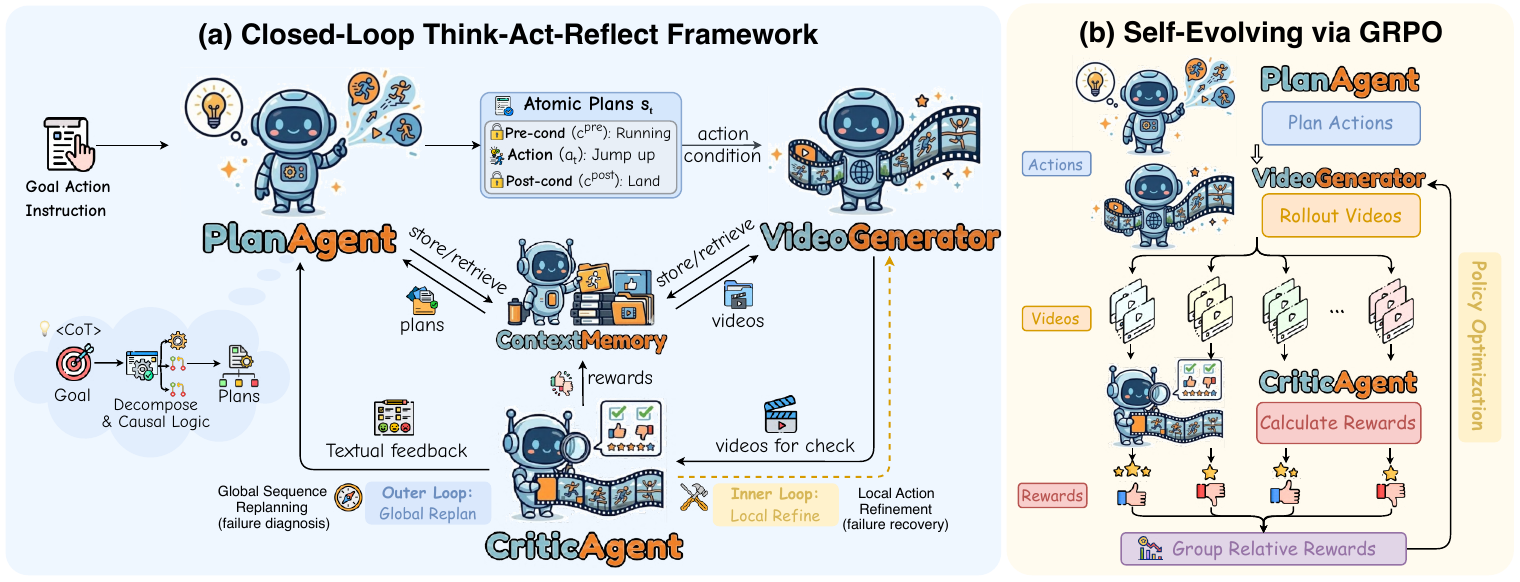}
    \caption{\textbf{\ourMethodColored{} Overview.} (a) \textbf{Closed-Loop Framework}: The PlanAgent decomposes abstract goals into atomic plans for action-conditioned video generation, while the CriticAgent evaluates action-video alignments and triggers dual-level (inner/outer) feedback for refinement (Fig.~\ref{fig:framework}). (b) \textbf{Self-Evolving via GRPO}: Guided by the PlanAgent, the VideoGenerator produces video rollouts, using CriticAgent rewards for policy optimization (Fig.~\ref{fig:grpo}).}
    \label{fig:pipeline}
    \vspace{-3mm}
\end{figure*}

\noindent \textbf{Overall Framework.}
\label{sec:method_overview}
To ensure long-term consistency and controllability, we propose a hierarchical, closed-loop framework in Fig.~\ref{fig:pipeline}(a) that operates iteratively:
\textbf{(1)} a \textbf{PlanAgent} ($\pi_{\text{plan}}$), serving as the high-level policy, employs CoT reasoning to decompose global goal $g$ into structured atomic plans $s_t$;
\textbf{(2)} a \textbf{VideoGenerator} ($\pi_{\text{gen}}$) functions as the execution policy, synthesizing video segments $v_t$ conditioned on the plan $s_t$ and context memory $\mathcal{M}_{t-1}$;
\textbf{(3)} a \textbf{CriticAgent} ($\mathcal{C}$) evaluates the action-video alignment via reward $r_t = \mathcal{C}(v_t, s_t)$, triggering refinement feedback upon detecting failures; and
\textbf{(4)} \textbf{ContextMemory} accumulates successful action-video transitions as $\mathcal{M}_t = \mathcal{M}_{t-1} \cup \{(s_t, v_t)\}$ to preserve global coherence.
Collectively, this cycle instantiates a \textbf{\emph{think-act-reflect}} loop (detailed in Fig.~\ref{fig:framework}), advancing beyond single-shot generation toward a robust plan-generate-verify formulation.

\subsection{PlanAgent: Structured Reasoning and Planning}
\label{sec:plan_agent}
Directly conditioning a video generator on an abstract goal $g$ in a single-shot manner often leads to incomplete execution or hallucinations.
To mitigate this, we employ a \textbf{PlanAgent} (parameterized by a VLM $\pi_{\text{plan}}$) to explicitly decompose high-level goals into executable action sequences.

Formally, given a goal $g$ and context memory $\mathcal{M}$, the agent employs Chain-of-Thought (CoT) reasoning to decompose the instruction, establish causal logic, and synthesize structured plans:
\begin{equation}
    \mathcal{S} = \{s_1, \dots, s_T\},
    \quad \mathcal{S} \sim \pi_{\text{plan}}(\cdot \mid g, \mathcal{M})
\end{equation}
Each atomic plan is defined as a tuple $s_t$ = $(a_t, c_t^{pre}, c_t^{post})$, where $a_t$ denotes the language action instruction, while $c_t^{pre}, c_t^{post}$ represent physical pre- and post-conditions, respectively.
For instance, for an action $a_t$ = $\text{`jump over a hurdle'}$, the agent specifies $c_t^{pre}$ = $\text{`running'}$ and $c_t^{post} = \text{`landed'}$.
Crucially, this CoT paradigm compels the agent to explicitly reason about causal dependencies, ensuring that planned actions are physically feasible, not merely semantically relevant.

We instantiate PlanAgent with Qwen3-VL-8B~\cite{Qwen3-VL} and execute a hierarchical training to enhance its causal planning capacity: we first conduct \emph{Instruction Tuning (IT)} on our curated \ourDataset{} (Sec.~\ref{sec:dataset}) to instill structured CoT-based planning capabilities, followed by \emph{Direct Preference Optimization (DPO)} to enforce alignment with physical reality and temporal logic.
This regimen empowers the agent to not only capture causal dependencies but also mitigate logical hallucinations in complex tasks.
Implementation details are provided in Appendix~\ref{app:plan_agent}.

\subsection{VideoGenerator: Action-Conditioned Video Generation}
\label{sec:video_generator}
The \textbf{VideoGenerator} (instantiated as a video diffusion model $\pi_{\text{gen}}$) serves as the execution policy, translating each atomic plan $s_t$ into a video segment.
For long-term consistency, synthesizing the current video segment $v_t$ is conditioned on both action instruction $a_t$ and historical memory $\mathcal{M}_{t-1}$:
\begin{equation}
    v_t \sim \pi_{\text{gen}}(\cdot \mid a_t, \text{Encoder}(\mathcal{M}_{t-1}))
\end{equation}
Specifically, instruction $a_t$ specifies the intended motion or interaction, while historical memory $\mathcal{M}_{t-1}$, stored as visual keyframes or latent KV-caches, provides contextual guidance.
The generation process is executed in a streaming manner, with each video segment corresponding to a single action step, and context memory preserving motion continuity over extended horizons.

Crucially, our framework supports \textbf{plug-and-play} integration of arbitrary video generators.
To endow standard short-clip generators with long-horizon synthesis capabilities, we adopt a \emph{Streaming Long-Tuning} strategy by performing supervised fine-tuning (SFT) on our \ourDataset{}, empowering diverse T2V and I2V backbones to faithfully execute step-wise action instructions over extended durations.
Appendix~\ref{app:video_generator} details adapting SVI~\cite{SVI} for long-horizon action-conditioned video generation.

\subsection{CriticAgent: Reward and Closed-Loop Feedback}
\label{sec:critic_agent}
Despite structured planning, open-loop generation remains susceptible to execution failures, such as missed actions, post-condition violations, and physical inconsistencies.
To address this, we employ a \textbf{CriticAgent} (parameterized by a VLM $\mathcal{C}$) to enforce closed-loop verification and feedback.

Specifically, the atomic plan $s_t$ generated by PlanAgent contains action instructions and physical pre- and post-conditions, which serve as critic anchors for CriticAgent to verify whether the generated video executes the intended action and satisfies the required conditions. Consequently, for each video segment $v_t$, CriticAgent assesses its alignment with plan $s_t$ across five dimensions: \textit{action adherence, object interaction, goal achievement, temporal coherence, and physical realism}, producing a scalar reward $r_t \in [0,1]$ to quantify generation quality, alongside textual feedback $f_t$ for diagnosis:
\begin{equation}
    r_t, f_t = \mathcal{C}(v_t, s_t)
\label{eq:critic}
\end{equation}
Leveraging these signals, we implement a dual-level feedback mechanism:
\textbf{(1) Inner Loop (Local Refinement):} If $r_t$ falls below a predefined threshold $\tau$, indicating minor artifacts or incomplete action execution, the feedback $f_t$ is used to refine the instruction $a_t$ (e.g., emphasizing unmet post-conditions) for immediate regeneration;
\textbf{(2) Outer Loop (Global Replanning):} If generation fails consecutively for $K$ attempts, implying an infeasible plan (e.g., missing conditions or incorrect action ordering), the failure propagates to the PlanAgent to trigger the replanning of trajectory $\mathcal{S}$ from step $t$.

To ensure robust verification, the CriticAgent undergoes a two-stage training regimen: first, we distill judgments from large VLMs (e.g., Gemini-3-Pro) via \textit{Supervised Fine-Tuning (SFT)}; subsequently, we refine the model via \textit{Pairwise Reward Modeling (RM)} using a Bradley-Terry objective to enhance discriminative accuracy.
Detailed implementation protocols are provided in Appendix~\ref{app:critic_agent}.

\subsection{Self-Evolving: Closed-Loop GRPO Optimization}
\label{sec:grpo_training}
While inference-time feedback mitigates immediate errors, we aim to internalize such corrections to permanently enhance VideoGenerator.
To this end, we propose a \textbf{Self-Evolving} strategy based on GRPO, enabling closed-loop reinforcement learning driven by PlanAgent-proposed actions and CriticAgent-derived rewards to continuously optimize the VideoGenerator's performance.

Formally, initialized from supervised-finetuned $\pi_{\text{gen}}$, VideoGenerator policy $\pi_{\theta}$ undergoes iterative refinement leveraging stochastic actions from $\pi_{\text{plan}}$ and rewards from $\mathcal{C}$ (Fig.~\ref{fig:grpo}).
For action step $s_t$, we sample $G$ video trajectories $\{v_{t,i}\}_{i=1}^G$ via previous policy $\pi_{\theta_{\mathrm{old}}}$. CriticAgent evaluates each sample, assigning reward set $\{r_i\}_{i=1}^G$.
We then compute advantage $A_{i}$ using group-wise normalization:
\begin{equation}
\vspace{-2mm}
\small
    A_{i} = \frac{r_{i} - \text{mean}(\{r_1, \dots, r_G\})}{\text{std}(\{r_1, \dots, r_G\}) + \delta}
\label{eq:advantage}
\end{equation}
\vspace{-2mm}

The VideoGenerator parameters $\theta$ are updated by maximizing the GRPO objective:

\vspace{-4mm}\begin{equation}
\small
\begin{aligned}
    \mathcal{J}(\theta) =
    \mathbb{E}_{\substack{
    s_t \sim \pi_{\text{plan}}(\cdot \mid g) \\
    v_{t,i} \sim \pi_{\theta_{\mathrm{old}}}(\cdot \mid s_t)
    }}
    \Bigg[
    \frac{1}{G} \sum_{i=1}^{G}
    \min \Big(
    \rho_{t,i} A_i, \mathrm{clip}(\rho_{t,i}, 1-\epsilon, 1+\epsilon)\, A_i
    \Big)
    - \beta\, D_{\mathrm{KL}}\!\left(\pi_{\theta} \,\|\, \pi_{\mathrm{gen}}\right)
    \Bigg]
\end{aligned}
\label{eq:grpo_objective}
\end{equation}
where $\rho_{t,i} =
\frac{\pi_{\theta}(v_{t,i} \mid s_t)}
     {\pi_{\theta_{\mathrm{old}}}(v_{t,i} \mid s_t)}$
is the importance sampling ratio between $\pi_{\theta}$ and $\pi_{\theta_{\mathrm{old}}}$ for sample $v_{t,i}$,
$D_{\mathrm{KL}}(\cdot)$ denotes KL divergence, and $\epsilon, \beta$ are hyperparameters for policy clipping and regularization strength, respectively.

This closed-loop optimization effectively distills PlanAgent's reasoning and CriticAgent's verification into the generator's execution policy.
Moreover, this framework naturally facilitates \emph{Curriculum Learning}. By progressively scaling action complexity and temporal horizons, the VideoGenerator evolves from generating atomic actions to executing complex, long-horizon procedural tasks, yielding sustained improvements over extended horizons. Additional implementation details are provided in Appendix~\ref{app:grpo}.
\vspace{-1mm}
\section{Dataset and Benchmark}
\begin{figure*}[t]
    \centering
    \includegraphics[width=0.98\linewidth]{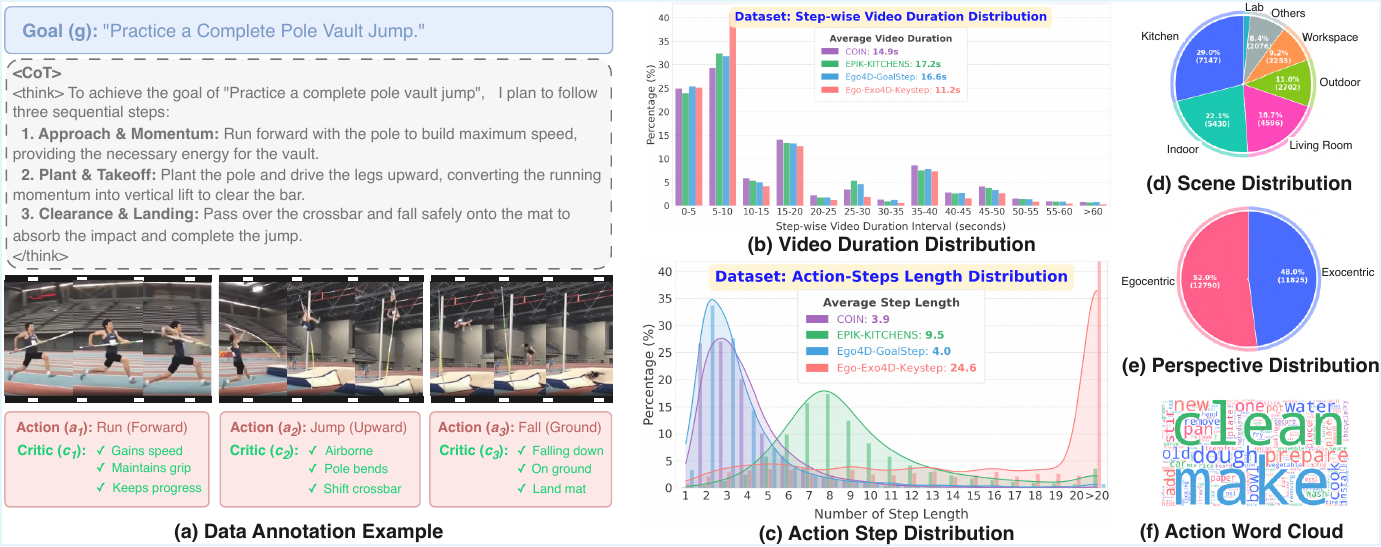}
    \caption{\textbf{Overview of ActVideoGen-Dataset.} (a) A structured data example comprising goal, CoT, and step-wise video-action-critic tuples. (b-f) Detailed dataset distributions across diversity, temporal scale, and action complexity.}
    \label{fig:dataset}
    \vspace{-1em}
\end{figure*}

\subsection{ActVideoGen-Dataset Construction}
\label{sec:dataset}
To support task-specific training, we introduce \textbf{\ourDataset{}}, providing the structured $\langle g, \text{CoT}, \mathcal{S}, \mathcal{V} \rangle$ supervision. It integrates global reasoning with step-wise actions grounded in video segments, providing a comprehensive foundation for long-horizon planning and video generation.

\noindent\textbf{Data Initialization.} To ensure diverse real-world task coverage, we aggregate videos from $4$ datasets: Ego4D~\cite{ego4d, ego4d-goal-step}, Ego-Exo4D~\cite{ego-exo4d}, COIN~\cite{coin}, and EPIC-KITCHENS~\cite{epickitchens}. These sources encompass a broad range of procedural activities across egocentric and exocentric viewpoints. To eliminate redundant or irrelevant content, we use an automated shot-detection pipeline to filter out low-quality clips, retaining only segments that are strictly aligned with significant action boundaries for annotation.

\noindent\textbf{Hierarchical Annotation Pipeline.}
We develop an automated pipeline to convert raw video clips into structured annotations $\langle g, \text{CoT}, \mathcal{S}, \mathcal{V} \rangle$, effectively bridging visual signals with explicit step-wise actions. The process consists of two strategic stages.
\textbf{Stage 1: Step-wise Action Annotation.}
This stage instantiates structured step tuple $s_t=(a_t, c_t^{pre}, c_t^{post})$ from video segments. Taking the video clip $v_t$ as input, we leverage a VLM (e.g., GLM-4.5V~\cite{glm-4.5}) to identify fine-grained action instruction $a_t$ (specifying verb-object-tool interactions) and infer physical state transitions $(c_t^{pre}, c_t^{post})$. This yields an action sequence $\mathcal{S} = \{s_1, \dots, s_T\}$ explicitly modeling physical dependencies.
\textbf{Stage 2: CoT Planning Generation.}
We employ an LLM (e.g., GPT-5.1) to orchestrate global reasoning logic based on goal $g$ and the action sequence $\mathcal{S}$. By synthesizing a reasoning narrative $\text{CoT}$ (e.g., {\small \texttt{<think> \ldots{} </think>}}), the model transforms disjoint steps into a coherent, logically grounded plan, providing intermediate supervision for long-horizon CoT-based planning.

\noindent\textbf{Quality Verification and Human Alignment.}
To ensure dataset integrity, we implement a hierarchical pipeline integrating \textbf{VLM-based filtering} with \textbf{human verification}. Specifically, Qwen3-VL-235B~\cite{Qwen3-VL} serves as the verifier, filtering out samples with suboptimal video-text alignment. Subsequent human verification on a random subset yields a 93\% agreement rate between annotations and corresponding videos, confirming the annotations faithfully capture action semantics and causal dependencies. This rigorous validation ensures a reliable foundation for training planning agents and video generators. Comprehensive verification details are provided in Appendix~\ref{app:dataset}.

\noindent\textbf{Dataset Statistics and Analysis.}
\ourDataset{} contains 24,616 tasks and 118,156 step-level action-video pairs. As shown in Fig.~\ref{fig:dataset}, the dataset is characterized by its diversity in scenes (indoor/outdoor), perspectives (ego/exocentric), and complexity (varying durations/steps). This scale and structure make \ourDataset{} ideal for training planning-oriented agents and action-conditioned video generation.

\subsection{ActVideoGen-Bench Construction}
\label{sec:benchmark}
To systematically evaluate long-horizon generation capacity of diverse T2V and I2V backbones and our closed-loop SPIRAL framework, we introduce \textbf{\ourBenchmark{}}, a comprehensive evaluation suite featuring hierarchical difficulty levels and multi-dimensional metrics.

\noindent\textbf{Eval Prompt Suite.}
We curate 300 evaluation prompts stratified into three difficulty levels to probe long-horizon generalization: \textit{Simple} (1-3 action steps, $<20$s video duration), \textit{Medium} (3-5 steps, 20-40s), and \textit{Hard} ($>5$ steps, $>40$s).
To ensure robust coverage, the suite spans diverse \textit{scenes} (indoor, outdoor), \textit{perspectives} (egocentric, exocentric), and \textit{procedural activities} (e.g., sports, cooking).

\noindent\textbf{Evaluation Metrics.}
We adopt a hybrid evaluation protocol, reporting \emph{Static Quality} and \emph{Dynamic Quality} via VBench~\cite{VBench} to assess general visual fidelity.
However, given the focus on procedural action-conditioned video generation, such generic metrics are insufficient to determine whether intended actions are correctly executed.
Consequently, the primary evaluation target is \emph{Action Quality}.
To this end, we introduce a multi-agent action evaluator utilizing GPT-5 as the reasoning engine, performing comprehensive assessments via multi-agent collaboration and cross-verified chain-of-query reasoning.
Specifically, this system assesses Action Completeness, Action Smoothness, Object Interaction, and Physical Fidelity.
Comprehensive details are provided in Appendix~\ref{app:benchmark}.

\section{Experiments}
\label{sec:exp}

\noindent\textbf{Implementation Details.}
\textbf{(1) PlanAgent} is built upon Qwen3-VL-8B and trained on \ourDataset{} with LoRA in two stages: \textit{Instruction Tuning (IT)} to induce CoT reasoning and structured plan formatting, followed by \textit{Direct Preference Optimization (DPO)} aligning generated plans with logical dependencies.
\textbf{(2) CriticAgent} undergoes two-stage training: initially performing \textit{Supervised Fine-Tuning (SFT)} by distilling judgments of a strong model (Gemini-3-Pro) on videos from diverse baselines (e.g., Sora, CogVideoX~\cite{yang2024cogvideox}, SkyReels~\cite{SkyReels}, VideoVerse~\cite{VideoVerse}), and subsequently conducting \textit{Reward Modeling (RM)} using the Bradley-Terry objective on the GAIA dataset~\cite{GAIA} for action quality preference learning.
\textbf{(3) VideoGenerator} is instantiated by diverse T2V and I2V backbones, undergoing \textit{Streaming Long-Tuning} on \ourDataset{} to instill action-following capabilities.

\begin{figure}[!t]
    \centering

    \begin{minipage}[c]{0.47\linewidth}
        \centering
        \renewcommand{\arraystretch}{1.0}
        \setlength{\tabcolsep}{3pt}
        \small
        \captionof{table}{\textbf{PlanAgent Performance on EgoPlan-Bench.} Success rates (\%) across In-Domain, Out-of-Domain, and All splits for long-horizon action planning. Instr.: Instruction Tuning; Pref.: Preference Alignment.}
        \label{tab:planagent_bench}
        \vspace{-0.5em}
        \resizebox{\linewidth}{!}{
            \begin{tabular}{l|c|cc|ccc}
            \toprule
            \multirow{2}{*}[-0.8em]{\textbf{Method}} & \multirow{2}{*}[-0.8em]{\textbf{Base Model}} & \multicolumn{2}{c|}{\textbf{Training Stage}} & \multicolumn{3}{c}{\textbf{EgoPlan-Bench}} \\
            \cmidrule(lr){3-7}
             & & \textbf{\begin{tabular}[c]{@{}c@{}}Instr.\\ Tuning\end{tabular}} & \textbf{\begin{tabular}[c]{@{}c@{}}Pref.\\ Align.\end{tabular}} & \textbf{\begin{tabular}[c]{@{}c@{}}In-\\ Domain\end{tabular}} & \textbf{\begin{tabular}[c]{@{}c@{}}Out-of-\\ Domain\end{tabular}} & \cellcolor{gray!10}\textbf{All} \\
            \midrule

            GPT-4V & - & - & - & 38.40 & 36.90 & \cellcolor{gray!10}37.98 \\
            GPT-5.1 & - & - & - & 55.08 & 54.37 & \cellcolor{gray!10}54.78 \\
            SEED-LLaMA~\cite{ge2023making} & LLaMA2-Chat-13B & - & - & - & - & \cellcolor{gray!10}29.93 \\
            DeepVideo-R1~\cite{park2025deepvideo} & Qwen2.5-VL-7B & - & - & 52.00 & 55.70 & \cellcolor{gray!10}- \\
            SEED-Bench-RL~\cite{chen2025exploring} & Qwen2-VL-7B & - & - & 46.01 & 50.16 & \cellcolor{gray!10}- \\
            GRPO-CARE~\cite{chen2025grpo} & Qwen2.5-VL-7B & - & - & 57.00 & \textbf{57.00} & \cellcolor{gray!10}57.00 \\
            \midrule

            Video-LLaMA~\cite{zhang2023videollama} & \multirow{3}{*}{LLaMA2-Chat-7B} &  &  & 27.88 & 30.44 & \cellcolor{gray!10}28.58 \\
            Video-LLaMA~\cite{zhang2023videollama} &  & \checkmark &  & 52.14 & 40.52 & \cellcolor{gray!10}48.94 \\
            Video-LLaMA~\cite{zhang2023videollama} &  & \checkmark & \checkmark & 54.65 & 44.42 & \cellcolor{gray!10}51.83 \\
            \midrule

            \rowcolor{tabcolor!25} PlanAgent (Ours) &  &  & & 36.60 & 35.72 & \cellcolor{gray!10}35.81 \\
            \rowcolor{tabcolor!25} PlanAgent + Mem. &  &  &  & 44.68 & 43.31 & \cellcolor{gray!10}43.63 \\
            \rowcolor{tabcolor!25} PlanAgent + Mem. &  & \checkmark &  & 56.49 & 50.17 & \cellcolor{gray!10}53.29 \\
            \rowcolor{tabcolor!25} PlanAgent + Mem. & \multirow{-4}{*}{Qwen3-VL-8B} & \checkmark & \checkmark & \textbf{62.46} & 54.30 & \cellcolor{gray!10}\textbf{58.72} \\
            \bottomrule
            \end{tabular}
        }
    \end{minipage}\hfill
    \begin{minipage}[c]{0.51\linewidth}
        \centering
        \renewcommand{\arraystretch}{1.0}
        \setlength{\tabcolsep}{3pt}
        \small
        \captionof{table}{\textbf{CriticAgent Performance on VideoGen-Reward Bench.} Preference accuracy (\%) across Visual Quality (VQ), Motion Quality (MQ), Text Alignment (TA), and Overall Quality, evaluating multi-dimensional judgment.}
        \label{tab:criticagent_bench}
        \vspace{-0.5em}
        \resizebox{\linewidth}{!}{
            \begin{tabular}{l|cc|cccccccc}
            \toprule
            \multirow{3}{*}[-0.6em]{\textbf{Method}} & \multicolumn{2}{c|}{\textbf{\begin{tabular}[c]{@{}c@{}}Training\\ Stage\end{tabular}}} & \multicolumn{8}{c}{\textbf{VideoGen-RewardBench}} \\
            \cmidrule(lr){2-3} \cmidrule(lr){4-11}

             & \multirow{2}{*}[-0.4em]{\textbf{SFT}} & \multirow{2}{*}[-0.4em]{\textbf{RM}} &
             \multicolumn{2}{c}{\textbf{VQ Acc.}} &
             \multicolumn{2}{c}{\textbf{MQ Acc.}} &
             \multicolumn{2}{c}{\textbf{TA Acc.}} &
             \multicolumn{2}{c}{\cellcolor{gray!10}\textbf{Overall Acc.}} \\

             \cmidrule(lr){4-5} \cmidrule(lr){6-7} \cmidrule(lr){8-9} \cmidrule(lr){10-11}
             & & &
             w/ Ties & w/o &
             w/ Ties & w/o &
             w/ Ties & w/o &
             \cellcolor{gray!10}w/ Ties & \cellcolor{gray!10}w/o \\
            \midrule

            VideoPhy2~\cite{bansal2025videophy} & \checkmark & & - & - & - & - & 37.04 & 22.14 & \cellcolor{gray!10}30.75 & \cellcolor{gray!10}26.41 \\
            AIGVE~\cite{liu2025aigve} & \checkmark & & 38.05 & 30.80 & - & - & 30.76 & 11.66 & \cellcolor{gray!10}37.09 & \cellcolor{gray!10}37.08 \\
            LiFT-Critic~\cite{wang2024lift} & \checkmark & & 47.53 & 55.97 & 59.04 & 54.91 & 33.79 & 55.43 & \cellcolor{gray!10}39.08 & \cellcolor{gray!10}57.26 \\
            VideoScore~\cite{he2024videoscore} & \checkmark & & 47.41 & 47.72 & 59.05 & 51.09 & 37.24 & 50.34 & \cellcolor{gray!10}41.80 & \cellcolor{gray!10}50.22 \\
            Q-Align~\cite{wu2023qalign} & \checkmark & & 32.01 & 52.98 & - & - & 35.77 & 51.06 & \cellcolor{gray!10}42.05 & \cellcolor{gray!10}52.52 \\
            UnifiedReward~\cite{wang2025unifiedreward} & \checkmark & \checkmark & 41.27 & 39.42 & - & - & 40.11 & 36.58 & \cellcolor{gray!10}53.31 & \cellcolor{gray!10}58.83 \\
            Dover~\cite{wu2023dover} & \checkmark & & 39.34 & \textbf{68.87} & - & - & 38.01 & 55.65 & \cellcolor{gray!10}54.27 & \cellcolor{gray!10}68.58 \\
            VideoScore2~\cite{he2025videoscore2} & & \checkmark & 34.67 & 65.87 & - & - & 48.70 & 65.92 & \cellcolor{gray!10}54.53 & \cellcolor{gray!10}65.59 \\
            Q-Save~\cite{wu2025qsave} & \checkmark & \checkmark & 40.34 & 67.49 & - & - & \textbf{50.52} & 63.99 & \cellcolor{gray!10}56.63 & \cellcolor{gray!10}65.13 \\
            VisionReward~\cite{VisionReward} & \checkmark & \checkmark & 47.43 & 59.03 & 59.03 & 60.98 & 46.56 & 61.15 & \cellcolor{gray!10}56.77 & \cellcolor{gray!10}67.59 \\
            \midrule

            \rowcolor{tabcolor!25} CriticAgent (Ours) & \checkmark & & 47.16 & 59.27 & 59.08 & 61.20 & 44.71 & 60.39 & \cellcolor{gray!10}53.20 & \cellcolor{gray!10}63.16 \\
            \rowcolor{tabcolor!25} CriticAgent (Ours) & \checkmark & \checkmark & \textbf{49.79} & 63.94 & \textbf{59.98} & \textbf{62.42} & 47.98 & \textbf{66.52} & \cellcolor{gray!10}\textbf{57.31} & \cellcolor{gray!10}\textbf{68.86} \\
            \bottomrule
            \end{tabular}
        }
    \end{minipage}

    \vspace{0.6em}

    \begin{minipage}[c]{0.48\linewidth}
        \centering
        \includegraphics[width=\linewidth]{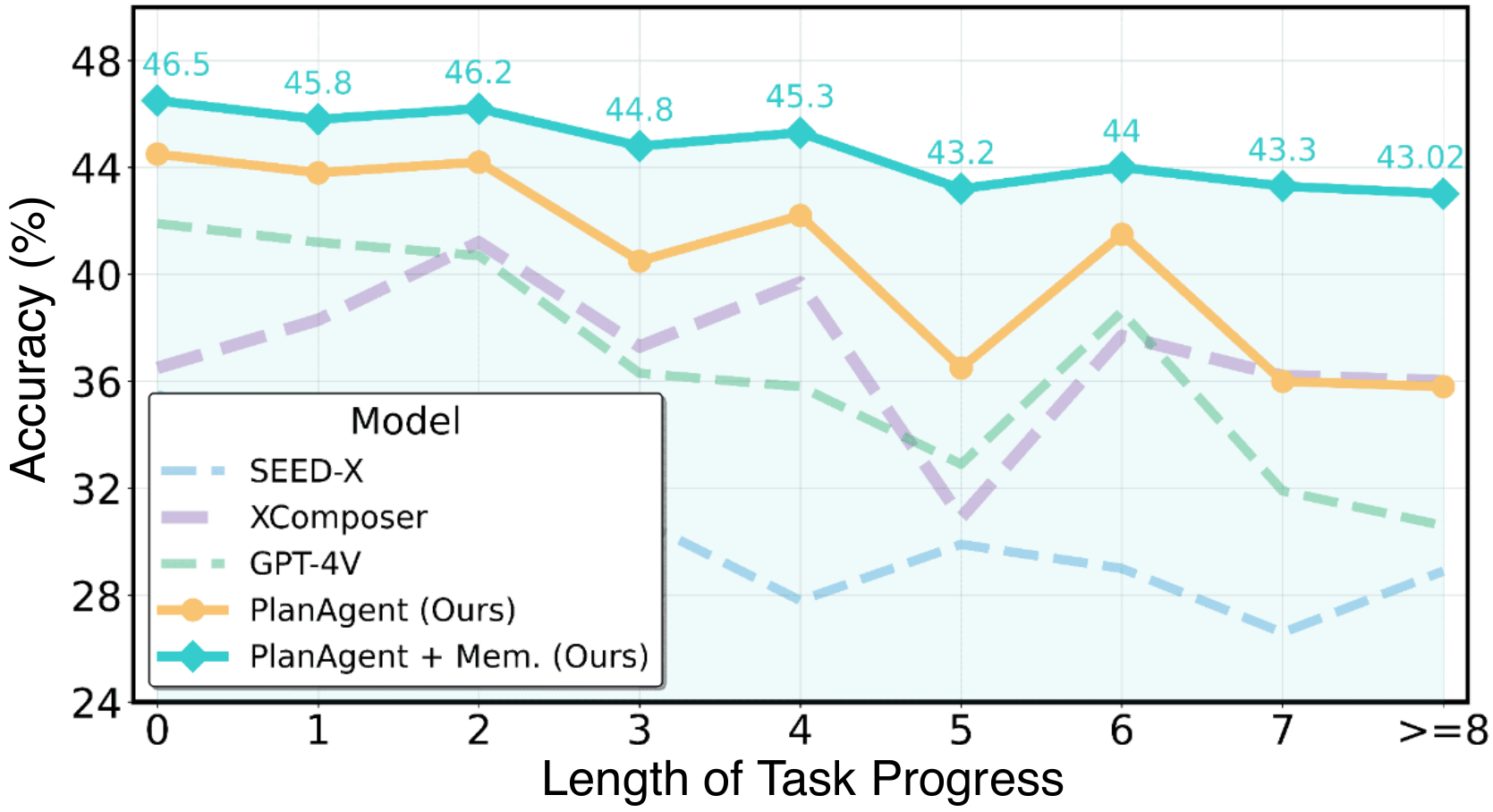}
        \vspace{-1.5em}
        \caption{\textbf{PlanAgent Robustness to Task Length.} Planning accuracy across varying temporal horizons.}
        \label{fig:planagent_length_curve}
    \end{minipage}\hfill
    \begin{minipage}[c]{0.50\linewidth}
        \centering
        \includegraphics[width=\linewidth]{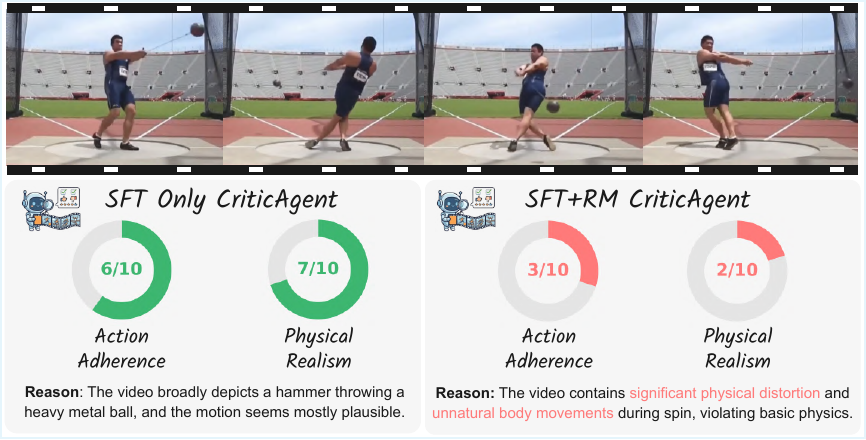}
        \vspace{-1.2em}
        \caption{\textbf{CriticAgent Discrimination.} Reward Modeling induces sharper scores for success-failure separation.}
        \label{fig:criticagent_score_dist}
    \end{minipage}

    \vspace{0.4em}

    \begin{minipage}{1.0\textwidth}
        \centering
        \includegraphics[width=0.96\linewidth]{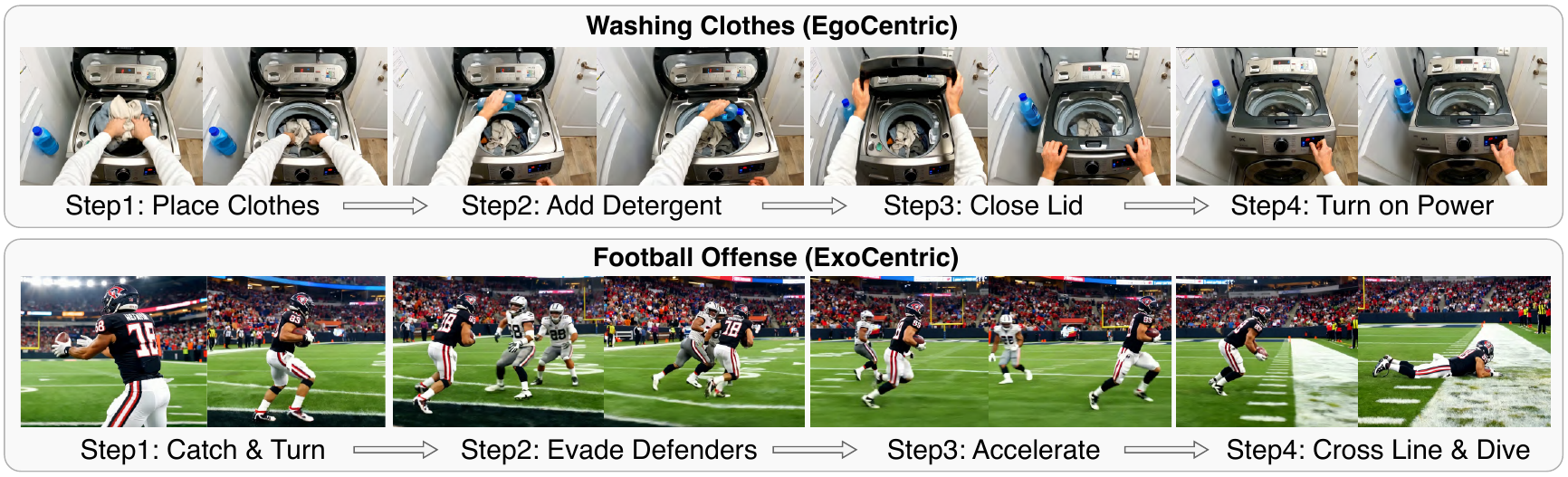}
        \vspace{-0.2em}
        \captionof{figure}{\textbf{Long-horizon Procedural Generation} results covering egocentric behavior generation and exocentric human kinematics synthesis, demonstrating coherent action execution across extended procedural steps.}
        \label{fig:vis_long_horizon_egoexo}
    \end{minipage}
    \vspace{-3.8mm}
\end{figure}

\noindent \textbf{Evaluation Benchmarks.}
We evaluate different components of our framework across three benchmarks:
\textbf{(1) EgoPlan-Bench}~\cite{EgoPlan-Bench} assesses the \emph{PlanAgent} on long-horizon procedural planning, using task success rate as the primary metric;
\textbf{(2) VideoGen-RewardBench}~\cite{VideoReward} evaluates the \emph{CriticAgent} as a video reward model, quantifying its alignment with human judgments via preference accuracy; and
\textbf{(3) \ourBenchmark{} (Ours)} holistically evaluates the video quality of \emph{VideoGenerator} across multiple dimensions, including visual fidelity, temporal coherence, and action accuracy under varying difficulty levels.

\subsection{Evaluations on PlanAgent}
\label{sec:exp_planagent}
\noindent\textbf{Superiority in Long-Horizon Planning.}
Table~\ref{tab:planagent_bench} reports PlanAgent's performance on EgoPlan-Bench~\cite{EgoPlan-Bench}. Across in-domain and out-of-domain (OOD; EPIC-KITCHENS to Ego4D) splits, integrating context memory boosts accuracy from 35.81\% to 43.63\%, demonstrating that preserving historical state is vital for long-horizon consistency. While our method achieves higher in-domain accuracy than GRPO-CARE~\cite{chen2025grpo}, GRPO-CARE generalizes better to OOD data. This likely occurs because GRPO fosters broader state exploration in the OOD domain, whereas our DPO strictly aligns with the training distribution.

\noindent\textbf{Impact of Hierarchical Training Stages.}
Ablating the training process highlights the complementary roles of both optimization phases. Instruction Tuning (IT) yields a 9.66\% gain over the memory-augmented baseline, confirming the necessity of structured supervision for explicit CoT reasoning. Subsequently, Preference Alignment via DPO adds a 5.43\% improvement, acting as a critical physical regularizer against logical inconsistencies and hallucinations. Ultimately, our two-stage training boosts overall accuracy to 58.72\%, outperforming GRPO-CARE~\cite{chen2025grpo} by 3.94\%.

\noindent\textbf{Robustness To Extended Horizons.}
Figure~\ref{fig:planagent_length_curve} details accuracy across varying task lengths. Without memory, performance degrades sharply from 44.5\% to 36.0\%. Conversely, integrating the memory module maintains remarkable stability, sustaining 43.02\% accuracy at extreme horizons. This empirically demonstrates that preserving historical context is critical to retaining long-term dependencies in long-range planning.

\begin{figure*}[t]
    \centering
    \begin{minipage}{\linewidth}
        \centering
        \renewcommand{\arraystretch}{0.85}
        \small
        \centering
        \captionof{table}{\textbf{Impact of SPIRAL on Long-Horizon Action-Conditioned Video Generation.} We integrate diverse T2V and I2V methods with SPIRAL, assessing static, dynamic, and action quality on VBench and \ourBenchmark{}.}
        \label{tab:videogen_bench}
        \vspace{-.3em}
        \resizebox{\linewidth}{!}{
            \begin{tabular}{l|cc|cc|ccc|cccc}
            \toprule
            \multirow{2}{*}[-0.8em]{\textbf{Method}} & \multicolumn{2}{c|}{\textbf{Video Params}} & \multicolumn{2}{c|}{\textbf{Static Quality}} & \multicolumn{3}{c|}{\textbf{Dynamic Quality}} & \multicolumn{4}{c}{\textbf{Action Quality}} \\
            
            \cmidrule(lr){2-12}
            
             & Res. & FPS &
             \begin{tabular}[c]{@{}c@{}}Aesthetic\\ Quality\end{tabular} & 
             \begin{tabular}[c]{@{}c@{}}Imaging\\ Quality\end{tabular} & 
             \begin{tabular}[c]{@{}c@{}}Subject\\ Consistency\end{tabular} & 
             \begin{tabular}[c]{@{}c@{}}Background\\ Consistency\end{tabular} & 
             \begin{tabular}[c]{@{}c@{}}Motion\\ Smoothness\end{tabular} & 
             \begin{tabular}[c]{@{}c@{}}Action\\ Completeness\end{tabular} & 
             \begin{tabular}[c]{@{}c@{}}Action\\ Smoothness\end{tabular} & 
             \begin{tabular}[c]{@{}c@{}}Object\\ Interaction\end{tabular} & 
             \begin{tabular}[c]{@{}c@{}}Physical\\ Fidelity\end{tabular} \\
            \midrule
            
            \multicolumn{12}{c}{\textbf{Text-to-Video (T2V)}} \\ 
            \midrule
            LongLive~\cite{LongLive} & 480P & 16 & 0.549 & 0.735 & 0.937 & 0.946 & 0.990 & 0.273 & 0.646 & 0.776 & 0.808 \\
            \rowcolor{tabcolor!25} LongLive~\cite{LongLive} + \ourMethodstr{} (Ours) & 480P & 16 & 0.554 & 0.727 & 0.936 & 0.950 & 0.990 & \textbf{0.721}$\mathrm{\scriptstyle{\hi{+44.8\%}}}$ & \textbf{0.902}$\mathrm{\scriptstyle{\hi{+25.6\%}}}$ & \textbf{0.860}$\mathrm{\scriptstyle{\hi{+8.4\%}}}$ & \textbf{0.924}$\mathrm{\scriptstyle{\hi{+11.6\%}}}$ \\
            Wan2.1~\cite{wan2025wan} & 480P & 16 & 0.560 & 0.729 & 0.919 & 0.944 & 0.985 & 0.565 & 0.834 & 0.756 & 0.894 \\
            \rowcolor{tabcolor!25} Wan2.1~\cite{wan2025wan} + \ourMethodstr{} (Ours) & 480P & 16 & 0.568 & 0.717 & 0.931 & 0.953 & 0.988 & \textbf{0.875}$\mathrm{\scriptstyle{\hi{+31.0\%}}}$ & \textbf{0.918}$\mathrm{\scriptstyle{\hi{+8.4\%}}}$ & \textbf{0.926}$\mathrm{\scriptstyle{\hi{+17.0\%}}}$ & \textbf{0.958}$\mathrm{\scriptstyle{\hi{+6.4\%}}}$ \\
            Sora~\cite{Sora} & 720P & 30 & 0.511 & 0.649 & 0.898 & 0.926 & 0.991 & 0.378 & \textbf{0.820} &  0.894  & \textbf{0.852} \\
            \rowcolor{tabcolor!25} Sora~\cite{Sora} + \ourMethodstr{} (Ours) & 720P & 30 & 0.503 & 0.648 & 0.838 & 0.898 & 0.991 & \textbf{0.540}$\mathrm{\scriptstyle{\hi{+16.2\%}}}$ & 0.714$\mathrm{\scriptstyle{\nohi{-10.6\%}}}$ & \textbf{0.910}$\mathrm{\scriptstyle{\hi{+1.6\%}}}$ & 0.844$\mathrm{\scriptstyle{\nohi{-0.8\%}}}$ \\
            Seedance1.0~\cite{gao2025seedance} & 1080P & 24 & 0.567 & 0.752 & 0.928 & 0.947 & 0.988 & 0.879 & 0.962 & 0.944 & 0.980 \\
            \rowcolor{tabcolor!25} Seedance1.0~\cite{gao2025seedance} + \ourMethodstr{} (Ours) & 1080P & 24 & 0.578 & 0.764 & 0.936 & 0.949 & 0.989 & \textbf{0.956}$\mathrm{\scriptstyle{\hi{+7.7\%}}}$ & \textbf{0.980}$\mathrm{\scriptstyle{\hi{+1.8\%}}}$  & \textbf{0.962}$\mathrm{\scriptstyle{\hi{+1.8\%}}}$ & \textbf{0.988}$\mathrm{\scriptstyle{\hi{+0.8\%}}}$ \\
            \midrule
            \multicolumn{12}{c}{\textbf{Image-to-Video (I2V)}} \\
            \midrule
            Wan2.1~\cite{wan2025wan} & 480P & 16 & 0.527 & 0.672 & 0.873 & 0.915 & 0.982 & 0.722 & 0.860 & 0.818 & 0.892 \\
            \rowcolor{tabcolor!25} Wan2.1~\cite{wan2025wan} + \ourMethodstr{} (Ours) & 480P & 16 & 0.528 & 0.679 & 0.858 & 0.897 & 0.984 & \textbf{0.827}$\mathrm{\scriptstyle{\hi{+10.5}}}$ & \textbf{0.934}$\mathrm{\scriptstyle{\hi{+7.4\%}}}$ & \textbf{0.894}$\mathrm{\scriptstyle{\hi{+7.6\%}}}$ & \textbf{0.970}$\mathrm{\scriptstyle{\hi{+7.8\%}}}$ \\
            Wan2.2~\cite{wan2025wan} & 480P & 24 & 0.492 & 0.642 & 0.892 & 0.925 & 0.974 & 0.709 & 0.916 & 0.818 & 0.934 \\
            \rowcolor{tabcolor!25} Wan2.2~\cite{wan2025wan} + \ourMethodstr{} (Ours) & 480P & 24 & 0.469 & 0.640 & 0.861 & 0.884 & 0.975 & \textbf{0.726}$\mathrm{\scriptstyle{\hi{+1.7\%}}}$ & \textbf{0.918}$\mathrm{\scriptstyle{\hi{+0.2\%}}}$ & \textbf{0.834}$\mathrm{\scriptstyle{\hi{+1.6\%}}}$ & \textbf{0.942}$\mathrm{\scriptstyle{\hi{+0.8\%}}}$ \\
            SVI~\cite{SVI} & 480P & 16 & 0.505 & 0.636 & 0.838 & 0.903 & 0.977 & 0.233 & 0.746 & 0.816  & 0.782 \\
            \rowcolor{tabcolor!25} SVI~\cite{SVI} + \ourMethodstr{} (Ours) & 480P & 16 & 0.509 & 0.649 & 0.846 & 0.902 & 0.984 & \textbf{0.353}$\mathrm{\scriptstyle{\hi{+12.0\%}}}$ & \textbf{0.754}$\mathrm{\scriptstyle{\hi{+0.8\%}}}$ & \textbf{0.838}$\mathrm{\scriptstyle{\hi{+2.2\%}}}$  & \textbf{0.808}$\mathrm{\scriptstyle{\hi{+2.6\%}}}$ \\
            Kling2.1~\cite{Kling} & 720P & 24 & 0.542 & 0.706 & 0.904 & 0.932 & 0.989 & 0.042 & 0.702 & 0.704 & 0.848 \\
            \rowcolor{tabcolor!25} Kling2.1~\cite{Kling} + \ourMethodstr{} (Ours) & 720P & 24 & 0.550 & 0.709 & 0.910 & 0.919 & 0.991 & \textbf{0.424}$\mathrm{\scriptstyle{\hi{+38.2\%}}}$ & \textbf{0.896}$\mathrm{\scriptstyle{\hi{+19.4\%}}}$ & \textbf{0.878}$\mathrm{\scriptstyle{\hi{+17.4\%}}}$ & \textbf{0.878}$\mathrm{\scriptstyle{\hi{+3.0\%}}}$ \\
            
            \bottomrule
            \end{tabular}
        }
        
        \vspace{0.8em}
        
        \includegraphics[width=0.95\linewidth]{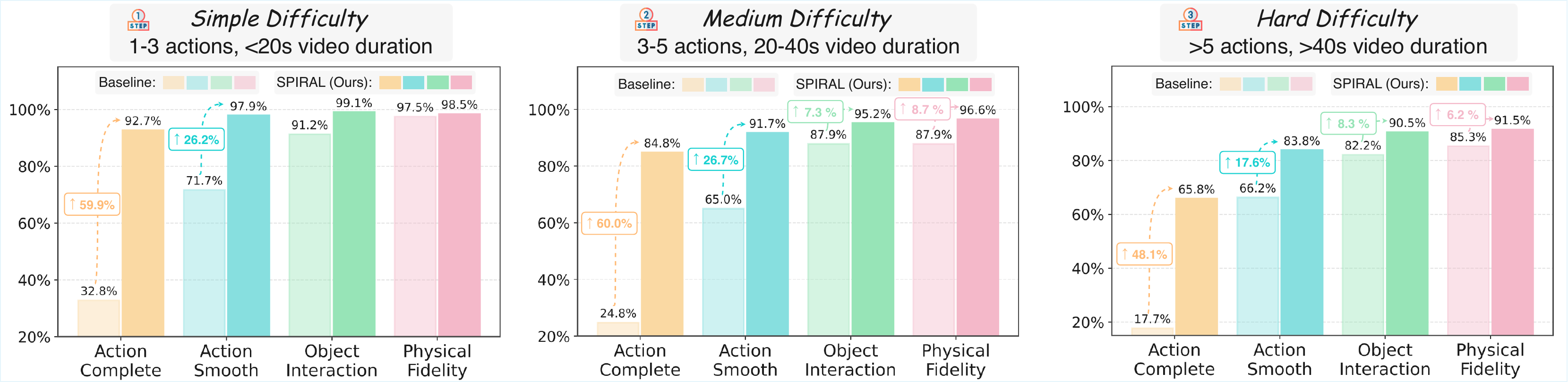}
        \vspace{-0.2em}
        \captionof{figure}{\textbf{Long-Horizon Video Generation Across Difficulties.} SPIRAL maintains high stability across simple, medium, and hard levels, mitigating the baseline's performance collapse on complex long-horizon tasks.}
        \label{fig:videogen_difficulty}
        
        \vspace{0.6em}

        \includegraphics[width=0.95\linewidth]{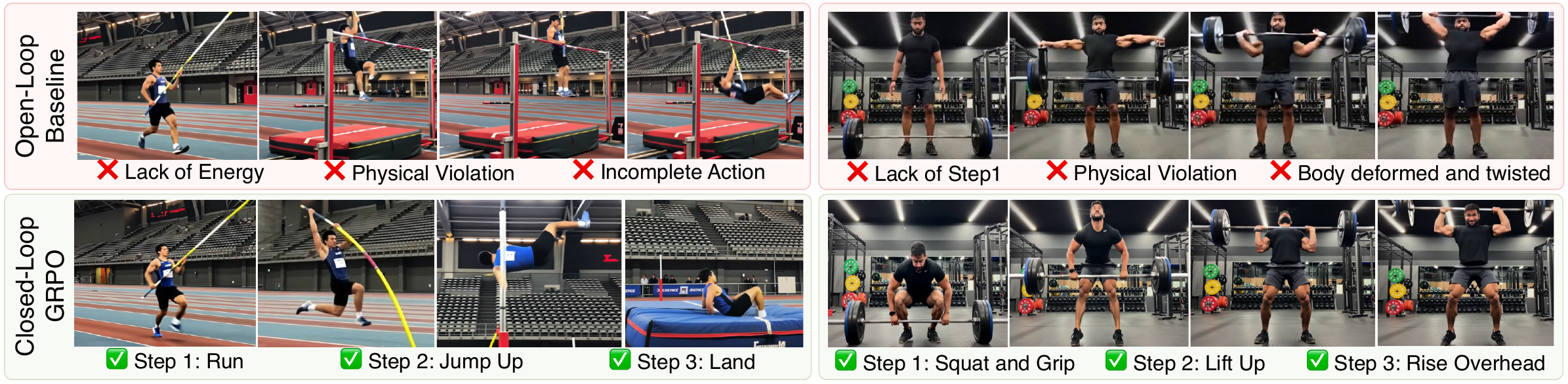}
        \vspace{-0.2em}
        \captionof{figure}{\textbf{Qualitative Comparison of Closed-Loop Self-Evolving.} Compared with the open-loop baseline, closed-loop GRPO produces more complete, temporally coherent, and physically plausible multi-step actions.}
        \label{fig:vis_compare_grpo}
        
    \end{minipage}
    \vspace{-3mm}
\end{figure*}

\subsection{Evaluations on CriticAgent}
\label{sec:exp_criticagent}
\noindent\textbf{Superiority in Reward Assessment.}
Table~\ref{tab:criticagent_bench} evaluates CriticAgent on the VideoGen-RewardBench~\cite{VideoReward}. It consistently outperforms previous methods in visual and motion quality, yielding a notable 5.37\% gain in Text Alignment. This improvement stems from our reward modeling optimization on the GAIA dataset~\cite{GAIA}, which enhances sensitivity to text-action alignment. Ultimately, CriticAgent achieves an overall accuracy of 68.86\%, validating our multi-dimensional scoring mechanism jointly captures visual fidelity and physical realism.

\noindent\textbf{Impact of Training Stages.}
Ablating the training stages reveals that while SFT provides a robust initialization by distilling foundation model judgments, the subsequent Reward Modeling (RM) phase is critical for discrimination. Integrating preference learning delivers a 5.7\% boost in overall performance, demonstrating its necessity for distinguishing execution failures, assessing text-video alignment, and ranking subtle action-quality differences.

\noindent\textbf{Enhanced Discriminative Capability.}
Figure~\ref{fig:criticagent_score_dist} illustrates the scoring distributions before and after incorporating reward modeling (RM). Following SFT, evaluations for failed executions are insufficiently sharp, leaving ambiguous penalties for incomplete or misaligned actions. Integrating reward modeling heavily polarizes these scores, causing failures to peak sharply near the minimum score.
This calibration provides sharper, definitive reward signals for action adherence and physical plausibility, which are essential for subsequent GRPO-based optimization.

\subsection{Evaluations on Closed-Loop Video Generation}
\label{sec:exp_videogen}
\noindent\textbf{Superiority in Action Execution.}
As shown in Table~\ref{tab:videogen_bench}, SPIRAL significantly enhances diverse T2V and I2V backbones. By leveraging the \textit{PlanAgent} to decompose complex goals into executable actions and the \textit{CriticAgent} to rectify execution errors, our closed-loop framework substantially improves overall Action Quality. Notably, action completeness achieves remarkable gains (e.g., +44.8\% for LongLive), while other aspects also improve across most backbones, demonstrating that explicit planning and verification are crucial for reliable long-horizon generation.

\noindent\textbf{Analysis of Generic Metrics.}
We observe minor fluctuations in generic visual and dynamic metrics, which are inherently biased toward short-term fidelity and limited motions. This bias inadvertently rewards open-loop baselines that fail to execute complete actions and collapse into short dynamics. In contrast, SPIRAL successfully executes full long-horizon tasks through staged atomic actions. Consequently, our framework achieves substantial gains in action completeness and procedural correctness, the true paramount metrics for long-horizon execution.

\noindent\textbf{Qualitative Results.}
Fig.~\ref{fig:vis_long_horizon_egoexo} visualizes closed-loop generation across egocentric interaction and exocentric human kinematics. As demonstrated, explicit planning and verification ensure coherent multi-step execution for both first-person manipulation and third-person motion control. Additional analyses, including ultra-long procedural chains and open-loop vs. closed-loop comparisons, are detailed in Appendix~\ref{app:additional_qualitative}.

\noindent\textbf{Difficulty Stratification.}
Fig.~\ref{fig:videogen_difficulty} details performance across task difficulties (Simple, Medium, Hard). While baselines adequately handle simple tasks, they collapse on hard tasks with extended durations (e.g., >40s) and long-range action dependencies. Conversely, \ourMethod{} remains highly stable across all sequence lengths, confirming that explicit planning, memory preservation, and iterative verification are essential for reliable long-horizon synthesis.

\begin{figure*}[t]
    \centering
    \begin{minipage}{0.49\textwidth}
        \centering
        \includegraphics[width=\linewidth]{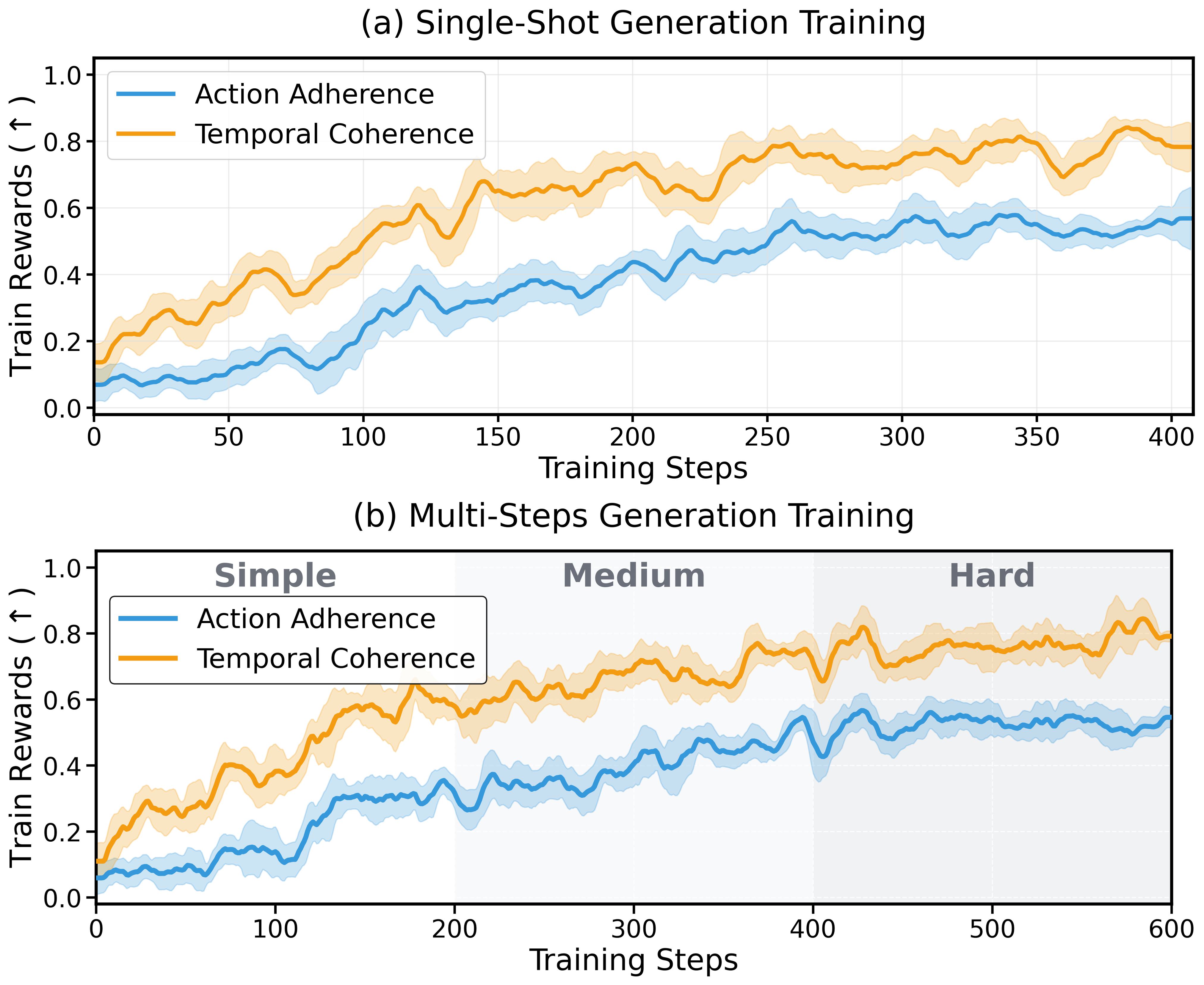}
        \vspace{-1.6em}
        \captionof{figure}{\textbf{GRPO Reward Convergence}. Curriculum learning (b) ensures continuous policy improvement across increasing task complexities and longer horizons.}
        \label{fig:grpo_reward_curve}
    \end{minipage}
    \hfill  
    \begin{minipage}{0.49\textwidth}
        \centering
        \renewcommand{\arraystretch}{1.0}
        \setlength{\tabcolsep}{3pt}
        \small
        \captionof{table}{Performance across Different Training Stages.}
        \label{tab:grpo_stage_results}
        \vspace{-0.5em}
        \resizebox{\linewidth}{!}{
            \begin{tabular}{l|cccc|c}
            \toprule
            \textbf{Training Stage} & \textbf{\begin{tabular}[c]{@{}c@{}}Action\\ Compl.\end{tabular}} & \textbf{\begin{tabular}[c]{@{}c@{}}Action\\ Smooth.\end{tabular}} & \textbf{\begin{tabular}[c]{@{}c@{}}Object\\ Interact.\end{tabular}} & \textbf{\begin{tabular}[c]{@{}c@{}}Physical\\ Fidelity\end{tabular}} & \textbf{\begin{tabular}[c]{@{}c@{}}Overall\\ Quality\end{tabular}} \\
            \midrule
            \textbf{SFT Only} & 0.273 & 0.646 & 0.776 & 0.808 & 0.626 \\
            \textbf{SFT + Closed Loop} & 0.721$\mathrm{\scriptstyle{\hi{+44.8\%}}}$ & 0.902$\mathrm{\scriptstyle{\hi{+25.6\%}}}$ & 0.860$\mathrm{\scriptstyle{\hi{+8.4\%}}}$ & 0.924$\mathrm{\scriptstyle{\hi{+11.6\%}}}$ & 0.852$\mathrm{\scriptstyle{\hi{+22.6\%}}}$ \\
            \rowcolor{tabcolor!25} \textbf{SFT + CL + GRPO} & \textbf{0.841}$\mathrm{\scriptstyle{\hi{+56.8\%}}}$ & 0.942$\mathrm{\scriptstyle{\hi{+29.6\%}}}$ & 0.864$\mathrm{\scriptstyle{\hi{+8.8\%}}}$ & 0.936$\mathrm{\scriptstyle{\hi{+12.8\%}}}$ & 0.896$\mathrm{\scriptstyle{\hi{+27.0\%}}}$ \\ 
            \bottomrule
            \end{tabular}
        }
        
        \vspace{0.9em}
        
        \includegraphics[width=\linewidth]{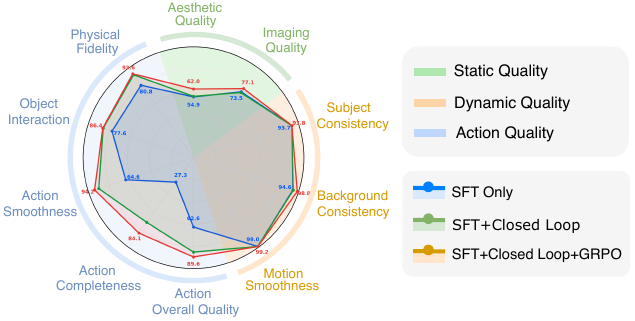}
        \vspace{-1.0em}
        \captionof{figure}{\textbf{Performance Gains via Closed Loop and GRPO.} SFT provides initial gains, while closed-loop feedback and GRPO further enhance action quality.}
        \label{fig:grpo_radar}
    \end{minipage}
    \vspace{-0.9em}
\end{figure*}
\begin{table}[t!]
    \centering
    \small
    \renewcommand{\arraystretch}{1.0}
    \begin{minipage}[t]{0.51\linewidth}
        \centering
        \caption{\textbf{Ablation on SPIRAL Modules.} Impact of integrating PlanAgent for task decomposition and CriticAgent for dual-level inner/outer feedback on action quality.}
        \vspace{0.4em}
        \label{tab:ablation_plan_critic}
        \setlength{\tabcolsep}{3pt} 
        \resizebox{\linewidth}{!}{
            \begin{tabular}{l|cc|cc|cccc}
                \toprule
                \multirow{2}{*}[-0.3em]{\textbf{Method}} & \multicolumn{2}{c|}{\textbf{Agent}} & \multicolumn{2}{c|}{\textbf{Feedback}} & \multicolumn{4}{c}{\textbf{Action Quality}} \\
                \cmidrule(lr){2-3} \cmidrule(lr){4-5} \cmidrule(lr){6-9}
                & \textbf{Plan} & \textbf{Critic} & \textbf{Inner} & \textbf{Outer} & \textbf{Compl.} & \textbf{Smooth.} & \textbf{Interact.} & \textbf{Fidelity} \\
                \midrule
                Baseline & & & & & 0.347 & 0.796 & 0.804 & 0.820 \\
                Plan only & $\checkmark$ & & & & 0.628 & 0.754 & 0.838 & 0.840 \\
                Plan + Critic & $\checkmark$ & $\checkmark$ & $\checkmark$ & & 0.715 & 0.816 & \textbf{0.880} & 0.848 \\
                \rowcolor{tabcolor!25} \textbf{Full SPIRAL} & $\checkmark$ & $\checkmark$ & $\checkmark$ & $\checkmark$ & \textbf{0.781} & \textbf{0.862} & 0.872 & \textbf{0.934} \\ 
                \bottomrule
            \end{tabular}
        }
    \end{minipage}\hfill
    \begin{minipage}[t]{0.47\linewidth}
        \centering
        \caption{\textbf{Ablation on Memory Horizons.} Our ContextMemory ensures stable long-horizon generation across varying memory lengths and temporal contexts.}
        \vspace{0.4em}
        \label{tab:ablation_memory}
        \setlength{\tabcolsep}{3pt} 
        \resizebox{\linewidth}{!}{
            \begin{tabular}{l|cc|ccc}
                \toprule
                \multirow{2}{*}[-0.3em]{\textbf{Memory}} & \multicolumn{2}{c|}{\textbf{Visual Consistency}} & \multicolumn{3}{c}{\textbf{Action Quality}} \\
                \cmidrule(lr){2-3} \cmidrule(lr){4-6}
                & \textbf{Subject} & \textbf{Background} & \textbf{Compl.} & \textbf{Smooth.} & \textbf{Fidelity} \\
                \midrule
                None ($I_0$) & 0.712 & 0.745 & 0.654 & 0.783 & 0.825 \\
                Last Frame & 0.835 & 0.852 & 0.712 & 0.815 & 0.868 \\
                Local Window & 0.890 & 0.884 & 0.748 & 0.841 & 0.902 \\
                \rowcolor{tabcolor!25} \textbf{ContextMem. (Ours)} & \textbf{0.938} & \textbf{0.941} & \textbf{0.781} & \textbf{0.862} & \textbf{0.934} \\ 
                \bottomrule
            \end{tabular}
        }
    \end{minipage}
    \vspace{-0.9em}
\end{table}

\subsection{Evaluations on GRPO-based Self-Evolving}
\label{sec:exp_grpo}
\noindent\textbf{Curriculum-Driven Training Dynamics.}
Fig.~\ref{fig:grpo_reward_curve} plots the GRPO reward curves for action adherence and temporal coherence. Driven by a \textbf{curriculum learning} regime (Fig.~\ref{fig:grpo_reward_curve}(b)) that progressively scales plan complexity and temporal horizons, the generator transitions from executing atomic actions to mastering complex procedural tasks. The steady convergence validates the efficacy of our critic-driven policy optimization under increasing horizons.

\noindent\textbf{Agent-Driven Optimization.}
By leveraging external reflective planning agents and critic-derived rewards, our self-evolving directly optimizes the generator's intrinsic weights (Fig.~\ref{fig:grpo_radar}). As detailed in Table~\ref{tab:grpo_stage_results}, GRPO optimization boosts action completeness (+12\%) and smoothness (+4\%) against the closed-loop baseline, proving that self-evolving translates inference-time verification and correctness into permanent model improvements.

\noindent\textbf{Qualitative Analysis.}
Fig.~\ref{fig:vis_compare_grpo} shows that the open-loop baseline suffers from incomplete motions and physical distortions, while closed-loop GRPO yields complete, temporally coherent sequences with more reliable object interactions. Additional qualitative results are provided in Appendix~\ref{app:qual_single_shot} and~\ref{app:qual_grpo} for comparisons.

\subsection{Ablation Study}
\label{sec:ablation}
\noindent\textbf{Ablation on PlanAgent and CriticAgent.}
Table~\ref{tab:ablation_plan_critic} ablates the planning and feedback mechanisms. PlanAgent significantly boosts Action Completeness (+28.1\%), demonstrating that explicit decomposition is vital for complex multi-step tasks and procedural ordering. Meanwhile, CriticAgent's dual-level feedback enhances overall quality through local refinement, failed-step correction, and global replanning. The full SPIRAL framework achieves optimal performance, proving that explicit planning and closed-loop verification are highly complementary.

\noindent\textbf{Ablation on ContextMemory.}
Table~\ref{tab:ablation_memory} evaluates varying memory horizons. Unlike limited-memory baselines that suffer from semantic drift, our global ContextMemory maintains visual consistency by preserving long-horizon visual priors, object identity, and scene layout. Furthermore, this persistent context ensures action continuity and cross-step consistency, thereby directly enhancing overall Action Quality (e.g., 3.3\% gains against local memory).

\section{Conclusion}
\label{sec:conclusion}
We present SPIRAL, a closed-loop think-act-reflect framework that significantly advances long-horizon action-conditioned video generation. SPIRAL synergizes a PlanAgent for task decomposition, a ContextMemory for visual persistence, and a CriticAgent for dual-level verification, effectively preventing semantic drift and incomplete execution. Moving beyond inference-time correction, we introduce a GRPO-based self-evolving strategy guided by curriculum learning, translating agent-driven feedback into permanent optimization of the video generator. We further contribute \ourDataset{} and \ourBenchmark{}, providing structured supervision and dedicated evaluation protocols for training and assessing long-horizon action-conditioned generation. Extensive experiments on \ourBenchmark{} confirm that SPIRAL consistently enhances action quality and temporal coherence across diverse T2V and I2V backbones, paving the way for robust long-horizon action-conditioned video synthesis.

\clearpage
\newgeometry{
  textheight=9in,
  textwidth=5.5in,
  top=1in,
  headheight=12pt,
  headsep=25pt,
  footskip=30pt
}
\newpage
\appendix
\onecolumn
\renewcommand{\thesection}{\Alph{section}}
\setcounter{section}{0}

\begin{center}
    \textbf{\Large \ourMethodColored: Self-Evolving Action-Conditioned}\\[0.5em]
    \textbf{\Large Video Generation via Reflective Planning Agents}\\[1.8em]
    \large Supplementary Material
\end{center}

\vspace{0.5em}
{\large \textbf{Contents}}

\startcontents[appendices]
\printcontents[appendices]{l}{1}{\setcounter{tocdepth}{3}}

\clearpage
\section{Additional Related Works}
\label{app:related_work}

\subsection{Agentic Visual Generation}

Recent visual generation research is moving beyond one-shot generation toward agentic systems that treat generation as one step in a closed-loop decision process~\cite{wu2026visual}. Existing systems span a spectrum from workflow-style pipelines to more autonomous visual agents. For example, GEMS~\cite{he2026gems} introduces a planner--decomposer--verifier--refiner workflow equipped with memory and skill selection for multimodal generation, while Gen-Searcher~\cite{feng2026gen} extends this paradigm with agentic web search and evidence collection for knowledge-intensive visual tasks. In the editing domain, JarvisArt~\cite{lin2025jarvisart} maps user intent into executable photo-retouching operations through reasoning and tool use, demonstrating that agentic generation can be embedded within professional visual software rather than merely producing pixels. Related embodied systems, such as CoT-VLA~\cite{zhao2025cot} and UniPi~\cite{du2023learning}, further treat visual generation or prediction as part of a perception--action loop for downstream control. Overall, these works indicate that the central challenge of agentic visual generation is shifting from rendering fidelity alone toward reliable planning, grounded verification, memory preservation, and self-correction over long-horizon visual trajectories.

\subsection{Agentic Video Generation}
Recent work has explored incorporating chain-of-thought (CoT) reasoning and intermediate planning into video generation to improve consistency and controllability~\cite{guo2025can}. These approaches typically decompose high-level prompts into structured intermediate signals, and have evolved from pre-planning~\cite{liao2025imagegen} or post-refinement~\cite{zhuo2025reflection} toward “thinking while generating” paradigms~\cite{guo2025thinking}. Representative examples include EditThinker~\cite{li2025editthinker}, which frames image editing as iterative reasoning, while VideoAgent~\cite{soni2025videoagent}, CoAgent~\cite{zeng2025coagent}, and A$^2$RD~\cite{long20262} cast video generation as a closed-loop, agentic process with planning, synthesis, and verification. In contrast to these works, we explicitly formulate a closed-loop framework with reflective planning agents and an explicit memory module, enabling sustained grounding of high-level actions over long temporal horizons.

\subsection{Self-Evolving Video Generation}
Recent years have witnessed growing interest in self-evolving paradigms across code generation~\cite{zhao2025absolute}, question answering~\cite{huang2025r, wang2025vision}, and agentic search~\cite{wu2025evolver, fu2025re}. A common theme is to leverage either internal signals or external feedback to progressively enhance solution quality. In video generation, early attempts have begun to extend these ideas to long-horizon generation and planning. VideoAgent~\cite{soni2025videoagent} proposes a self-improving video planning framework that iteratively refines text-to-video generation via self-conditioned consistency, employs a VLM to guide plan selection, and leverages online environment feedback to mitigate hallucinations. VISTA~\cite{long2025vista} further advances this line of work by introducing a test-time, multi-agent iterative self-improvement framework that emulates human-like prompt refinement and jointly optimizes visual, audio, and contextual aspects of video generation through structured planning, critique, and targeted revision. In this work, we propose a self-evolution strategy using group relative policy optimization, enabling closed-loop reinforcement learning driven by critic-derived signals.

\subsection{Video World Models}
World models aim to learn the dynamics of the environment by predicting future observations conditioned on past observations and actions~\cite{ha2018world,chu2026agentic}. Early studies focused on games and simulated environments~\cite{alonso2024diffusion}, with recent progress extending to robotics~\cite{zhang2025world,liao2025genie}, autonomous driving~\cite{kong20253d,liang2025worldlens,yang2025x}, and embodied settings. More recently, fueled by rapid progress in video foundation models~\cite{yang2024cogvideox, Wan}, video world models have advanced toward higher long-horizon generation quality and temporal consistency~\cite{mei2024dreamforge, wu2025video}, faster inference for interactive or online use~\cite{LongLive,chen2025sana}, richer and more diverse action controllability~\cite{tang2025hunyuan}, and improved physical fidelity in motion, dynamics, and contact interactions~\cite{agarwal2025cosmos}. Unlike prior works that primarily use text prompts to describe global scenes or short-term changes~\cite{henschel2025streamingt2v}, our work focuses on persistent, object-grounded control, treating language as high-level semantic actions driving long-horizon behavior.

\clearpage
\section{Additional Implementation Details of SPIRAL} 
We provide comprehensive implementation details of our SPIRAL framework, elaborating on the PlanAgent (Sec.~\ref{app:plan_agent}), VideoGenerator (Sec.~\ref{app:video_generator}), CriticAgent (Sec.~\ref{app:critic_agent}), inference-time closed-loop pipeline (Sec.~\ref{app:closed_loop}), and training-time GRPO-based self-evolving strategy (Sec.~\ref{app:grpo}).

\subsection{PlanAgent Implementation Details}
\label{app:plan_agent}
We detail the PlanAgent ($\pi_{\text{plan}}$), covering its architecture, the two-stage training protocol (SFT and DPO), preference data construction, and closed-loop inference logic.

\begin{table}[htbp]
    \centering
    \small
    \renewcommand{\arraystretch}{1.05}
    \setlength{\tabcolsep}{4pt}
    \begin{threeparttable}
        \caption{\textbf{Hyperparameter settings for PlanAgent training}. Parameters are differentiated by two training phases: Supervised Fine-Tuning (SFT) and Direct Preference Optimization (DPO).}
        \label{tab:plan_hyperparameters}
        \begin{tabularx}{0.95\textwidth}{@{}lccX@{}}
            \toprule
            \textbf{Parameter} & \textbf{SFT} & \textbf{DPO} & \textbf{Description} \\
            \midrule
            \rowcolor{tabcolor!20}
            \multicolumn{4}{@{}l}{\textbf{\textit{Optimization \& Training Strategy}}} \\
            Base model & \multicolumn{2}{c}{Qwen3-VL-8B-Thinking} & Backbone vision-language model \\
            Learning rate & $1.0 \times 10^{-4}$ & $5.0 \times 10^{-6}$ & Peak learning rate for AdamW \\
            LR scheduler & Cosine & Cosine & Decay schedule type \\
            Warmup ratio & 0.1 & 0.1 & Ratio of warmup steps \\
            Max epochs & 3.0 & 3.0 & Number of training epochs \\
            Batch size/GPU & 1 & 1 & Per-device batch size \\
            Gradient accumulation & 8 & 8 & Steps for effective batch construction \\
            Precision & bf16 & bf16 & Mixed-precision training format \\
            Max sequence length & 8192 & 8192 & Maximum context length in tokens \\
            \midrule
            \rowcolor{tabcolor!20}
            \multicolumn{4}{@{}l}{\textbf{\textit{Model Architecture \& LoRA Configuration}}} \\
            LoRA rank ($r$) & 16 & 16 & Rank dimension for adaptation \\
            LoRA alpha ($\alpha$) & 32 & 32 & Scaling factor ($2 \times r$) \\
            LoRA dropout & 0.05 & 0.05 & Dropout probability for LoRA layers \\
            Target modules & \multicolumn{2}{l}{all linear} & Attention and MLP layers applied \\
            \midrule
            \rowcolor{tabcolor!20}
            \multicolumn{4}{@{}l}{\textbf{\textit{DPO-Specific Configuration}}} \\
            Beta ($\beta$) & - & 0.1 & KL penalty coefficient \\
            Loss function & - & Sigmoid & Preference loss type \\
            \bottomrule
        \end{tabularx}
    \end{threeparttable}
\end{table}

\noindent\textbf{Architecture and Training Configuration.}
We instantiate the PlanAgent using Qwen3-VL-8B~\cite{Qwen3-VL} as the backbone. To enable structured planning while preserving pre-trained multimodal capabilities, we apply Low-Rank Adaptation (LoRA) to all linear layers in the attention and feed-forward networks.

Training proceeds in two sequential stages. First, \textbf{Supervised Fine-Tuning (SFT)} teaches the model to produce explicit Chain-of-Thought (CoT) reasoning and to follow the required JSON schema. Second, \textbf{Direct Preference Optimization (DPO)} further aligns the model with physical plausibility and temporal consistency by contrasting valid plans against carefully constructed flawed alternatives. Both stages use the AdamW optimizer with a cosine scheduler and a global batch size of 8 via gradient accumulation. We set the context window to 8,192 tokens to accommodate interleaved visual history and long reasoning traces. Table~\ref{tab:plan_hyperparameters} summarizes the hyperparameters.

\noindent\textbf{DPO Preference Construction.}
A critical component of DPO training is constructing high-quality preference pairs $(y_w, y_l)$ that expose hallucinations, invalid object interactions, and temporal inconsistencies. Unlike standard datasets that provide only ground-truth plans, we employ a teacher-student distillation pipeline to synthesize paired responses from \ourDataset{}:

\begin{itemize}[leftmargin=1.25em, itemsep=0pt, topsep=0pt]
    \item \textbf{Chosen Response ($y_w$).} We prompt a strong teacher (GPT-5.1) to generate a preferred response that first decomposes the goal through a rigorous CoT rationale inside \textit{\textless think\textgreater} tags, and then outputs a valid JSON plan. The resulting plan must realize the global goal depicted in the ground-truth video.

    \item \textbf{Rejected Response ($y_l$).} We prompt the teacher to generate a hard negative response that appears plausible locally but fails to complete the goal because of injected planning errors. These errors include missing preconditions (e.g., pouring without opening), temporal disorder, invalid object interactions, or object hallucinations.
\end{itemize}

This model-in-the-loop synthesis makes the rejected responses subtle and contextually relevant, providing stronger learning signals than random negative sampling. Before DPO training, all generated pairs are filtered to ensure strict schema adherence.

\paragraph{Inference Logic.}
The PlanAgent supports two inference modes, corresponding to initial planning and feedback-driven correction.

\begin{itemize}[leftmargin=1.25em, itemsep=0pt, topsep=0pt]
    \item \textbf{Standard Planning Mode.} At the beginning of a task, the agent receives the goal and visual observations, including the current frame and uniformly sampled historical frames. We generate an action plan via temperature sampling, then parse the output to extract the reasoning trace and the structured step list $\mathcal{S}$. To simplify downstream execution, the agent is constrained to output only the \emph{\textless think\textgreater}-block and the JSON object.

    \item \textbf{Closed-Loop Re-planning Mode.} When the CriticAgent detects a failure, the PlanAgent enters a re-planning state conditioned on a structured \textit{Failure Analysis} context. This context contains the \textit{execution history}, the \textit{failed attempt}, and the \textit{critical diagnosis}. The agent follows a diagnose-and-fix procedure: it rewrites the failed step according to the critique and revises subsequent steps to preserve logical continuity. The revised output starts from the failed step ID, enabling seamless replacement of the remaining plan.
\end{itemize}

\paragraph{System Prompts.}
We provide the exact system prompts used to steer the PlanAgent in both modes, designed to enforce the structured output format and specific reasoning requirements.

\definecolor{PlanAgentBack}{HTML}{F0F7FF}
\definecolor{PlanAgentFrame}{HTML}{A9D3FF}
\definecolor{PlanAgentText}{HTML}{0B74E5}
\definecolor{PlanAgentAccent}{HTML}{0084FF}

\tcbset{
  planagentboxstyle/.style={
    enhanced,
    boxrule=0.8pt,
    arc=3pt,
    boxsep=2pt,
    left=5pt,
    right=5pt,
    top=4pt,
    bottom=4pt,
    colback=PlanAgentBack,
    fonttitle=\bfseries\color{PlanAgentText},
    title=System Prompt,
    toptitle=1mm,
    bottomtitle=1mm,
    coltitle=PlanAgentText,
    colframe=PlanAgentFrame,
    colbacktitle=PlanAgentFrame!45,
    fontupper=\footnotesize\linespread{0.95}\selectfont,
    before upper={\setlist[itemize]{itemsep=0pt, parsep=0pt, topsep=0pt, partopsep=0pt, leftmargin=10pt}},
    breakable
  }
}

\tcbset{
  planagentjsonbox/.style={
    enhanced,
    colframe=PlanAgentFrame,
    colback=white,
    colbacktitle=PlanAgentFrame!35,
    coltitle=PlanAgentText,
    fonttitle=\small\bfseries\color{PlanAgentText},
    boxrule=0.8pt,
    arc=3pt,
    left=4pt,
    right=4pt,
    top=4pt,
    bottom=4pt,
    boxsep=5pt,
    listing only,
    listing options={
      language=json,
      basicstyle=\footnotesize\ttfamily,
      keywordstyle=\color{PlanAgentAccent},
      stringstyle=\color{PlanAgentText},
      commentstyle=\color{gray},
      morekeywords={true,false,null},
      literate=%
        *{,}{{\textcolor{PlanAgentAccent}{,}}}{1}%
         {:}{{\textcolor{PlanAgentAccent}{:}}}{1}%
         {\{}{{\textcolor{PlanAgentAccent}{\{}}}{1}%
         {\}}{{\textcolor{PlanAgentAccent}{\}}}}{1}%
    },
    breakable,
  }
}

\lstdefinelanguage{json}{
    basicstyle=\ttfamily\footnotesize,
    showstringspaces=false,
    breaklines=true,
    breakatwhitespace=true,
    morestring=[b]",
    stringstyle=\color{PlanAgentText},
    morecomment=[s]{/*}{*/},
    commentstyle=\color{gray},
    morekeywords={true,false,null},
    keywordstyle=\color{PlanAgentAccent},
    literate=%
      *{,}{{\textcolor{PlanAgentAccent}{,}}}{1}%
       {:}{{\textcolor{PlanAgentAccent}{:}}}{1}%
       {\{}{{\textcolor{PlanAgentAccent}{\{}}}{1}%
       {\}}{{\textcolor{PlanAgentAccent}{\}}}}{1}%
}

\begin{itemize}[leftmargin=1.25em]
  \item \textbf{Standard Planning Prompt.} This prompt is used for the initial decomposition of the global goal into atomic actions.

\begin{tcolorbox}[planagentboxstyle, title=System Prompt for PlanAgent (Standard Planning), width=0.90\textwidth, center]

You are an \textbf{\textcolor{PlanAgentAccent}{Action Planner}} in a video generation system.
Your role is to plan a sequence of executable steps that will guide a world model to produce a video achieving the given goal from the current visual state.

\vspace{1ex}
\noindent\textbf{\textcolor{PlanAgentAccent}{Input Data:}}
\begin{itemize}[leftmargin=10pt]
  \item \textbf{\textcolor{PlanAgentText}{image}}: ONE current reference image (the most recent frame of the scene).
  \item \textbf{\textcolor{PlanAgentText}{history}}: A few history images showing previous context.
  \item \textbf{\textcolor{PlanAgentText}{GOAL}}: A textual description of what needs to be achieved.
\end{itemize}

\vspace{1ex}
\noindent\textbf{\textcolor{PlanAgentAccent}{Instructions:}}
\begin{itemize}[leftmargin=10pt]
  \item Carefully observe the provided images to understand the current state.
  \item Reason step by step inside \textbf{\textcolor{PlanAgentText}{\textless think\textgreater\ldots\textless/think\textgreater}} to plan how to achieve the GOAL.
  \item Output a valid JSON with a single key \texttt{steps} (array), describing the planned sequence.
  \item \textbf{\textcolor{PlanAgentAccent}{Constraint}}: Each step must include specific fields: \texttt{sid}, \texttt{text}, \texttt{actions}, \texttt{pre}, and \texttt{post}.
  \item Use lemma forms for verbs and nouns in the \texttt{actions} field.
\end{itemize}

\vspace{1ex}
\begin{tcolorbox}[planagentjsonbox, title=Expected Output Structure (JSON):]
\vspace{-3mm}
\begin{lstlisting}[language=json, basicstyle=\scriptsize\ttfamily, frame=none]
{
  "steps": [
    {
      "sid": 1,
      "action instruction": "Concise execution instruction (<= 36 words)",
      "actions": [
        {"verb": "lemma", "objects": ["obj1"], "tool": "tool_name"}
      ],
      "pre": ["pre-condition1", "pre-condition2"],
      "post": ["post-condition1", "post-condition2"]
    },
    ...
  ]
}
\end{lstlisting}
\vspace{-3mm}
\end{tcolorbox}

\end{tcolorbox}

\vspace{2ex}


  \item \textbf{Re-Planning Prompt.} This prompt is activated during the feedback loop to correct specific execution failures.

\begin{tcolorbox}[planagentboxstyle, title=System Prompt for PlanAgent (Closed-Loop Re-planning), width=0.90\textwidth, center]

You are an \textbf{\textcolor{PlanAgentAccent}{Expert Plan Editor}} for a procedural video generation system.
Your goal is to recover from a failure by creating a \textbf{\textcolor{PlanAgentAccent}{New, Corrected}} sequence of future actions based on visual evidence and error diagnosis.

\vspace{1ex}
\noindent\textbf{\textcolor{PlanAgentAccent}{Input Context:}}
\begin{itemize}[leftmargin=10pt]
  \item \textbf{\textcolor{PlanAgentText}{global\_goal}}: The final objective.
  \item \textbf{\textcolor{PlanAgentText}{failed\_attempt}}: The specific step definition that failed execution.
  \item \textbf{\textcolor{PlanAgentText}{critic\_feedback}}: Detailed natural language diagnosis of WHY the failure occurred.
  \item \textbf{\textcolor{PlanAgentText}{remaining\_steps}}: The original draft for the future steps.
\end{itemize}

\vspace{1ex}
\noindent\textbf{\textcolor{PlanAgentAccent}{Re-planning Rules (Strict):}}
\begin{itemize}[leftmargin=10pt]
  \item \textbf{\textcolor{PlanAgentAccent}{Diagnose and Fix}}: Analyze the \textbf{\textcolor{PlanAgentText}{critic\_feedback}}. You must discard the \texttt{failed\_attempt} and replace it with valid steps. Fix the step's \texttt{text}, \texttt{actions}, and \texttt{post-conditions}.
  \item \textbf{\textcolor{PlanAgentAccent}{Revise Future}}: Review the \texttt{remaining\_steps}. Ensure they logically follow your NEW corrected step or steps. You may modify, add, or delete steps to restore logical continuity.
  \item \textbf{\textcolor{PlanAgentAccent}{Formatting}}: The output sequence must start with the \textbf{Same SID} as the failed step.
\end{itemize}

\vspace{1ex}
\begin{tcolorbox}[planagentjsonbox, title=Example Output (Re-planning):]
\vspace{-3mm}
\begin{lstlisting}[language=json, basicstyle=\scriptsize\ttfamily, frame=none]
<think>
The critic indicates the previous attempt failed because the object was not grasped firmly. I need to insert a grasp adjustment step before the lift action.
</think>
{
  "steps": [
    {
      "sid": 3,  // Same ID as the failed step
      "text": "Adjust grip to ensure a firm hold on the handle.",
      "actions": [{"verb": "adjust", "objects": ["grip"]}],
      "pre": ["hand near object"],
      "post": ["firm grip established"]
    },
    {
      "sid": 4,  // Revised subsequent step
      "text": "Lift the object vertically.",
      ...
    }
  ]
}
\end{lstlisting}
\vspace{-3mm}
\end{tcolorbox}

\end{tcolorbox}
\end{itemize}

\subsection{VideoGenerator Implementation Details}
\label{app:video_generator}
The VideoGenerator is designed as a \textbf{flexible, model-agnostic module} that supports plug-and-play integration with Text-to-Video (T2V) and Image-to-Video (I2V) backbones. Its goal is to execute each structured action instruction produced by the PlanAgent and generate the corresponding video segment. We adapt generic video generation backbones using the structured instruction-action pairs from \ourDataset{} through Supervised Fine-Tuning (SFT), enabling fine-grained action control and long-horizon instruction following.

\noindent\textbf{Model Instantiation.}
To demonstrate the effectiveness of our SFT strategy, we present a concrete instantiation using Stable-Video-Infinity~\cite{SVI}, which leverages the Wan2.1-I2V-14B~\cite{Wan} as the foundational backbone. To adapt this model for precise action control while preserving its massive pre-trained generative prior, we apply Low-Rank Adaptation (LoRA) to the query, key, value, output projection, and feed-forward network layers ($q, k, v, o, ffn$) of the DiT blocks. We set the LoRA rank $r=64$ and alpha $\alpha=128$. The model processes video data encoded by the Wan2.1 VAE and text instructions via the T5-XXL encoder. Training is conducted using DeepSpeed Stage 2 with BF16 mixed precision to optimize memory efficiency.

\noindent\textbf{Multi-Resolution Training.}
Since \ourDataset{} contains videos with diverse aspect ratios, we adopt Multi-Resolution Training to improve generation quality across spatial formats. Training data is organized into resolution buckets, and the VideoGenerator is fine-tuned in three stages:

\begin{itemize}[leftmargin=1.25em, itemsep=0pt, topsep=0pt]
    \item \textbf{Stage 1: Base Adaptation.} We train on 480$\times$480 square clips for 10 epochs with a learning rate of $1e-4$. This stage establishes basic alignment between action instructions and motions.

    \item \textbf{Stage 2: Rectangular Adaptation.} We increase the resolution to 640$\times$480 and reduce the learning rate to $6e-5$ for 3 epochs. This stage adapts the generator to rectangular video layouts.

    \item \textbf{Stage 3: High-Resolution Adaptation.} We further fine-tune on 832$\times$480 inputs with a learning rate of $4e-5$ for 3 epochs. This stage improves high-resolution generation capacity.
\end{itemize}

This multi-resolution setup exposes the VideoGenerator to diverse spatial layouts and instruction complexities, improving aspect-ratio robustness without losing the motion priors.

\subsection{CriticAgent Implementation Details}
\label{app:critic_agent}
We provide a comprehensive implementation for the CriticAgent, including its architecture, two-stage training protocol (SFT and RM), dataset construction strategies, and closed-loop inference logic.

\vspace{-1em}
\begin{table}[htbp]
    \centering
    \small
    \renewcommand{\arraystretch}{1.05}
    \setlength{\tabcolsep}{4pt}
    \begin{threeparttable}
        \caption{\textbf{Hyperparameter settings for CriticAgent training}. Parameters are differentiated by the two training phases: Supervised Fine-Tuning (SFT) and Pairwise Reward Modeling (RM).}
        \label{tab:critic_hyperparameters}
        \begin{tabularx}{0.95\textwidth}{@{}lccX@{}}
            \toprule
            \textbf{Parameter} & \textbf{SFT} & \textbf{RM} & \textbf{Description} \\
            \midrule
            \rowcolor{tabcolor!20}
            \multicolumn{4}{@{}l}{\textbf{\textit{Optimization \& Training Strategy}}} \\
            Base model & \multicolumn{2}{c}{Qwen3-VL-8B-Instruct} & Backbone vision-language model \\
            Learning rate & $1.0 \times 10^{-4}$ & $5.0 \times 10^{-6}$ & Peak learning rate for AdamW \\
            LR scheduler & Cosine & Cosine & Decay schedule type \\
            Warmup ratio & 0.1 & 0.1 & Ratio of warmup steps \\
            Max epochs & 3.0 & 1.0 & Number of training epochs \\
            Batch size/GPU & 1 & 1 & Per-device batch size \\
            Gradient accumulation & 8 & 8 & Steps for effective batch construction \\
            Precision & bf16 & bf16 & Mixed-precision training format \\
            Max sequence length & 8192 & 8192 & Maximum context length in tokens \\
            \midrule
            \rowcolor{tabcolor!20}
            \multicolumn{4}{@{}l}{\textbf{\textit{Model Architecture \& LoRA Configuration}}} \\
            LoRA rank ($r$) & 16 & 16 & Rank dimension for adaptation \\
            LoRA alpha ($\alpha$) & 32 & 32 & Scaling factor ($2 \times r$) \\
            LoRA dropout & 0.05 & 0.05 & Dropout probability for LoRA layers \\
            Target modules & \multicolumn{2}{l}{all linear} & Attention and MLP layers applied \\
            \midrule
            \rowcolor{tabcolor!20}
            \multicolumn{4}{@{}l}{\textbf{\textit{RM-Specific Configuration}}} \\
            Loss function & - & Bradley-Terry & Pairwise ranking loss objective \\
            Margin & - & 0.0 & Margin for ranking separation \\
            \bottomrule
        \end{tabularx}
    \end{threeparttable}
\end{table}

\noindent\textbf{Architecture and Training Configuration.}
Analogous to the PlanAgent, the CriticAgent is initialized using the Qwen3-VL-8B~\cite{Qwen3-VL} backbone. To enable fine-grained video evaluation capabilities while preserving generalization, we apply Low-Rank Adaptation (LoRA) across all linear layers.

The training regimen proceeds in two sequential stages. First, \textbf{Supervised Fine-Tuning (SFT)} conditions the model to act as a rigorous judge by internalizing explicit evaluation dimensions, such as action adherence, object interaction, temporal coherence, and physical realism. Second, \textbf{Pairwise Reward Modeling (RM)} improves discriminative precision through a Bradley-Terry objective, aligning the model's scores with human or oracle preference rankings. Both stages use the AdamW optimizer with a cosine learning rate scheduler; detailed hyperparameters are summarized in Table~\ref{tab:critic_hyperparameters}.

\noindent\textbf{Reward Data Construction.}
We employ a hybrid data construction strategy to balance reasoning depth with fine-grained discriminative sensitivity.

\begin{itemize}[leftmargin=1.25em, itemsep=0pt, topsep=0pt]
    \item \textbf{SFT Data.} We implement a teacher-student distillation pipeline based on the VideoVerse benchmark~\cite{VideoVerse}. Following its protocol, we synthesize diverse video samples using state-of-the-art video generators, including CogVideoX-1.5-5B~\cite{yang2024cogvideox}, SkyReels-V2-14B~\cite{SkyReels}, HunyuanVideo~\cite{kong2024hunyuanvideo}, OpenSora2.0~\cite{peng2025opensora2}, Wan2.1-14B~\cite{Wan}, Wan2.2-A14B~\cite{Wan}, Hailuo~\cite{hailuo}, Veo3, and Sora-2. We use Gemini-3-Pro as the oracle judge to generate CoT critiques and scalar scores over five dimensions: \textit{Action Adherence}, \textit{Object Interaction}, \textit{Goal Achievement}, \textit{Temporal Coherence}, and \textit{Physical Realism}. These signals teach the CriticAgent a structured, evidence-based evaluation schema.

    \item \textbf{RM Data.} We leverage GAIA~\cite{GAIA} to strengthen action-level sensitivity through pairwise reward modeling. Preference pairs $(y_w, y_l)$ are constructed from two sources: \textit{Quality Ranking} pairs, derived from human ratings of videos depicting the same action; and \textit{Semantic Negative} pairs, formed by pairing a valid video with a semantically similar but incorrect action description (e.g., ``sipping'' vs. ``drinking''). This contrastive setup encourages the CriticAgent to inspect fine-grained motion evidence rather than relying on coarse scene context.
\end{itemize}

\noindent\textbf{Inference Logic.}
The CriticAgent provides actionable feedback inside the closed-loop pipeline through three operations:

\begin{itemize}[leftmargin=1.25em, itemsep=0pt, topsep=0pt]
    \item \textbf{Multi-Dimensional Scoring.} Given a generated segment $v_t$ and its corresponding plan step $s_t$, the agent evaluates action-video alignment across predefined dimensions. The output is constrained to a JSON object with scalar scores $r \in [0, 1]$ and textual evidence justifying each judgment.

    \item \textbf{Thresholding and Feedback.} We use the primary metric, \textit{Action Adherence}, to trigger correction. If its score falls below the success threshold $\tau=0.7$, the CriticAgent produces a \textit{revised action instruction} that explicitly addresses missing or violated conditions, such as ensuring that the hand clearly grasps the handle before lifting.

    \item \textbf{Streaming Evaluation.} For long-horizon tasks, evaluation is performed at the segment level. The agent inspects the clip corresponding to the current step $v_t$ rather than the full accumulated video, keeping the critique focused on the immediate action dynamics specified by $s_t$.
\end{itemize}

\noindent\textbf{System Prompts.}
We provide the exact system prompts used to steer the CriticAgent for evaluation, designed to enforce the structured output format and specific evidence-based reasoning requirements.

\vspace{-.5em}
\definecolor{CriticAgentBack}{HTML}{F1FBF1}
\definecolor{CriticAgentFrame}{HTML}{A8DFA7}
\definecolor{CriticAgentText}{HTML}{159A2F}
\definecolor{CriticAgentAccent}{HTML}{00B83A}

\tcbset{
  criticagentboxstyle/.style={
    enhanced,
    boxrule=0.8pt,
    arc=3pt,
    boxsep=2pt,
    left=5pt,
    right=5pt,
    top=4pt,
    bottom=4pt,
    colback=CriticAgentBack,
    fonttitle=\bfseries\color{CriticAgentText},
    title=System Prompt,
    toptitle=1mm,
    bottomtitle=1mm,
    coltitle=CriticAgentText,
    colframe=CriticAgentFrame,
    colbacktitle=CriticAgentFrame!45,
    fontupper=\footnotesize\linespread{0.95}\selectfont,
    before upper={\setlist[itemize]{itemsep=0pt, parsep=0pt, topsep=0pt, partopsep=0pt, leftmargin=10pt}},
    breakable
  }
}

\tcbset{
  criticagentjsonbox/.style={
    enhanced,
    colframe=CriticAgentFrame,
    colback=white,
    colbacktitle=CriticAgentFrame!35,
    coltitle=CriticAgentText,
    fonttitle=\small\bfseries\color{CriticAgentText},
    boxrule=0.8pt,
    arc=3pt,
    left=4pt,
    right=4pt,
    top=4pt,
    bottom=4pt,
    boxsep=5pt,
    listing only,
    listing options={
      language=json,
      basicstyle=\footnotesize\ttfamily,
      keywordstyle=\color{CriticAgentAccent},
      stringstyle=\color{CriticAgentText},
      commentstyle=\color{gray},
      morekeywords={true,false,null},
      literate=%
        *{,}{{\textcolor{CriticAgentAccent}{,}}}{1}%
         {:}{{\textcolor{CriticAgentAccent}{:}}}{1}%
         {\{}{{\textcolor{CriticAgentAccent}{\{}}}{1}%
         {\}}{{\textcolor{CriticAgentAccent}{\}}}}{1}%
    },
    breakable,
  }
}

\lstdefinelanguage{json}{
    basicstyle=\ttfamily\footnotesize,
    showstringspaces=false,
    breaklines=true,
    breakatwhitespace=true,
    morestring=[b]",
    stringstyle=\color{CriticAgentText},
    morecomment=[s]{/*}{*/},
    commentstyle=\color{gray},
    morekeywords={true,false,null},
    keywordstyle=\color{CriticAgentAccent},
    literate=%
      *{,}{{\textcolor{CriticAgentAccent}{,}}}{1}%
       {:}{{\textcolor{CriticAgentAccent}{:}}}{1}%
       {\{}{{\textcolor{CriticAgentAccent}{\{}}}{1}%
       {\}}{{\textcolor{CriticAgentAccent}{\}}}}{1}%
}

\begin{itemize}[leftmargin=1.25em]
  \item \textbf{Multi-Dimensional Scoring Prompt.} This prompt performs a rigorous evaluation of the video across five key dimensions.

\begin{tcolorbox}[criticagentboxstyle, title=System Prompt for CriticAgent (Multi-Dimensional Scoring), width=0.90\textwidth, center]

You are a meticulous \textbf{\textcolor{CriticAgentAccent}{Video Critic}}.
Given a generated video and its task specification, your role is to evaluate the video across multiple dimensions using a structured, evidence-based methodology.

\vspace{1ex}
\noindent\textbf{\textcolor{CriticAgentAccent}{Input Data:}}
\begin{itemize}[leftmargin=10pt]
  \item \textbf{\textcolor{CriticAgentText}{global\_goal}}: High-level textual description of the objective.
  \item \textbf{\textcolor{CriticAgentText}{action\_plan\_list}}: Ordered steps including \texttt{text}, \texttt{actions}, \texttt{pre}, and \texttt{post} conditions.
  \item \textbf{\textcolor{CriticAgentText}{Video}}: The AI-generated video frames.
\end{itemize}

\vspace{1ex}
\noindent\textbf{\textcolor{CriticAgentAccent}{Evaluation Dimensions:}}
\begin{itemize}[leftmargin=10pt]
  \item \textbf{A. Action Adherence}: Deduct points for missing or swapped steps.
  \item \textbf{B. Object Interaction}: Verify correct verb-tool-object usage.
  \item \textbf{C. Goal Achievement}: Check if \texttt{post} conditions are met.
  \item \textbf{D. Temporal Coherence}: Evaluate motion continuity and smoothness.
  \item \textbf{E. Visual \& Physics Realism}: Assess lighting, gravity, and collisions.
\end{itemize}

\vspace{1ex}
\begin{tcolorbox}[criticagentjsonbox, title=Expected Output Structure (JSON):]
\vspace{-3mm}
\begin{lstlisting}[language=json, basicstyle=\scriptsize\ttfamily, frame=none]
{
  "scores": {
    "action_adherence": {
      "score": 0.85, 
      "reason": "Concise reason..."
    },
    "object_interaction": {
      "reason": "General assessment...",
      "per_action": [
        {
          "verb": "pour", "tool": "kettle", "match": "yes",
          "score": 1.0, "reason": "Water flows naturally."
        }
      ]
    },
    "goal_achievement": {
      "reason": "General assessment...",
      "per_event": [
        {"event_id": 1, "score": 1.0, "reason": "Cup is full."}
      ]
    },
    "temporal_coherence": {"score": 0.9, "reason": "..."},
    "visual_physics_realism": {"score": 0.8, "reason": "..."}
  }
}
\end{lstlisting}
\vspace{-3mm}
\end{tcolorbox}

\end{tcolorbox}
\end{itemize}

\subsection{Inference-Time Closed-Loop Pipeline}
\label{app:closed_loop}

\begin{figure*}[t]
    \centering
    \includegraphics[width=0.98\linewidth]{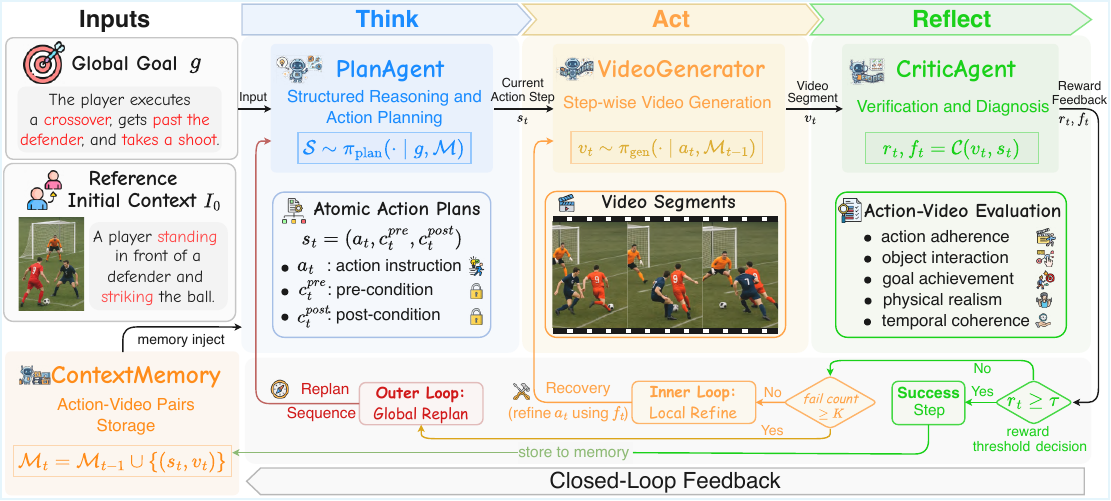}
    \caption{\textbf{Closed-Loop Think-Act-Reflect Pipeline.} PlanAgent decomposes the goal and reference context into atomic action plans, VideoGenerator executes each step into video segments, and CriticAgent verifies action-video alignment. Successful steps are stored in ContextMemory, while critic feedback triggers \emph{local refinement} or \emph{global replanning} for \textbf{closed-loop inference-time correction}.}
    \label{fig:framework}
    \vspace{-1em}
\end{figure*}

Figure~\ref{fig:framework} illustrates the \textbf{inference-time closed-loop pipeline} used by SPIRAL. Given a global goal and a reference initial state, the system iteratively plans, generates, verifies, and corrects video segments until the goal is completed. The pipeline proceeds as follows:

\begin{itemize}[leftmargin=1.25em, itemsep=0pt, topsep=0pt]
    \item \textbf{Goal and Reference Initialization.} The system receives the textual goal $g$ and the reference initial state, which define the target task and the starting scene configuration.

    \item \textbf{PlanAgent Planning.} The PlanAgent $\pi_{\text{plan}}$ observes the goal and current context memory $\mathcal{M}_{t-1}$, then decomposes the task into structured atomic steps $\mathcal{S}=\{s_1,\dots,s_T\}$. Each step specifies an action instruction together with pre- and post-conditions that make the plan executable.

    \item \textbf{VideoGenerator Execution.} For the current step $s_t$, the VideoGenerator $\pi_{\text{gen}}$ synthesizes a video segment $v_t$ conditioned on the action instruction and the accumulated visual context. This produces a local execution attempt for the planned action.

    \item \textbf{CriticAgent Verification.} The CriticAgent $\mathcal{C}$ evaluates whether $v_t$ faithfully executes $s_t$, checking action adherence, object interaction, goal achievement, temporal coherence, and physical realism. It outputs a reward score $r_t$ and textual feedback $f_t$ that diagnoses any observed failure.

    \item \textbf{Feedback for Local Refinement.} If the score falls below the success threshold but the plan remains feasible, the feedback is used to locally refine the current action instruction. The VideoGenerator then regenerates the same step with a more explicit or corrected instruction, such as emphasizing an unmet precondition or post-condition.

    \item \textbf{Feedback for Global Replanning.} If repeated local refinement fails, the error is treated as a planning-level issue. The CriticAgent feedback is passed back to the PlanAgent, which replans from the failed step $s_t$ onward while preserving the successful prefix stored in ContextMemory.
\end{itemize}

After a step is verified as successful, its action-video pair is appended to ContextMemory, $\mathcal{M}_t=\mathcal{M}_{t-1}\cup\{(s_t,v_t)\}$. The system then repeats the Think-Act-Reflect loop for subsequent steps, applying local refinement or global replanning whenever necessary, until the complete goal is achieved.

\subsection{Training-Time GRPO-based Self-Evolving}
\label{app:grpo}

\begin{figure*}[t]
    \centering
    \includegraphics[width=0.98\linewidth]{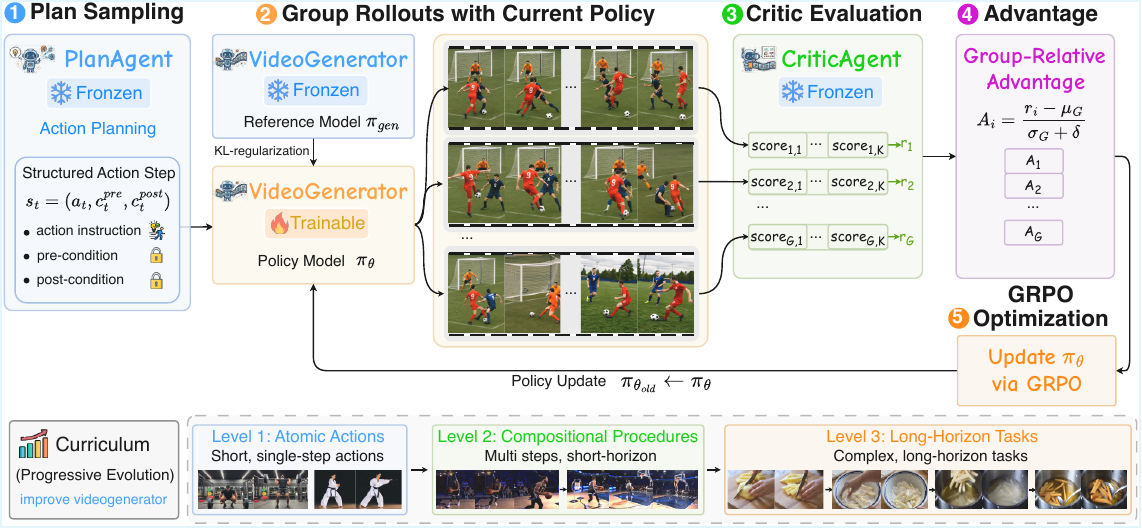}
    \caption{\textbf{Training-Time Self-Evolving via GRPO.} PlanAgent samples action steps, VideoGenerator produces group rollouts, and CriticAgent assigns rewards for GRPO-based policy optimization. A \emph{curriculum} gradually increases task complexity from atomic actions to long-horizon procedures, enabling VideoGenerator to \emph{internalize} planning and verification signals for \textbf{progressive-evolution}.}
    \label{fig:grpo}
    \vspace{-1em}
\end{figure*}

Figure~\ref{fig:grpo} summarizes the \textbf{training-time self-evolving procedure}. While inference-time feedback can correct individual failures, our goal is to internalize such corrections into the VideoGenerator itself. We therefore optimize the supervised VideoGenerator $\pi_{\text{gen}}$ with Group Relative Policy Optimization (GRPO), using PlanAgent-generated action steps as tasks and CriticAgent scores as rewards.

The training loop follows a closed-loop think-act-reflect structure:

\begin{itemize}[leftmargin=1.25em, itemsep=0pt, topsep=0pt]
    \item \textbf{Plan Sampling:} PlanAgent samples a structured action sequence $\mathcal{S}$ from a training goal $g$.
    \item \textbf{Group Rollouts:} For each action step $s_t$, the current VideoGenerator policy samples a group of $G$ candidate video segments $\{v_{t,i}\}_{i=1}^{G}$.
    \item \textbf{Critic Evaluation:} CriticAgent evaluates rewards $\{r_i\}_{i=1}^{G}$ based on action-video alignment.
    \item \textbf{GRPO Optimization:} GRPO normalizes rewards within the group, computes relative advantages, and updates the VideoGenerator while regularizing it toward the supervised initialization.
\end{itemize}

Below, we provide the mathematical details of this procedure. We first cast the denoising process inside VideoGenerator as a Markov Decision Process (MDP), then introduce stochastic exploration via SDE sampling, and finally define the GRPO objective used to update the VideoGenerator.

\noindent\textbf{Video Generation as a Markov Decision Process.}
While the high-level framework operates on plan steps $t \in \{1, \dots, T\}$, the low-level generation of a single video segment $v_t$ involves an iterative denoising process. We formulate this inner loop as a finite-horizon MDP defined by the tuple $\mathcal{M}_{\text{denoise}} = \langle \mathbf{S}, \mathbf{A}, P, R \rangle$:

\begin{itemize}[leftmargin=1.25em, itemsep=0pt, topsep=0pt]
    \item \textbf{State Space ($\mathbf{S}$):} A state is defined as $\mathbf{x}_\tau = (\mathbf{c}_t, \tau, \mathbf{z}_\tau)$, where $\tau$ represents the current denoising timestep (from $K$ down to $0$), $\mathbf{z}_\tau$ denotes the noisy video latent, and $\mathbf{c}_t = (s_t, \text{Encoder}(\mathcal{M}_{t-1}))$ encodes the atomic plan $s_t$ and memory context $\mathcal{M}_{t-1}$ at the $t$-th step.
    \item \textbf{Action Space ($\mathbf{A}$):} The action $\mathbf{a}_\tau$ corresponds to the VideoGenerator's prediction at step $\tau$. For a Rectified Flow model, $\mathbf{a}_\tau = \pi_{\theta}(\mathbf{x}_\tau)$ predicts the velocity field $\mathbf{u}_\tau$, transitioning the state from $\mathbf{z}_\tau$ to $\mathbf{z}_{\tau-1}$.
    \item \textbf{Transition Probability ($P$):} The transition $P(\mathbf{x}_{\tau-1} \mid \mathbf{x}_\tau, \mathbf{a}_\tau)$ is governed by the sampling solver (defined via SDEs below).
    \item \textbf{Reward Function ($R$):} We employ a sparse reward structure where signals are provided only at the terminal state $\tau=0$ (the clean video $v_t$). The reward is determined by the CriticAgent $\mathcal{C}$:
    \begin{equation}
        R(\mathbf{x}_\tau) = 
        \begin{cases}
             \mathcal{C}(v_t, s_t) & \text{if } \tau = 0 \\
            0 & \text{otherwise}
        \end{cases}
    \end{equation}
\end{itemize}

\noindent\textbf{Stochastic Exploration via SDEs.}
Standard sampling in Rectified Flow models follows an Ordinary Differential Equation (ODE): $\mathrm{d}\mathbf{z} = \mathbf{u}_t \mathrm{d}\tau$. However, GRPO requires stochastic exploration to generate diverse trajectories $\{v_{t,i}\}_{i=1}^G$ for relative advantage estimation. We therefore follow DanceGRPO~\cite{xue2025dancegrpo} and adopt a reverse-time \textbf{Stochastic Differential Equation (SDE)} formulation.

\newcommand{\grpoalgstep}[1]{%
    \begingroup
    \setlength{\fboxsep}{1.5pt}%
    \colorbox{tabcolor!20}{#1}%
    \endgroup
}

\begin{algorithm}[t]
    \centering
    \caption{Training-Time Self-Evolving via Closed-Loop GRPO}
    \label{algo:grpo}
    \small
        \begin{algorithmic}[1]
        \REQUIRE Supervised VideoGenerator $\pi_{\text{gen}}$, PlanAgent $\pi_{\text{plan}}$, CriticAgent $\mathcal{C}$, Dataset of goals $\mathcal{D}_g$
        \ENSURE Evolved VideoGenerator $\pi_{\theta}$ initialized from $\pi_{\text{gen}}$
        
        \FOR{iteration $= 1, \dots, M$}
            \STATE Sample global goal $g \sim \mathcal{D}_g$
            \STATE \textbf{\grpoalgstep{Think}:} Generate structured plan sequence $\mathcal{S} \sim \pi_{\text{plan}}(\cdot \mid g)$
            \STATE Initialize ContextMemory $\mathcal{M}_0 = \emptyset$
            \STATE Sync sampling policy $\pi_{\theta_{\text{old}}} \leftarrow \pi_{\theta}$
            
            \FOR{step $t = 1, \dots, T$} 
                \STATE \COMMENT{Iterate through plan steps}
                \STATE Let current atomic plan $s_t \in \mathcal{S}$
                \STATE \textbf{\grpoalgstep{Act}} (Shared Noise): Sample initialization $\mathbf{z}_{K} \sim \mathcal{N}(\mathbf{0}, \mathbf{I})$
                \STATE Generate $G$ video candidates $\{v_{t,i}\}_{i=1}^G$ via $\pi_{\theta_{\text{old}}}$ using SDE solver (Eq.~\ref{eq:reverse_sde})
                \STATE \quad $v_{t,i} \leftarrow \text{SolveSDE}(s_t, \mathcal{M}_{t-1}, \mathbf{z}_{K})$   \COMMENT{Conditioned on shared noise}
                    
                \STATE \textbf{\grpoalgstep{Reflect}:} Evaluate alignment via CriticAgent (Eq.~\ref{eq:critic}):
                \STATE \quad \quad $r_i \leftarrow \mathcal{C}(v_{t,i}, s_t)$ for $i=1 \dots G$
                
                \STATE \textbf{\grpoalgstep{Advantage}:} Compute $A_i$ using group statistics (Eq.~\ref{eq:advantage}):
                \STATE \quad \quad $A_{i} = \frac{r_{i} - \text{mean}(\{r_1, \dots, r_G\})}{\text{std}(\{r_1, \dots, r_G\}) + \delta}$
                
                \STATE \textbf{\grpoalgstep{Update}:} Maximize objective $\mathcal{J}(\theta)$ (Eq.~\ref{eq:grpo_objective}) via gradient ascent:
                \STATE \quad \quad $\mathcal{L}_{\text{surr}} = \frac{1}{G} \sum_{i=1}^{G} \min \Big( \rho_{t,i} A_i, \mathrm{clip}(\rho_{t,i}, 1-\epsilon, 1+\epsilon)\, A_i \Big)$
                \STATE \quad \quad $\theta \leftarrow \theta + \eta \nabla_\theta \left( \mathcal{L}_{\text{surr}} - \beta D_{\mathrm{KL}}(\pi_{\theta} \,\|\, \pi_{\mathrm{gen}}) \right)$
                
                \STATE Select best segment $v_t^* = \text{argmax}_{v_{t,i}} (r_i)$
                \STATE Update ContextMemory $\mathcal{M}_t \leftarrow \mathcal{M}_{t-1} \cup \{(s_t, v_t^*)\}$
            \ENDFOR
        \ENDFOR
        \end{algorithmic}
\end{algorithm}
\vspace{-0.2em}

Inspired by recent stochastic sampling theories, we introduce a diffusion term into the flow matching process. The reverse SDE for generation is given by:
\begin{equation}
    \mathrm{d}\mathbf{z}_\tau = \underbrace{\left(\mathbf{u}_\tau(\mathbf{z}_\tau) - \frac{1}{2}\eta_\tau^2 \nabla_{\mathbf{z}} \log p_\tau(\mathbf{z}_\tau)\right)}_{\text{Drift Term}} \mathrm{d}\tau + \underbrace{\eta_\tau \mathrm{d}\mathbf{w}}_{\text{Diffusion Term}}
    \label{eq:reverse_sde}
\end{equation}
where $\mathbf{u}_\tau$ is the velocity predicted by $\pi_\theta$, $\mathrm{d}\mathbf{w}$ denotes a standard Wiener process (Brownian motion), and $\eta_\tau$ is a time-dependent scalar controlling stochasticity. The score term $\nabla \log p_\tau(\mathbf{z}_\tau)$ guides the sample towards data distribution, derived from Gaussian assumption $p_\tau(\mathbf{z}_\tau) = \mathcal{N}(\mathbf{z}_\tau \mid \alpha_\tau \mathbf{x}, \sigma_\tau^2 \mathbf{I})$ as:
\begin{equation}
    \nabla \log p_\tau(\mathbf{z}_\tau) = -\frac{\mathbf{z}_\tau - \alpha_\tau \mathbf{x}_{\text{pred}}}{\sigma_\tau^2}
\end{equation}
By solving Eq.~\eqref{eq:reverse_sde} numerically, we sample a group of diverse candidates conditioned on the same action step $s_t$, satisfying the exploration requirement for GRPO.

\noindent\textbf{Shared Noise Initialization.}
A critical detail is the initialization of the starting noise $\mathbf{z}_K$. As demonstrated by~\cite{xue2025dancegrpo}, using independent noise for each group member leads to \emph{reward hacking}, where the model overfits to specific noise artifacts.
Therefore, we enforce a shared noise strategy:
\begin{equation}
    \forall i \in \{1, \dots, G\}: \quad \mathbf{z}_{K, i} = \mathbf{z}_{\text{shared}}, \quad \text{where } \mathbf{z}_{\text{shared}} \sim \mathcal{N}(\mathbf{0}, \mathbf{I})
\end{equation}
This ensures that the variance in rewards $r_i$ mainly reflects differences in stochastic sampling paths (Eq.~\eqref{eq:reverse_sde}) rather than unrelated initial noise, stabilizing the gradient estimate.

\noindent\textbf{Objective Function.}
We optimize the VideoGenerator policy $\pi_\theta$ using GRPO. For a plan step $s_t$, we sample a group of $G$ trajectories $\{v_{t,i}\}_{i=1}^G$ using the old policy $\pi_{\theta_{\text{old}}}$. CriticAgent assigns rewards $\{r_i\}_{i=1}^{G}$, and the advantages are computed via group-wise normalization:
\begin{equation}
    A_i = \frac{r_i - \mu_G}{\sigma_G + \delta}, \quad \text{where } \mu_G = \frac{1}{G}\sum_{j=1}^G r_j, \quad \sigma_G = \sqrt{\frac{1}{G}\sum_{j=1}^G (r_j - \mu_G)^2}
\end{equation}
The final objective function maximizes the surrogate gain with importance sampling clipping and KL-regularization:
\begin{equation}
\begin{aligned}
    \mathcal{J}(\theta) =
    \mathbb{E}_{\substack{
    s_t \sim \pi_{\text{plan}}(\cdot \mid g) \\
    v_{t,i} \sim \pi_{\theta_{\mathrm{old}}}(\cdot \mid s_t)
    }}
    \Bigg[
    \frac{1}{G} \sum_{i=1}^{G}
    \min \Big(
    \rho_{t,i} A_i,
    \mathrm{clip}(\rho_{t,i}, 1-\epsilon, 1+\epsilon)\, A_i
    \Big)
    - \beta\, D_{\mathrm{KL}}\!\left(\pi_{\theta} \,\|\, \pi_{\mathrm{gen}}\right)
    \Bigg]
\end{aligned}
\end{equation}
where $\rho_{t,i} = \frac{\pi_{\theta}(v_{t,i} \mid s_t)}{\pi_{\theta_{\text{old}}}(v_{t,i} \mid s_t)}$ is the importance sampling ratio. The KL term $D_{\mathrm{KL}}(\pi_{\theta} \| \pi_{\mathrm{gen}})$ prevents the updated policy from drifting excessively from the supervised VideoGenerator initialization.

\paragraph{Full Training Algorithm.}
The complete closed-loop GRPO procedure is summarized in Algorithm~\ref{algo:grpo}. It integrates PlanAgent's structured reasoning and CriticAgent's verification signals to iteratively refine the VideoGenerator, turning feedback that was previously used only at inference time into a persistent training signal.

\clearpage
\section{Additional Details of ActVideoGen-Dataset}
\label{app:dataset}

\subsection{Dataset Annotation and Verification}
\noindent\textbf{Annotation process.}
As introduced in Sec.~\ref{sec:dataset}, we convert each procedural video into a structured record $\langle g, \text{CoT}, \mathcal{S}, \mathcal{V} \rangle$, where $g$ is the global goal, $\mathcal{V}=\{v_t\}_{t=1}^{T}$ denotes ordered video segments, and $\mathcal{S}=\{s_t\}_{t=1}^{T}$ contains step-level annotations.
The construction follows a two-stage pipeline:

\begin{itemize}[leftmargin=1.25em, itemsep=0pt, topsep=0pt]
    \item \textbf{Stage 1: Step-wise action annotation:} For datasets with temporal step boundaries, each segment $v_t$ is clipped with short pre- and post-context and annotated by GLM-4.5V~\cite{glm-4.5}. The model outputs a valid JSON object containing an atomic instruction, canonical verb-object-tool actions, 1-3 visual preconditions and postconditions, and the preserved time span. For goal-level clips without sub-step boundaries, we first infer ordered sub-steps and temporal spans from the continuous video, then normalize them to the same schema. This stage yields step tuples $s_t=(a_t,c_t^{pre},c_t^{post})$ that encode both executable actions and physical state transitions.

    \item \textbf{Stage 2: CoT planning generation:} Given the refined goal and ordered step tuples, GPT-5.1 generates the global planning rationale. The prompt exposes the step instructions, actions, preconditions, and postconditions, requiring the model to explain step ordering, dependency relations, and the observable termination condition. Thus, each final annotation contains both grounded action-video pairs and a causal plan connecting them.
\end{itemize}

\noindent\textbf{Verification process.}
To ensure annotation quality, we validate the generated records through a two-step pipeline that combines model-based filtering with human verification:

\begin{itemize}[leftmargin=1.25em, itemsep=0pt, topsep=0pt]
    \item \textbf{Step 1: VLM-based filtering:} After annotation, we employ a VLM-based filtering pipeline to exclude low-quality samples. Qwen3-VL-235B~\cite{Qwen3-VL} is used as an independent verifier rather than the original GLM-4.5V annotator. For each candidate step, the verifier checks whether the main action is visible within the specified time span, whether the target objects/tools are correctly involved, whether the postconditions are visually supported, and whether the text introduces hallucinated content. Low-confidence or clearly mismatched samples are discarded.

    \item \textbf{Step 2: Human verification:} After VLM filtering, we ask three human annotators to independently verify each action-video pair. Annotators were shown the video segment, step instruction, action tuple, and pre-/postconditions. A pair is counted as \emph{fully aligned} only if the segment executes the annotated main action, uses the specified objects/tools when applicable, starts from states consistent with the preconditions, and reaches the stated postconditions without contradiction. Missing actions, wrong objects/tools, incorrect temporal localization, or unsupported postconditions are counted as misalignment, while minor wording differences that preserve the executable action and resulting state are allowed. We report pair-level alignment by \emph{majority vote} among the three annotators; under this criterion, about 93\% of the sampled pairs are fully aligned.
\end{itemize}

\subsection{Dataset Examples and Statistics}
\begin{figure*}[t]
    \centering
    \includegraphics[width=0.98\linewidth]{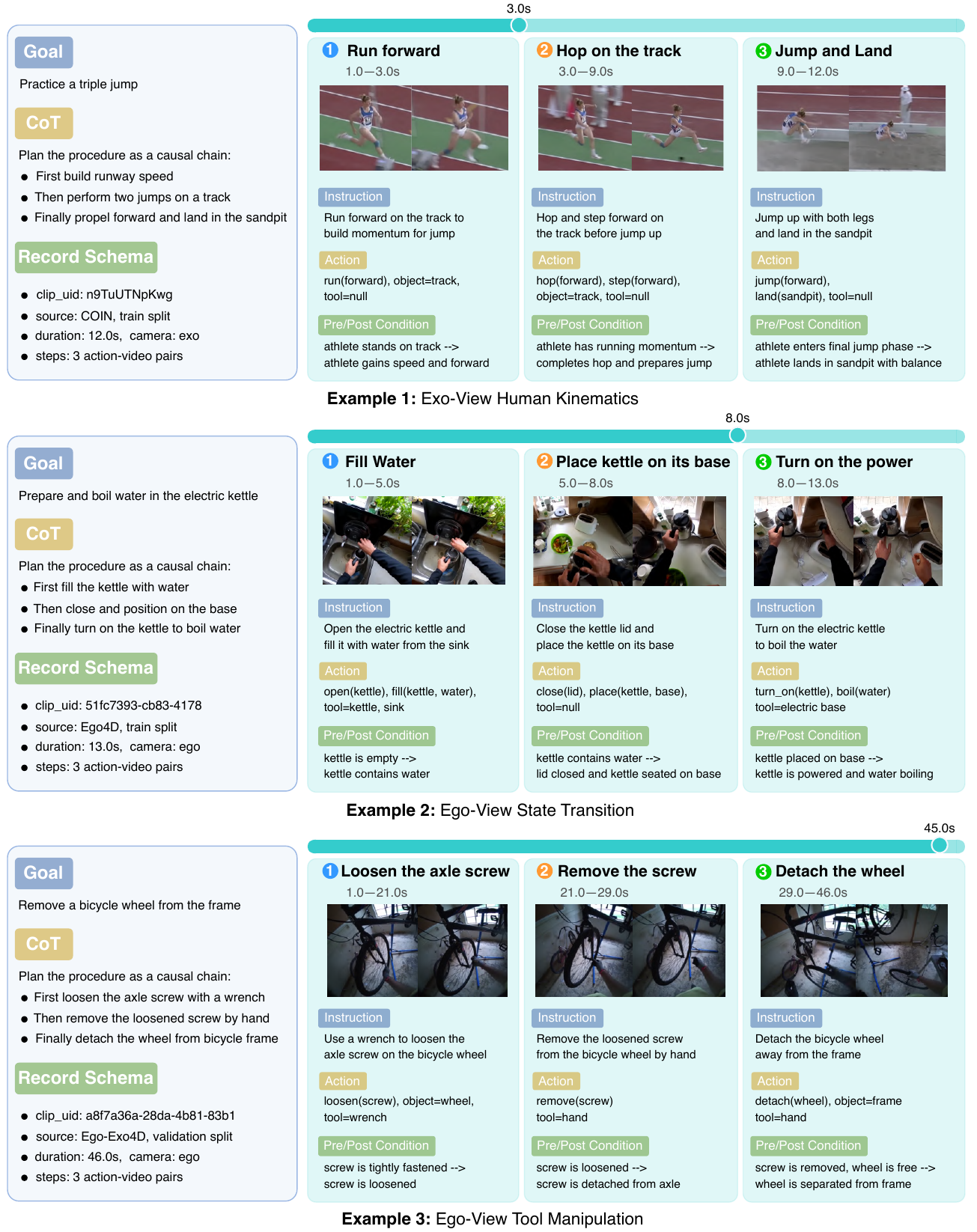}
    \caption{\textbf{Detailed annotation example from \ourDataset{}.} A procedural task is represented by a global goal and CoT planning rationale, then decomposed into temporally grounded action-video pairs. Each step contains an action instruction, an object-tool tuple, and pre-/post-conditions, providing structured supervision for long-horizon action planning and video generation.}
    \label{fig:dataset_example}
    \vspace{-1em}
\end{figure*}

\noindent\textbf{Dataset example.}
Fig.~\ref{fig:dataset_example} presents three representative annotation examples from \ourDataset{}, illustrating the breadth of procedural behaviors covered by our dataset:

\begin{itemize}[leftmargin=1.25em, itemsep=0pt, topsep=0pt]
    \item \textbf{Exo-view human kinematics (Example 1):} A triple-jump sequence is decomposed into running, hop-step motion, and final landing.
    \item \textbf{Ego-view state transition (Example 2):} Boiling water with an electric kettle requires filling the kettle, placing it on the base, and turning it on.
    \item \textbf{Ego-view tool manipulation (Example 3):} Removing a bicycle wheel involves loosening the screw with a wrench, removing the screw, and detaching the wheel.
\end{itemize}

Despite their different visual perspectives and action types, all examples share the same structured annotation format.
Each sample is organized around a global goal and a short CoT rationale, followed by temporally grounded action-video pairs.
For each step, we provide an executable action instruction, canonical action-object-tool tuple, and pre-/post-conditions, making the annotation directly usable for both planning supervision and step-wise video generation.

This format preserves the causal structure of procedural activities: earlier postconditions establish the visual and physical context needed by later actions.

\noindent\textbf{Dataset statistics.}
\ourDataset{} contains 24,616 procedural tasks and 118,156 step-level action-video pairs, with an average of approximately 4.8 steps per task.
In terms of source composition, it aggregates four resources: Ego4D~\cite{ego4d, ego4d-goal-step}, Ego-Exo4D~\cite{ego-exo4d}, COIN~\cite{coin}, and EPIC-KITCHENS~\cite{epickitchens}, covering both egocentric and exocentric perspectives.
For complexity, the dataset spans short atomic procedures as well as longer multi-step activities, with varied task durations, step counts, and action transition patterns.
For diversity, the collected tasks span a broad range of scenes and activity categories, including cooking, household manipulation, sports, and other real-world behaviors.
Together, these statistics show that \ourDataset{} provides diverse and causal supervision for training and evaluating structured planning and action-conditioned video generation.

\clearpage
\section{Additional Details of ActVideoGen-Bench}
\label{app:benchmark}

\subsection{Multi-Agent Evaluation Framework} 
As introduced in Sec.~\ref{sec:benchmark}, \ourBenchmark{} targets action quality beyond generic visual fidelity metrics. Since procedural videos require verifying ordered step execution, post-condition satisfaction, object interaction, and physical realism, single-turn QA evaluation is often insufficient. We therefore design a \textbf{multi-agent evaluation framework} based on the Chain-of-Query mechanism~\citep{han2025video}. The framework couples a VLM-based host evaluator with two LLM assistant auditors: the host grounds the evaluation in video frames and action data, the auditors raise complementary verification queries, and the final scoring module parses structured outputs into video-level metrics.

The evaluation follows a four-stage pipeline:
\begin{itemize}[leftmargin=1.25em, itemsep=0pt, topsep=0pt]
    \item \textbf{Stage 1: Video grounding.} A video description agent (VLM) observes the sampled frames and produces a chronological description grounded in the required action steps and post-conditions.
    \item \textbf{Stage 2: Query generation.} Two assistant agents (LLMs) generate complementary verification queries. The action-completeness auditor checks action execution and post-condition satisfaction, while the multi-dimensional quality auditor focuses on object interaction, motion smoothness, and physical fidelity.
    \item \textbf{Stage 3: Evidence-based answering.} An answer agent (VLM) revisits the video frames with the synthesized chain of queries and provides evidence-based answers, including step-wise completion judgments and quality-related observations.
    \item \textbf{Stage 4: Final scoring.} A score agent (VLM) integrates the initial description, query-answer history, original prompt, and action decomposition data to output structured metrics.
\end{itemize}
This collaborative pipeline enables cross-verified evaluation of both semantic action completion and fine-grained visual quality, providing robust action quality evaluation.

\subsection{Evaluation Metrics Definition}
The evaluator produces four action-quality metrics, denoted as \texttt{AC}, \texttt{AS}, \texttt{OI}, and \texttt{PF}. These metrics serve as the action verification criteria introduced in Sec.~\ref{sec:benchmark}, and support the video generation analyses in Sec.~\ref{sec:exp_videogen}. The final score agent outputs them in a structured \texttt{<METRICS>} dictionary, and all reported scores are normalized to $[0,1]$.

\begin{itemize}[leftmargin=1.25em, itemsep=0pt, topsep=0pt]
    \item \textbf{Action Completeness (AC).} Measures whether the generated video completes the required action sequence. A step is counted as complete only if the key action is visibly executed and all specified post-conditions are met. This is reported as the ratio $X/Y$, where $X$ represents the number of successfully completed steps and $Y$ is the total number of required steps.
    \item \textbf{Action Smoothness (AS).} Evaluates the temporal continuity of the generated actions. It accounts for transition quality between steps, motion continuity, inter-frame consistency, and the absence of abrupt cuts or jumps. The metric is evaluated on a 1–5 scale and normalized to $[0, 1]$ for reporting.
    \item \textbf{Object Interaction (OI).} Assesses the correctness and realism of interactions between the subject and any tools or objects. Key evaluation factors include contact quality, tool-use accuracy, object persistence, and the absence of clipping or floating artifacts. If no object interaction is involved, it is marked as \texttt{N/A}; otherwise, it is scored on a 1–5 scale and normalized to $[0, 1]$.
    \item \textbf{Physical Fidelity (PF).} Measures adherence to real-world physical constraints and natural motion patterns. It focuses on gravity, contact dynamics, and object deformation, penalizing artifacts like unnatural morphing or clipping. Scores are initially assigned on a 1–5 scale, ranging from severe physics violations to realistic body mechanics, and normalized to $[0, 1]$.
\end{itemize}

\subsection{Case Studies of Evaluation}
\begin{figure*}[t]
    \centering
    \includegraphics[width=0.98\linewidth]{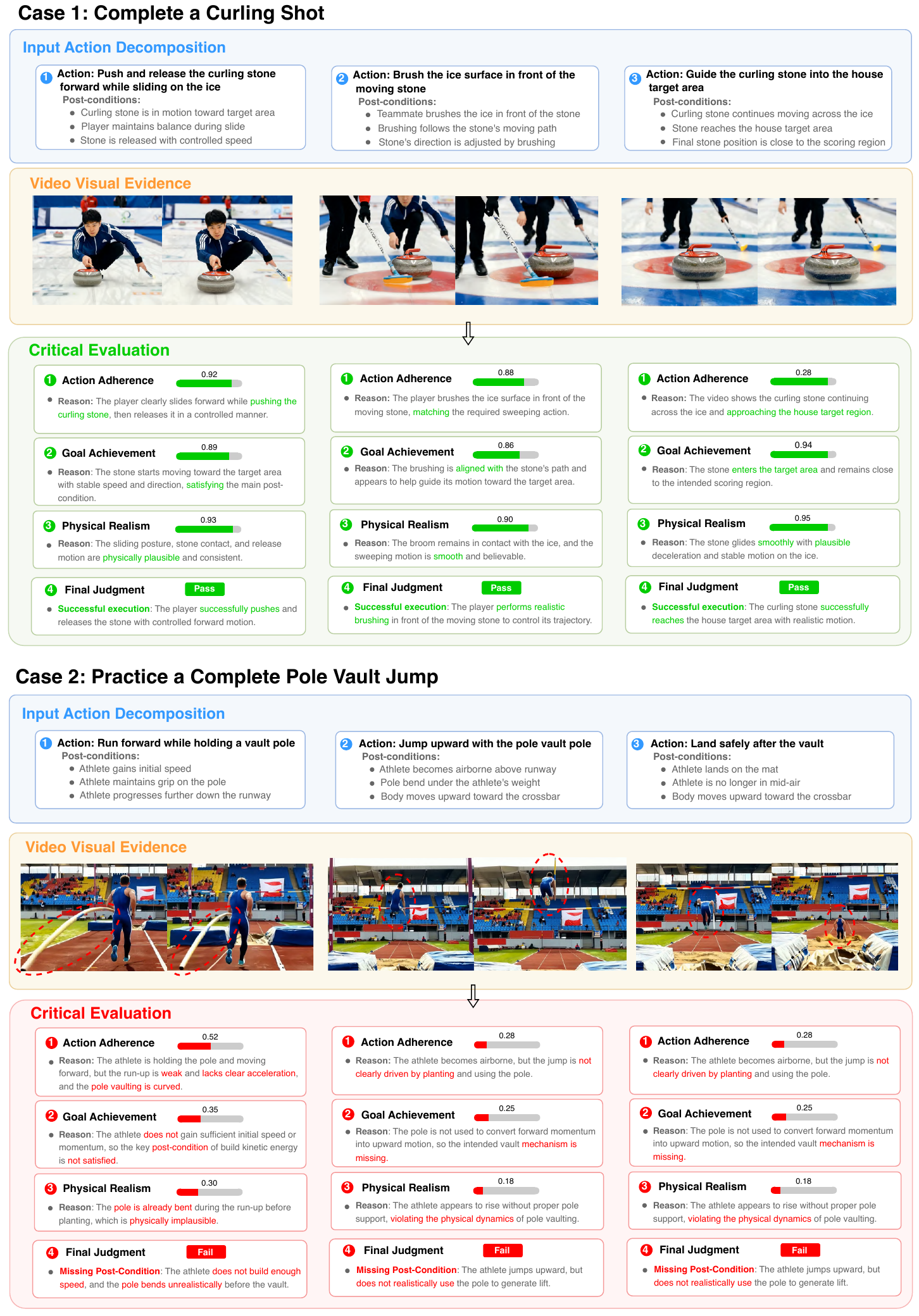}
    \caption{\textbf{Case studies of our multi-agent evaluator.}
    The evaluator grounds step-wise judgments in action decompositions and visual evidence, providing multi-dimensional assessments of action quality. Successful executions satisfy post-conditions, while failed actions reveal implausible motions.}
    \label{fig:vis_critic_case}
\end{figure*}

Fig.~\ref{fig:vis_critic_case} shows two representative cases, covering successful and failed action execution. Each case includes the action decomposition, visual evidence, step-wise evaluation, and final judgment.

These cases demonstrate that our evaluator provides interpretable step-level diagnoses rather than only coarse video-level scores. It assigns high scores when the visual evidence satisfies the required post-conditions and penalizes concrete failures, such as missing steps or unrealistic motion, providing robust action-quality metrics for our \ourBenchmark{}.

\clearpage
\section{Additional Quantitative Results} 
\label{app:additional_quantitative}

This section provides additional quantitative analyses. (i) Sec.~\ref{app:sft_generalization} evaluates whether SFT overfits VideoGenerator to planner-style prompts, (ii) Sec.~\ref{app:evaluator_stability} assesses the stability of our multi-agent evaluator, (iii) Sec.~\ref{app:inference_time} analyzes the inference cost and quality trade-off of our closed-loop pipeline.

\vspace{-1em}

\begin{table}[H]
    \centering
    \small
    \renewcommand{\arraystretch}{1.0}
    \setlength{\tabcolsep}{3pt}
    \begin{minipage}[t]{0.51\linewidth}
        \centering
        \caption{\textbf{VideoGenerator Generalization before and after SFT.} We evaluate the pre- and post-SFT SVI models~\cite{SVI} on official SVI general prompts.}
        \label{tab:sft_generalization}
        \resizebox{\linewidth}{!}{
            \begin{tabular}{l|ccccc}
                \toprule
                \textbf{Method} & \textbf{\begin{tabular}[c]{@{}c@{}}Aesthetic\\ Quality\end{tabular}} & \textbf{\begin{tabular}[c]{@{}c@{}}Imaging\\ Quality\end{tabular}} & \textbf{\begin{tabular}[c]{@{}c@{}}Subject\\ Consistency\end{tabular}} & \textbf{\begin{tabular}[c]{@{}c@{}}Motion\\ Smoothness\end{tabular}} & \textbf{\begin{tabular}[c]{@{}c@{}}Dynamic\\ Degree\end{tabular}} \\
                \midrule
                Pre-SFT (Base) & 63.84\% & 71.88\% & 98.13\% & 98.93\% & 17.61\% \\
                \rowcolor{tabcolor!25} Post-SFT (Ours) & 63.41\% & 73.07\% & 98.30\% & 99.05\% & 18.52\% \\
                $\Delta$ & -0.43\% & +1.19\% & +0.17\% & +0.12\% & +0.91\% \\
                \bottomrule
            \end{tabular}
        }
    \end{minipage}\hfill
    \begin{minipage}[t]{0.47\linewidth}
        \setlength{\tabcolsep}{3pt}
        \centering
        \caption{\textbf{Evaluator Consistency Across Independent Sessions.} We re-score 150 samples on action quality in two independent sessions.}
        \label{tab:evaluator_consistency}
        \resizebox{\linewidth}{!}{
            \begin{tabular}{l|cccc|c}
                \toprule
                \textbf{Metric} & \textbf{\begin{tabular}[c]{@{}c@{}}Action\\ Compl.\end{tabular}} & \textbf{\begin{tabular}[c]{@{}c@{}}Action\\ Smooth.\end{tabular}} & \textbf{\begin{tabular}[c]{@{}c@{}}Object\\ Interact.\end{tabular}} & \textbf{\begin{tabular}[c]{@{}c@{}}Physical\\ Fidelity\end{tabular}} & \textbf{\begin{tabular}[c]{@{}c@{}}Overall\\ Quality\end{tabular}} \\
                \midrule
                Agreement & 88.5\% & 76.2\% & 79.5\% & 82.0\% & \cellcolor{tabcolor!25}81.0\% \\
                Pearson $r$ & 0.812 & 0.685 & 0.764 & 0.792 & \cellcolor{tabcolor!25}0.753 \\
                \bottomrule
            \end{tabular}
        }
    \end{minipage}
    \vspace{-2em}
\end{table}

\begin{table}[H]
    \centering
    \small
    \renewcommand{\arraystretch}{1.0}
    \caption{\textbf{Inference Cost and Generation Quality Analysis.} We report average inference time, memory usage, and action quality. Experiments are conducted on the same NVIDIA A100 GPUs.}
    \label{tab:inference_time_analysis}
    \resizebox{\linewidth}{!}{
        \begin{tabular}{l|l|cc|ccccc}
            \toprule
            \multirow{2}{*}[-0.6em]{\textbf{Setting}} & \multirow{2}{*}[-0.6em]{\textbf{Model}} & \multicolumn{2}{c|}{\textbf{Inference Cost}} & \multicolumn{5}{c}{\textbf{Action Quality}} \\
            \cmidrule(lr){3-4} \cmidrule(lr){5-9}
             & & \textbf{\begin{tabular}[c]{@{}c@{}}Avg. Time\\ (s)\end{tabular}} & \textbf{\begin{tabular}[c]{@{}c@{}}Memory\\ (GB)\end{tabular}} & \textbf{\begin{tabular}[c]{@{}c@{}}Action\\ Completeness\end{tabular}} & \textbf{\begin{tabular}[c]{@{}c@{}}Action\\ Smoothness\end{tabular}} & \textbf{\begin{tabular}[c]{@{}c@{}}Object\\ Interaction\end{tabular}} & \textbf{\begin{tabular}[c]{@{}c@{}}Physical\\ Fidelity\end{tabular}} & \textbf{\begin{tabular}[c]{@{}c@{}}Overall\\ Quality\end{tabular}} \\
            \midrule
            \rowcolor{gray!10} Single-Shot Baseline & LongLive-2B~\cite{LongLive} & 8 & 28 & 0.273 & 0.646 & 0.776 & 0.808 & 0.626 \\
            \midrule
            PlanAgent (Ours) & Qwen3-VL-8B~\cite{Qwen3-VL} & 4 & 19 & - & - & - & - & - \\
            VideoGenerator (Per-step) & LongLive-2B~\cite{LongLive} & 8 & 28 & - & - & - & - & - \\
            CriticAgent (Per-step) & Qwen3-VL-8B~\cite{Qwen3-VL} & 7 & 22 & - & - & - & - & - \\
            \midrule
            \rowcolor{tabcolor!25} \ourMethodstr{} (T=3, Retry=1) & Plan+Gen+Critic & 49 & 69 & 0.721$\mathrm{\scriptstyle{\hi{+44.8\%}}}$ & \textbf{0.902}$\mathrm{\scriptstyle{\hi{+25.6\%}}}$ & 0.860$\mathrm{\scriptstyle{\hi{+8.4\%}}}$ & 0.924$\mathrm{\scriptstyle{\hi{+11.6\%}}}$ & 0.852$\mathrm{\scriptstyle{\hi{+22.6\%}}}$ \\
            \rowcolor{tabcolor!25} \ourMethodstr{} (T=3, Retry=2) & Plan+Gen+Critic & 64 & 69 & \textbf{0.786}$\mathrm{\scriptstyle{\hi{+51.3\%}}}$ & 0.862$\mathrm{\scriptstyle{\hi{+21.6\%}}}$ & \textbf{0.872}$\mathrm{\scriptstyle{\hi{+9.6\%}}}$ & \textbf{0.930}$\mathrm{\scriptstyle{\hi{+12.2\%}}}$ & \textbf{0.867}$\mathrm{\scriptstyle{\hi{+24.1\%}}}$ \\
            \bottomrule
        \end{tabular}
    }
    \vspace{-0.6em}
\end{table}

\subsection{VideoGenerator Generalization}
\label{app:sft_generalization}
We examine whether SFT on our \ourDataset{} causes the VideoGenerator to overfit to planner-generated prompts. To test this, we compare the \textbf{base SVI model}~\cite{SVI} with our \textbf{SFT-tuned model} using the official SVI evaluation prompts on general scenarios.

As shown in Table~\ref{tab:sft_generalization}, general video quality remains stable across all dimensions, with $|\Delta| < 1.2\%$. The post-SFT model slightly improves imaging quality and dynamic degree, while showing only a minor decrease in aesthetic quality. These results suggest that LoRA-based SFT preserves general generation capability while specializing the VideoGenerator for action-conditioned generation.

\subsection{Evaluator Stability}
\label{app:evaluator_stability}
As detailed in Appendix~\ref{app:benchmark}, we use a \textbf{multi-agent evaluation framework} for action quality assessment. This design reduces single-inference variance by requiring multiple agents to ground their judgments in shared visual evidence. To test evaluator stability, we re-score 150 samples across two independent sessions and report agreement rates and Pearson correlations.

Table~\ref{tab:evaluator_consistency} shows that agreement rates range from 76.2\% to 88.5\%, while Pearson correlations range from 0.685 to 0.812. These results indicate that the multi-agent evaluator is reasonably stable for \ourBenchmark{}, especially for fine-grained action-conditioned video assessment.

\subsection{Inference-Time Cost}
\label{app:inference_time}
\ourMethodstr{} incurs additional inference cost compared with a single-shot baseline, as it performs explicit planning, step-wise video generation, and critic-based verification. We compare both inference cost and generation quality between the \textbf{open-loop baseline} and our \textbf{closed-loop framework}.

Table~\ref{tab:inference_time_analysis} shows that \ourMethodstr{} requires more computation than single-shot generation. However, the overhead is predictable and bounded: runtime scales mainly with the number of action steps and retry rounds, while memory usage is dominated by the largest concurrently loaded components. For example, increasing the retry budget from 1 to 2 raises runtime from 49s to 64s, due to one additional generation-and-critic phase, while leaving memory usage unchanged at 69GB.

In return, action completeness improves from 0.273 to 0.786 (+51.3\%) and overall quality improves from 0.626 to 0.867 (+24.1\%). The added runtime is spent on the plan-generate-critic loop, which provides explicit correction and verification mechanisms that directly support these quality gains.

\clearpage
\section{Additional Qualitative Results} 
\label{app:additional_qualitative}

This section provides additional qualitative results that complement the quantitative analyses in experiments~\ref{sec:exp}. We first visualize the complete inference-time pipeline of \ourMethod{} from goal decomposition to final video composition (Sec.~\ref{app:qual_full_pipeline}), then examine how closed-loop feedback refinement corrects failed intermediate steps (Sec.~\ref{app:qual_feedback_regenerate}). We further present long-horizon and ultra-long procedural generation results across diverse action modes and extended task chains (Sec.~\ref{app:qual_long_horizon}), compare open-loop single-shot generation with our closed-loop framework (Sec.~\ref{app:qual_single_shot}), and finally show qualitative evidence for GRPO-based self-evolution (Sec.~\ref{app:qual_grpo}).

\subsection{End-to-End Pipeline Trace}
\label{app:qual_full_pipeline}
\begin{figure*}[t]
    \centering
    \includegraphics[width=1.0\linewidth]{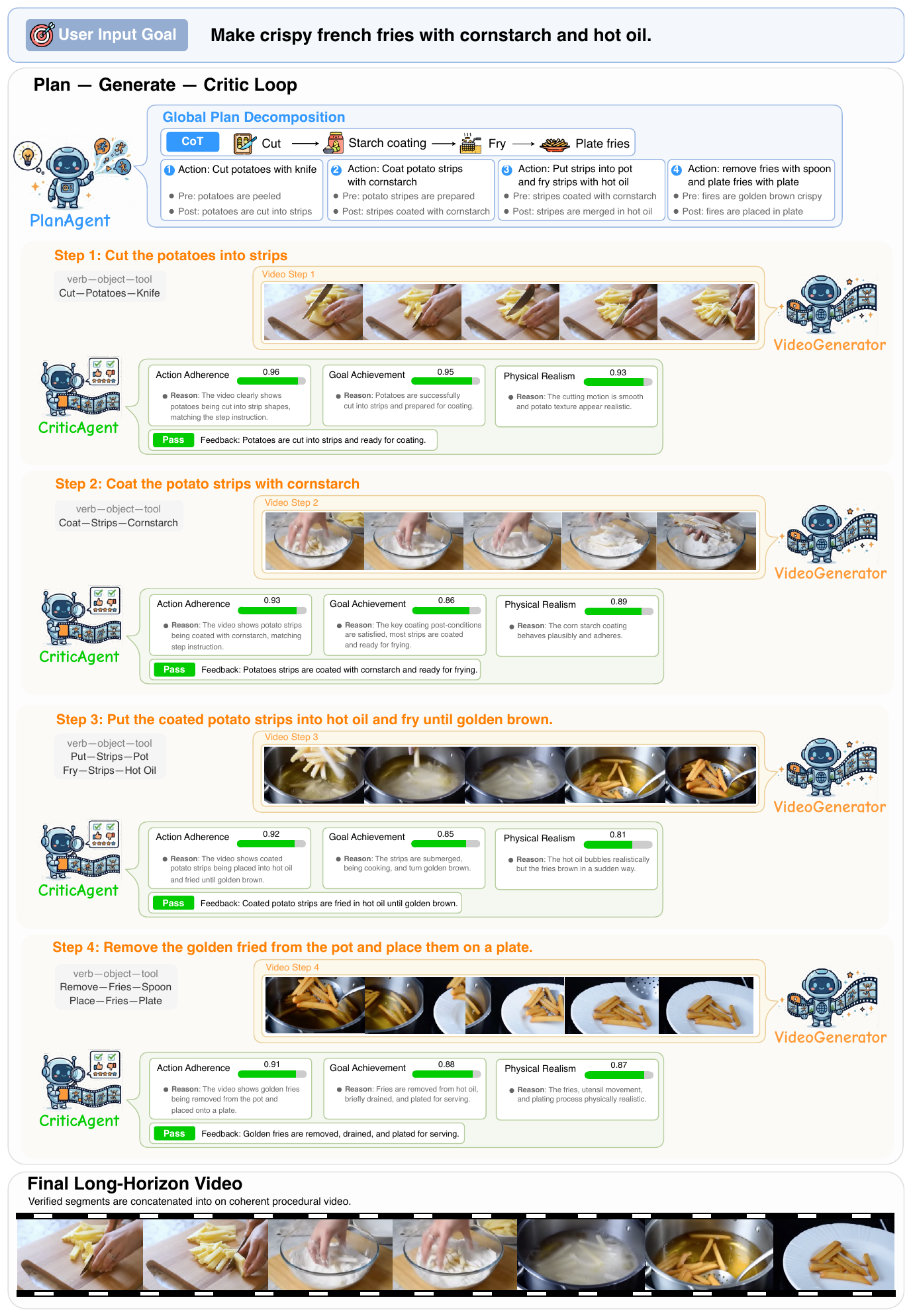}
    \vspace{-1em}
    \caption{\textbf{End-to-end pipeline of \ourMethod{}.} Given a user goal, PlanAgent first decomposes the task into step-wise actions with explicit pre-/post-conditions. At each step, VideoGenerator synthesizes a video segment, while CriticAgent evaluates action-video alignment and returns judgment with feedback. Verified segments are sequentially accumulated to compose the final long-horizon video.}
    \label{fig:vis_full_pipeline}
\end{figure*}

Figure~\ref{fig:vis_full_pipeline} illustrates a complete execution trace of \ourMethod{} for a long-horizon user goal, covering the full process from task decomposition to final video composition.

\begin{itemize}[leftmargin=1.25em, itemsep=0pt, topsep=0pt]
    \item \textbf{PlanAgent:} Starting from the high-level input instruction, PlanAgent decomposes the goal into a sequence of executable action steps with explicit preconditions and postconditions, providing a structured procedural plan for long-horizon generation.
    \item \textbf{VideoGenerator:} Given each planned step and the current visual context, VideoGenerator synthesizes a local video segment that executes the corresponding action while preserving continuity with previously verified segments.
    \item \textbf{CriticAgent:} After each generation step, CriticAgent evaluates whether the video segment satisfies the planned action and returns a pass/fail judgment with corrective feedback when failures occur.
\end{itemize}

This highlights the interpretability of the \textbf{plan-generate-critic loop}. Instead of producing a long video in one pass, \ourMethod{} exposes intermediate plans, per-step generations, critic judgments, and accumulated visual states. The verified segments are composed into a coherent, long-horizon video, providing a transparent mechanism for reliable procedural generation.

\subsection{Closed-Loop Feedback Refinement}
\label{app:qual_feedback_regenerate}
\begin{figure*}[t]
    \centering
    \includegraphics[width=1.0\linewidth]{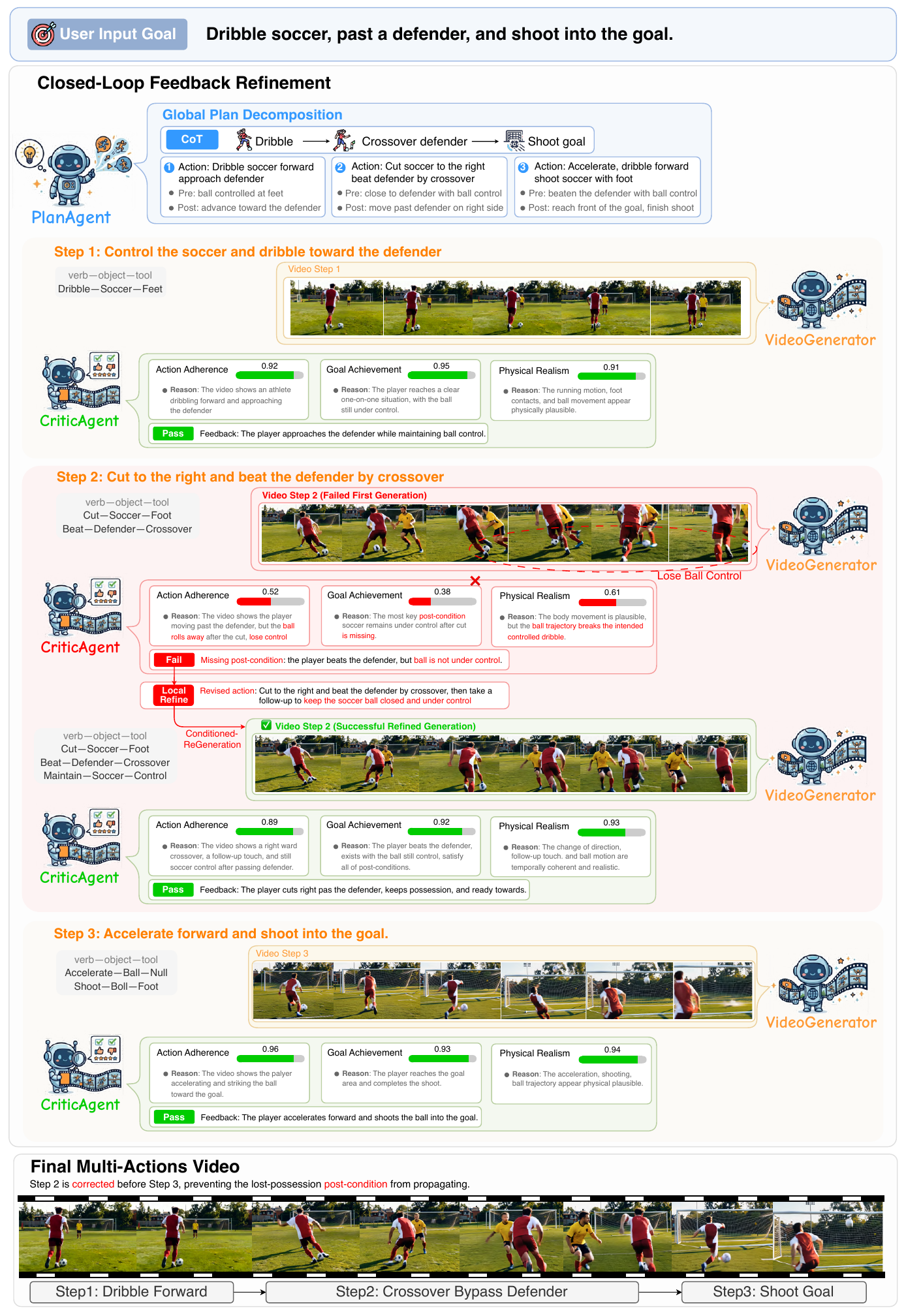}
    \vspace{-1em}
    \caption{\textbf{Closed-loop feedback refinement.} For the soccer procedure, CriticAgent detects that the Step~2 first generation bypasses the defender but loses ball control, violating the planned post-condition. \ourMethod{} uses this feedback to locally refine the action instruction, regenerate a successful crossover while preserving possession, and continue to final shooting step without propagating error.}
    \label{fig:vis_feedback_refine}
\end{figure*}

Figure~\ref{fig:vis_feedback_refine} visualizes the closed-loop feedback refinement process on a long-horizon soccer procedure, where the goal is decomposed into dribbling forward, bypassing defender, and shooting into the goal.

\begin{itemize}[leftmargin=1.25em, itemsep=0pt, topsep=0pt]
    \item \textbf{Failure Detection:} After each generated step, CriticAgent evaluates action adherence, goal achievement, and physical realism. In Step~2, the first generation loses ball control while bypassing the defender, violating the planned post-condition and causing the segment to fail.
    \item \textbf{Local Feedback Refinement:} Based on the critic feedback, \ourMethod{} explicitly refines the current instruction with the missing post-condition. The regenerated Step~2 corrects the failure while maintaining temporal continuity, allowing safely continue to the final shooting step.
\end{itemize}

Overall, this example shows how the closed-loop mechanism prevents local errors from accumulating across a long action chain. By detecting failures immediately and refining only the problematic step, \ourMethod{} preserves the successful prefix and composes a coherent final multi-action video.

\subsection{Long-Horizon Procedure Generation}
\label{app:qual_long_horizon}
\begin{figure*}[t]
    \centering
    \includegraphics[width=1.0\linewidth]{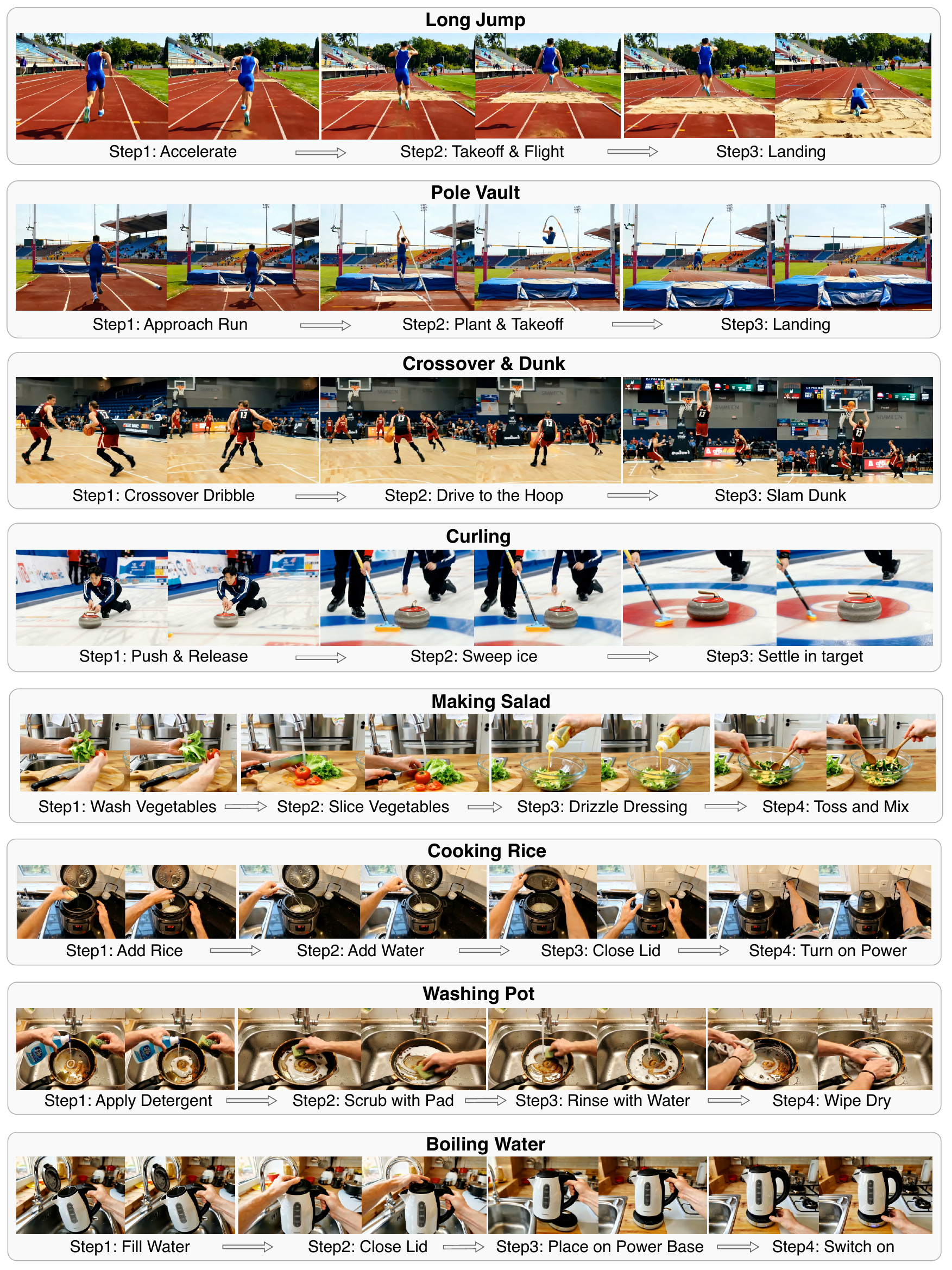}
    \vspace{-1em}
    \caption{\textbf{Procedural action video generation results.} We visualize two procedural generation modes: \emph{third-person human kinematics}, where actions control full-body motion (e.g., sports behaviors); and \emph{egocentric behavior generation}, where actions are executed from a first-person perspective.}
    \label{fig:vis_long_horizon}
\end{figure*}

\begin{figure*}[t]
    \centering
    \includegraphics[width=1.0\linewidth]{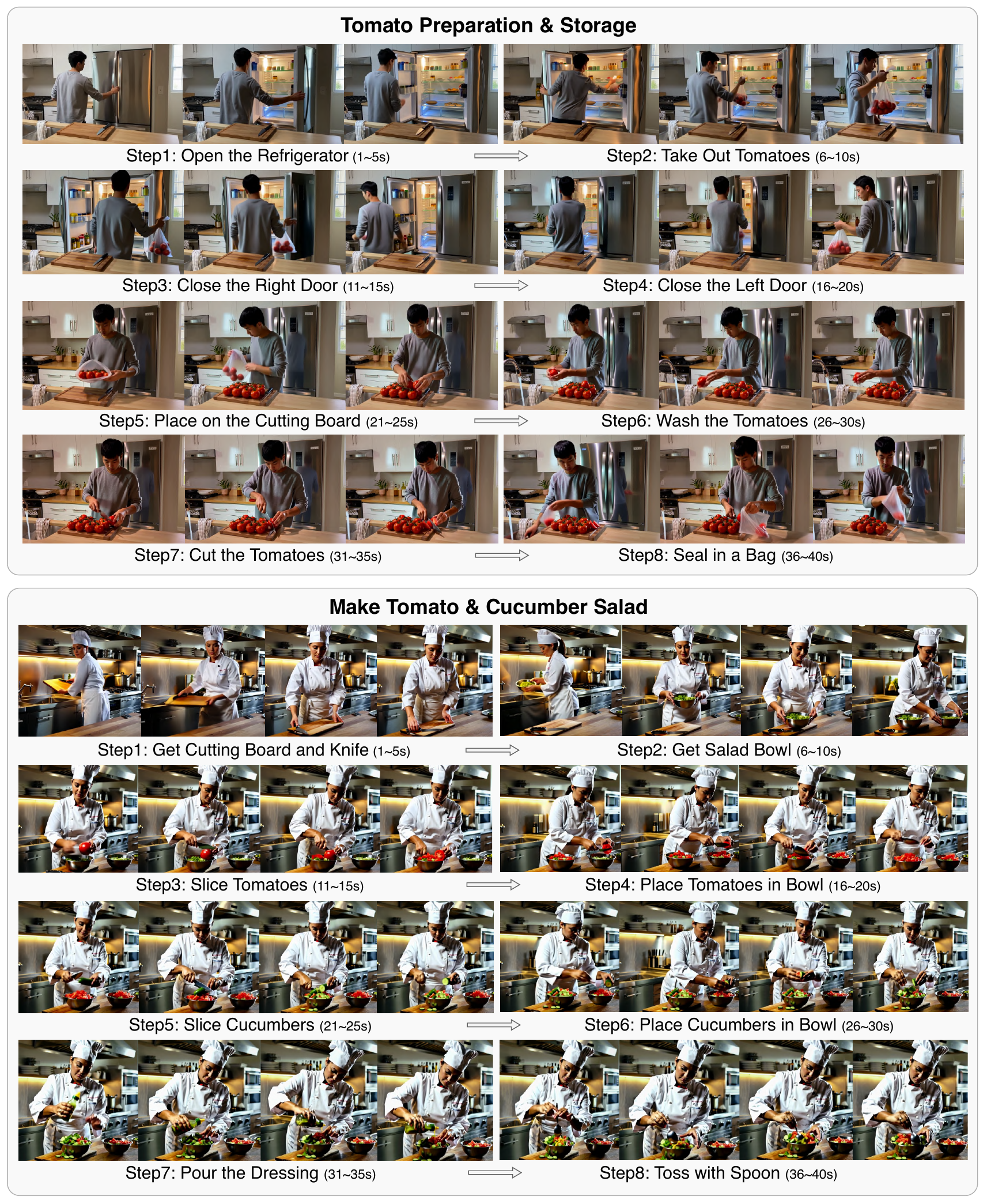}
    \vspace{-1em}
    \caption{\textbf{Ultra-long procedural generation results.} We visualize two eight-step kitchen procedures: tomato preparation and storage, and making tomato and cucumber salad. \ourMethod{} maintains coherent scene context, object states, and action ordering across extended multi-step manipulation.}
    \label{fig:vis_ultra_long_horizon}
\end{figure*}

Figure~\ref{fig:vis_long_horizon} and Figure~\ref{fig:vis_ultra_long_horizon} present long-horizon procedural generation results across diverse action types and extended task lengths.

\begin{itemize}[leftmargin=1.25em, itemsep=0pt, topsep=0pt]
    \item \textbf{Diverse Action Modes:} Figure~\ref{fig:vis_long_horizon} shows two representative procedural generation modes. Third-person human kinematics focuses on full-body motion control, such as sports behaviors, while egocentric behavior generation emphasizes first-person action execution and interaction intent.
    \item \textbf{Ultra-Long Procedural Chains:} Figure~\ref{fig:vis_ultra_long_horizon} further evaluates \ourMethod{} on ultra long-horizon kitchen procedures (>30s video duration). SPIRAL demonstrates strong preservation in scene layout, object states, and action dependencies over a substantially longer horizon.
\end{itemize}

Overall, these results demonstrate the importance of explicit step decomposition for long-horizon action-conditioned generation. By restricting each segment to a focused atomic action and maintaining context across steps, \ourMethod{} can translate abstract procedural goals into temporally ordered videos while preserving identity, viewpoint consistency, and procedural continuity.

\subsection{Comparison with Open-Loop Generation}
\label{app:qual_single_shot}
\begin{figure*}[t]
    \centering
    \includegraphics[width=1.0\linewidth]{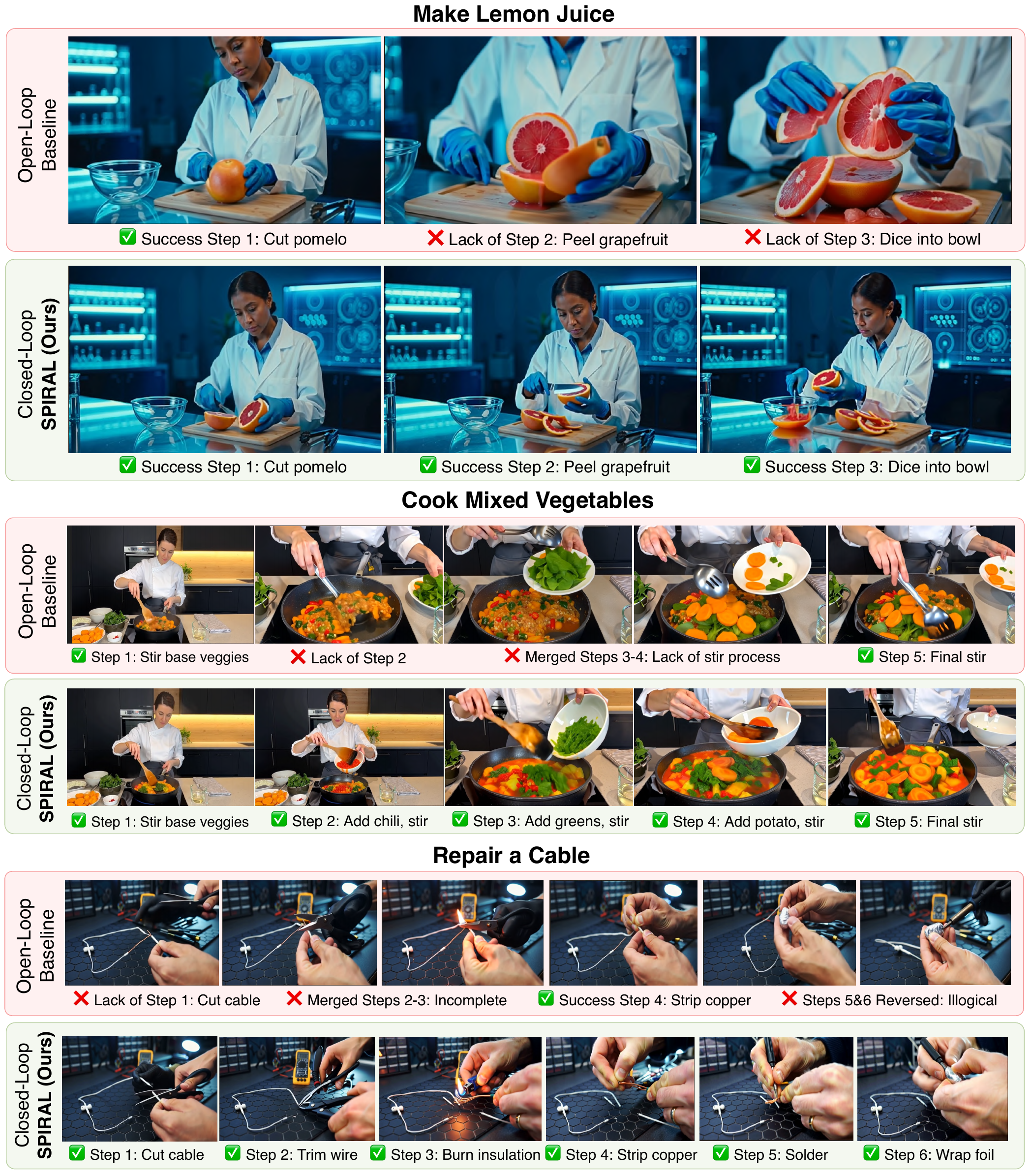}
    \vspace{-1em}
    \caption{\textbf{Single-shot open-loop vs. closed-loop generation on long-horizon procedural tasks.} The open-loop baseline often fails to complete the full procedure, exhibiting missing actions, incorrect ordering, and implausible transitions. In contrast, {\ourMethod{}} leverages explicit planning and verification, generating complete and long-chain action sequences with coherent ordering and procedural fidelity.}
    \label{fig:vis_compare_1}
\end{figure*}

\begin{figure*}[t]
    \centering
    \includegraphics[width=1.0\linewidth]{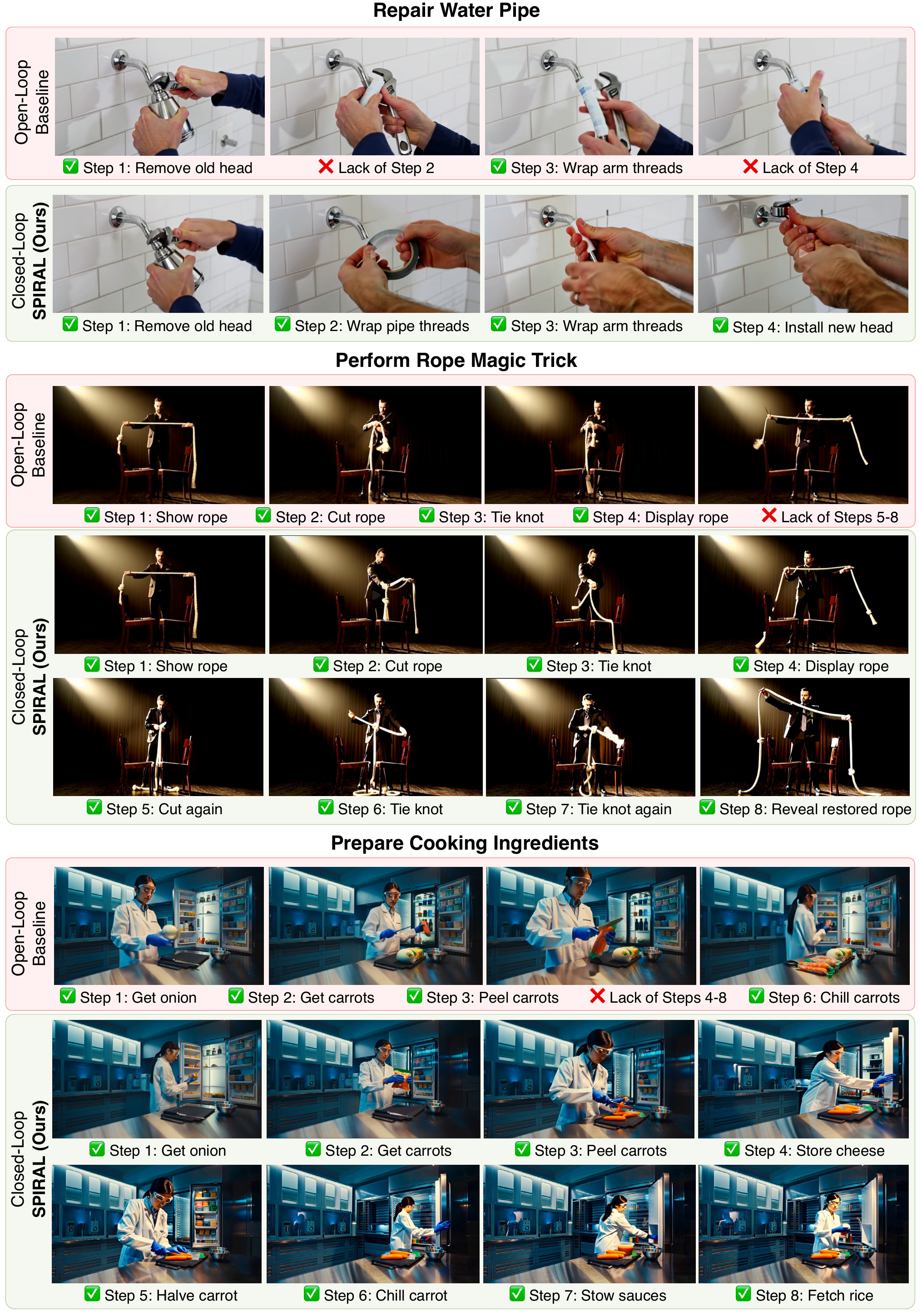}
    \vspace{-1em}
    \caption{\textbf{Single-shot open-loop vs. closed-loop generation on long-horizon procedural tasks.} The open-loop baseline often fails to complete the full procedure, exhibiting missing actions, incorrect ordering, and implausible transitions. In contrast, {\ourMethod{}} leverages explicit planning and verification, generating complete and long-chain action sequences with coherent ordering and procedural fidelity.}
    \label{fig:vis_compare_2}
\end{figure*}

We present qualitative comparisons in Fig.~\ref{fig:vis_compare_1} and Fig.~\ref{fig:vis_compare_2} across long-horizon procedural tasks to demonstrate the advantage of closed-loop generation.

\begin{itemize}[leftmargin=1.25em, itemsep=0pt, topsep=0pt]
    \item \textbf{Open-Loop Baseline:} The baseline must generate the entire procedure from a single prompt, requiring the model to implicitly infer all intermediate states and action dependencies at once. As shown in the comparisons, this often leads to missing actions, mixed actions, incorrect step ordering, and physically implausible transitions, especially when the target procedure contains a long chain of dependent actions.
    \item \textbf{Closed-Loop \ourMethodstr{}:} \ourMethod{} explicitly decomposes high-level goals into executable action steps and verifies the generated result after each step. This planning-and-verification loop prevents local failures from propagating, preserves procedural continuity, and enables more complete long-chain action generation with coherent temporal ordering and higher procedural fidelity.
\end{itemize}

Overall, these comparisons show that long-horizon procedural generation benefits from explicit closed-loop control. By replacing single-shot generation with step-wise planning and verification, \ourMethod{} improves both local action correctness and global procedural consistency.

\subsection{GRPO-Based Self-Evolution Visualization}
\label{app:qual_grpo}
\begin{figure*}[t]
    \centering
    \includegraphics[width=1.0\linewidth]{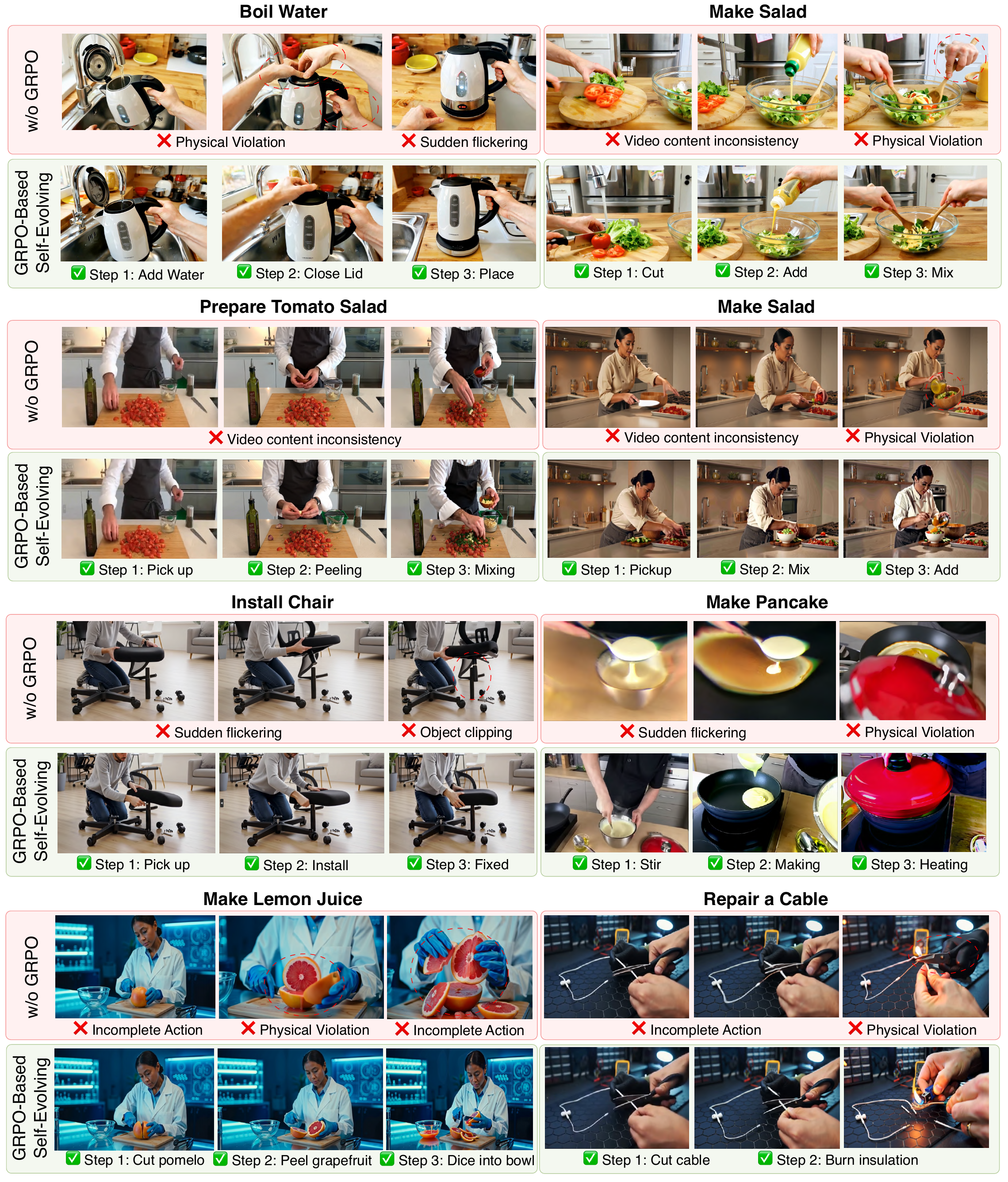}
    \vspace{-1em}
    \caption{\textbf{Effect of GRPO-based self-evolving.} Compared with the model without GRPO, GRPO-based self-evolving improves action completion, temporal coherence, and physical plausibility across long-horizon procedural generation by leveraging critic-derived reward signals during post-training.}
    \label{fig:vis_compare_3}
\end{figure*}

We further provide the comparison in Fig.~\ref{fig:vis_compare_3} to demonstrate the effect of GRPO-based self-evolving on long-horizon action generation.

\begin{itemize}[leftmargin=1.25em, itemsep=0pt, topsep=0pt]
    \item \textbf{Without Self-Evolving:} The generator can still suffer from incomplete actions, unstable motion, inconsistent object interactions, and physically implausible intermediate states. These failures indicate that supervised tuning alone may not fully internalize the action-following and verification signals required for reliable long-horizon execution.
    \item \textbf{GRPO-Based Self-Evolving:} By optimizing the generator with critic-derived rewards, GRPO encourages more complete action execution, smoother temporal transitions, and higher physical plausibility. As a result, the generator better follows procedural instructions and demonstrates stronger intrinsic long-horizon generation capability.
\end{itemize}

Overall, GRPO-based self-evolving complements inference-time correction by improving the generator itself. The resulting model better internalizes critical feedback, leading to more reliable action execution and smoother long-horizon procedural videos.

\clearpage
\section{Potential Societal Impact \& Limitations}
\label{sec:limitation}

\subsection{Societal Impact}
\label{sec:impact}
\ourMethodColored{} aims to improve controllable long-horizon video generation by explicit action planning, intermediate verification, and corrective feedback. Its societal impacts include both potential benefits and risks.
\begin{itemize}[leftmargin=1.25em, itemsep=0pt, topsep=0pt]
    \item \textbf{Positive impact.} \ourMethod{} can support content creation, education, and procedural simulation by synthesizing step-wise videos from high-level goals. Its explicit plan-generate-critic loop also provides interpretable traces, helping users diagnose action failures. It also encourages evaluation beyond generic visual fidelity toward action execution, object interaction, and physical consistency, bridging the gap toward more advanced action-conditioned video world models.
    \item \textbf{Negative impact.} More controllable and temporally coherent video generation may increase risks of misleading synthetic media, impersonation, or unsafe instructional content if deployed without safeguards. Models may also inherit social biases, cultural stereotypes, or privacy-sensitive patterns from large-scale video data. Practical deployment should therefore include prompt filtering, provenance or watermarking mechanisms, human review for sensitive domains, and restrictions on harmful or deceptive use cases.
\end{itemize}

\subsection{Known Limitations}
\label{sec:known_limitation}
While \ourMethodColored{} demonstrates strong performance in long-horizon action-conditioned video generation, it still has several limitations.
\begin{itemize}[leftmargin=1.25em, itemsep=0pt, topsep=0pt]
    \item \textbf{Inference cost.} The closed-loop think--act--reflect process introduces additional system complexity and inference latency compared with single-shot generation. A promising direction is to amortize planning and verification into the video generator itself.
    \item \textbf{Action representation.} The current formulation focuses on sequential, high-level semantic actions and does not explicitly model fine-grained control signals or tightly coupled physical interactions. Richer action representations and tighter low-level motion modeling remain open directions.
    \item \textbf{Automatic evaluation.} Our evaluation protocol relies on VLM-based agents. Although Appendix~\ref{app:evaluator_stability} shows reasonable consistency, automatic evaluation can still be affected by model bias, imperfect visual grounding, and ambiguity in open-ended procedural tasks.
    \item \textbf{Generalization scope.} Our experiments mainly focus on the distribution covered by \ourDataset{} and \ourBenchmark{}. Generalization to specialized domains, safety-critical procedures, rare object interactions, or culturally diverse activities remains to be systematically studied.
\end{itemize}
Addressing these limitations is essential for building more robust, efficient, and responsibly deployable long-horizon video generation models.

\clearpage
\section{Public Resources Used}
\label{sec:resources}

\subsection{Public Datasets Used}
\begin{itemize}
    \item Ego4D\footnote{\url{https://ego4d-data.org/}} \dotfill Ego4D Asset License Agreement
    
    \item Ego-Exo4D\footnote{\url{https://ego-exo4d-data.org/}} \dotfill Ego-Exo4D Asset License Agreement

    \item COIN\footnote{\url{https://coin-dataset.github.io/}} \dotfill COIN Asset License Agreement

    \item EPIC-KITCHENS\footnote{\url{https://epic-kitchens.github.io/}} \dotfill CC BY-NC 4.0

    \item VideoVerse\footnote{\url{https://github.com/Zeqing-Wang/VideoVerse}} \dotfill CC BY-NC 4.0

    \item GAIA\footnote{\url{https://github.com/zijianchen98/GAIA}} \dotfill Apache License 2.0
\end{itemize}

\subsection{Public Implementations Used}
\begin{itemize}
    \item Qwen3\footnote{\url{https://github.com/QwenLM/qwen3}} \dotfill Apache License 2.0
    
    \item Qwen3-VL\footnote{\url{https://github.com/QwenLM/Qwen3-VL}} \dotfill Apache License 2.0
    
    \item GLM-4.5V\footnote{\url{https://github.com/zai-org/GLM-V/}} \dotfill Apache License 2.0

    \item EgoPlan-Bench\footnote{\url{https://github.com/ChenYi99/EgoPlan}} \dotfill BSD 3-Clause License

    \item VideoGen-RewardBench\footnote{\url{https://github.com/KlingAIResearch/VideoAlign}} \dotfill MIT License

    \item VBench\footnote{\url{https://github.com/Vchitect/VBench}} \dotfill Apache License 2.0

    \item LLaMA-Factory\footnote{\url{https://github.com/hiyouga/LlamaFactory}} \dotfill Apache License 2.0

    \item LongLive\footnote{\url{https://github.com/NVlabs/LongLive}} \dotfill Apache License 2.0

    \item SVI\footnote{\url{https://github.com/vita-epfl/Stable-Video-Infinity}} \dotfill MIT License

    \item SANA-Video\footnote{\url{https://github.com/NVlabs/Sana}} \dotfill Apache License 2.0

    \item Wan2.1\footnote{\url{https://github.com/Wan-Video/Wan2.1}} \dotfill Apache License 2.0

    \item Wan2.2\footnote{\url{https://github.com/Wan-Video/Wan2.2}} \dotfill Apache License 2.0

    \item DanceGRPO\footnote{\url{https://github.com/XueZeyue/DanceGRPO}} \dotfill Apache License 2.0
\end{itemize}

\medskip
\clearpage
{
\small
\bibliographystyle{unsrt}  
\bibliography{ref}

@inproceedings{ego4d,
  title={Ego4d: Around the world in 3,000 hours of egocentric video},
  author={Grauman, Kristen and Westbury, Andrew and Byrne, Eugene and Chavis, Zachary and Furnari, Antonino and Girdhar, Rohit and Hamburger, Jackson and Jiang, Hao and Liu, Miao and Liu, Xingyu and others},
  booktitle={Proceedings of the IEEE/CVF conference on computer vision and pattern recognition},
  pages={18995--19012},
  year={2022}
}

@article{ego4d-goal-step,
  title={Ego4d goal-step: Toward hierarchical understanding of procedural activities},
  author={Song, Yale and Byrne, Eugene and Nagarajan, Tushar and Wang, Huiyu and Martin, Miguel and Torresani, Lorenzo},
  journal={Advances in Neural Information Processing Systems},
  volume={36},
  pages={38863--38886},
  year={2023}
}

@article{liao2025genie,
  title={Genie envisioner: A unified world foundation platform for robotic manipulation},
  author={Liao, Yue and Zhou, Pengfei and Huang, Siyuan and Yang, Donglin and Chen, Shengcong and Jiang, Yuxin and Hu, Yue and Cai, Jingbin and Liu, Si and Luo, Jianlan and others},
  journal={arXiv preprint arXiv:2508.05635},
  year={2025}
}

@inproceedings{ego-exo4d,
  title={Ego-exo4d: Understanding skilled human activity from first-and third-person perspectives},
  author={Grauman, Kristen and Westbury, Andrew and Torresani, Lorenzo and Kitani, Kris and Malik, Jitendra and Afouras, Triantafyllos and Ashutosh, Kumar and Baiyya, Vijay and Bansal, Siddhant and Boote, Bikram and others},
  booktitle={Proceedings of the IEEE/CVF Conference on Computer Vision and Pattern Recognition},
  pages={19383--19400},
  year={2024}
}

@inproceedings{coin,
  title={Coin: A large-scale dataset for comprehensive instructional video analysis},
  author={Tang, Yansong and Ding, Dajun and Rao, Yongming and Zheng, Yu and Zhang, Danyang and Zhao, Lili and Lu, Jiwen and Zhou, Jie},
  booktitle={Proceedings of the IEEE/CVF Conference on Computer Vision and Pattern Recognition},
  pages={1207--1216},
  year={2019}
}

@inproceedings{epickitchens,
  title={Scaling egocentric vision: The epic-kitchens dataset},
  author={Damen, Dima and Doughty, Hazel and Farinella, Giovanni Maria and Fidler, Sanja and Furnari, Antonino and Kazakos, Evangelos and Moltisanti, Davide and Munro, Jonathan and Perrett, Toby and Price, Will and others},
  booktitle={Proceedings of the European conference on computer vision (ECCV)},
  pages={720--736},
  year={2018}
}

@article{glm-4.5,
  title={Glm-4.5: Agentic, reasoning, and coding (arc) foundation models},
  author={Zeng, Aohan and Lv, Xin and Zheng, Qinkai and Hou, Zhenyu and Chen, Bin and Xie, Chengxing and Wang, Cunxiang and Yin, Da and Zeng, Hao and Zhang, Jiajie and others},
  journal={arXiv preprint arXiv:2508.06471},
  year={2025}
}

@article{Qwen3-VL,
      title={Qwen3-VL Technical Report}, 
      author={Shuai Bai and Yuxuan Cai and Ruizhe Chen and Keqin Chen and Xionghui Chen and Zesen Cheng and Lianghao Deng and Wei Ding and Chang Gao and Chunjiang Ge and Wenbin Ge and Zhifang Guo and Qidong Huang and Jie Huang and Fei Huang and Binyuan Hui and Shutong Jiang and Zhaohai Li and Mingsheng Li and Mei Li and Kaixin Li and Zicheng Lin and Junyang Lin and Xuejing Liu and Jiawei Liu and Chenglong Liu and Yang Liu and Dayiheng Liu and Shixuan Liu and Dunjie Lu and Ruilin Luo and Chenxu Lv and Rui Men and Lingchen Meng and Xuancheng Ren and Xingzhang Ren and Sibo Song and Yuchong Sun and Jun Tang and Jianhong Tu and Jianqiang Wan and Peng Wang and Pengfei Wang and Qiuyue Wang and Yuxuan Wang and Tianbao Xie and Yiheng Xu and Haiyang Xu and Jin Xu and Zhibo Yang and Mingkun Yang and Jianxin Yang and An Yang and Bowen Yu and Fei Zhang and Hang Zhang and Xi Zhang and Bo Zheng and Humen Zhong and Jingren Zhou and Fan Zhou and Jing Zhou and Yuanzhi Zhu and Ke Zhu},
	  journal={arXiv preprint arXiv:2511.21631},
      year={2025}
}

@article{LongLive,
  title={Longlive: Real-time interactive long video generation},
  author={Yang, Shuai and Huang, Wei and Chu, Ruihang and Xiao, Yicheng and Zhao, Yuyang and Wang, Xianbang and Li, Muyang and Xie, Enze and Chen, Yingcong and Lu, Yao and others},
  journal={arXiv preprint arXiv:2509.22622},
  year={2025}
}

@article{SVI,
  title={Stable video infinity: Infinite-length video generation with error recycling},
  author={Li, Wuyang and Pan, Wentao and Luan, Po-Chien and Gao, Yang and Alahi, Alexandre},
  journal={arXiv preprint arXiv:2510.09212},
  year={2025}
}

@article{Sora,
  title={Video generation models as world simulators},
  author={Brooks, Tim and Peebles, Bill and Holmes, Connor and DePue, Will and Guo, Yufei and Jing, Li and Schnurr, David and Taylor, Joe and Luhman, Troy and Luhman, Eric and others},
  journal={OpenAI Blog},
  volume={1},
  number={8},
  pages={1},
  year={2024}
}

@article{Kling,
  title={Kling-Omni Technical Report},
  author={Team, Kling and Chen, Jialu and Ci, Yuanzheng and Du, Xiangyu and Feng, Zipeng and Gai, Kun and Guo, Sainan and Han, Feng and He, Jingbin and He, Kang and others},
  journal={arXiv preprint arXiv:2512.16776},
  year={2025}
}

@article{Wan,
  title={Wan: Open and advanced large-scale video generative models},
  author={Wan, Team and Wang, Ang and Ai, Baole and Wen, Bin and Mao, Chaojie and Xie, Chen-Wei and Chen, Di and Yu, Feiwu and Zhao, Haiming and Yang, Jianxiao and others},
  journal={arXiv preprint arXiv:2503.20314},
  year={2025}
}

@article{SkyReels,
  title={Skyreels-v2: Infinite-length film generative model},
  author={Chen, Guibin and Lin, Dixuan and Yang, Jiangping and Lin, Chunze and Zhu, Junchen and Fan, Mingyuan and Zhang, Hao and Chen, Sheng and Chen, Zheng and Ma, Chengcheng and others},
  journal={arXiv preprint arXiv:2504.13074},
  year={2025}
}

@article{VideoReward,
  title={Improving video generation with human feedback},
  author={Liu, Jie and Liu, Gongye and Liang, Jiajun and Yuan, Ziyang and Liu, Xiaokun and Zheng, Mingwu and Wu, Xiele and Wang, Qiulin and Xia, Menghan and Wang, Xintao and others},
  journal={arXiv preprint arXiv:2501.13918},
  year={2025}
}

@inproceedings{VBench,
  title={Vbench: Comprehensive benchmark suite for video generative models},
  author={Huang, Ziqi and He, Yinan and Yu, Jiashuo and Zhang, Fan and Si, Chenyang and Jiang, Yuming and Zhang, Yuanhan and Wu, Tianxing and Jin, Qingyang and Chanpaisit, Nattapol and others},
  booktitle={Proceedings of the IEEE/CVF Conference on Computer Vision and Pattern Recognition},
  pages={21807--21818},
  year={2024}
}

@article{VideoVerse,
  title={VideoVerse: How Far is Your T2V Generator from a World Model?},
  author={Wang, Zeqing and Wei, Xinyu and Li, Bairui and Guo, Zhen and Zhang, Jinrui and Wei, Hongyang and Wang, Keze and Zhang, Lei},
  journal={arXiv preprint arXiv:2510.08398},
  year={2025}
}

@article{GAIA,
  title={Gaia: Rethinking action quality assessment for ai-generated videos},
  author={Chen, Zijian and Sun, Wei and Tian, Yuan and Jia, Jun and Zhang, Zicheng and Jiarui, Wang and Huang, Ru and Min, Xiongkuo and Zhai, Guangtao and Zhang, Wenjun},
  journal={Advances in Neural Information Processing Systems},
  volume={37},
  pages={40111--40144},
  year={2024}
}

@article{EgoPlan-Bench,
  title={Egoplan-bench: Benchmarking multimodal large language models for human-level planning},
  author={Chen, Yi and Ge, Yuying and Ge, Yixiao and Ding, Mingyu and Li, Bohao and Wang, Rui and Xu, Ruifeng and Shan, Ying and Liu, Xihui},
  journal={arXiv preprint arXiv:2312.06722},
  year={2023}
}

@article{zhao2025absolute,
  title={Absolute zero: Reinforced self-play reasoning with zero data},
  author={Zhao, Andrew and Wu, Yiran and Yue, Yang and Wu, Tong and Xu, Quentin and Lin, Matthieu and Wang, Shenzhi and Wu, Qingyun and Zheng, Zilong and Huang, Gao},
  journal={arXiv preprint arXiv:2505.03335},
  year={2025}
}

@article{huang2025r,
  title={R-zero: Self-evolving reasoning llm from zero data},
  author={Huang, Chengsong and Yu, Wenhao and Wang, Xiaoyang and Zhang, Hongming and Li, Zongxia and Li, Ruosen and Huang, Jiaxin and Mi, Haitao and Yu, Dong},
  journal={arXiv preprint arXiv:2508.05004},
  year={2025}
}

@article{wang2025vision,
  title={Vision-zero: Scalable vlm self-improvement via strategic gamified self-play},
  author={Wang, Qinsi and Liu, Bo and Zhou, Tianyi and Shi, Jing and Lin, Yueqian and Chen, Yiran and Li, Hai Helen and Wan, Kun and Zhao, Wentian},
  journal={arXiv preprint arXiv:2509.25541},
  year={2025}
}

@article{wu2025evolver,
  title={Evolver: Self-evolving llm agents through an experience-driven lifecycle},
  author={Wu, Rong and Wang, Xiaoman and Mei, Jianbiao and Cai, Pinlong and Fu, Daocheng and Yang, Cheng and Wen, Licheng and Yang, Xuemeng and Shen, Yufan and Wang, Yuxin and others},
  journal={arXiv preprint arXiv:2510.16079},
  year={2025}
}

@article{fu2025re,
  title={RE-Searcher: Robust Agentic Search with Goal-oriented Planning and Self-reflection},
  author={Fu, Daocheng and Mei, Jianbiao and Wen, Licheng and Yang, Xuemeng and Yang, Cheng and Wu, Rong and Hu, Tao and Li, Siqi and Shen, Yufan and Cai, Xinyu and others},
  journal={arXiv preprint arXiv:2509.26048},
  year={2025}
}

@inproceedings{soni2025videoagent,
  title={Videoagent: Self-improving video generation for embodied planning},
  author={Soni, Achint and Venkataraman, Sreyas and Chandra, Abhranil and Fischmeister, Sebastian and Liang, Percy and Dai, Bo and Yang, Sherry},
  booktitle={NeurIPS 2025 Workshop on Bridging Language, Agent, and World Models for Reasoning and Planning},
  year={2025}
}

@article{long2025vista,
  title={VISTA: A Test-Time Self-Improving Video Generation Agent},
  author={Long, Do Xuan and Wan, Xingchen and Nakhost, Hootan and Lee, Chen-Yu and Pfister, Tomas and Ar{\i}k, Sercan {\"O}},
  journal={arXiv preprint arXiv:2510.15831},
  year={2025}
}

@article{ha2018world,
  title={World models},
  author={Ha, David and Schmidhuber, J{\"u}rgen},
  journal={arXiv preprint arXiv:1803.10122},
  volume={2},
  number={3},
  year={2018}
}

@article{yang2025x,
  title={X-Scene: Large-Scale Driving Scene Generation with High Fidelity and Flexible Controllability},
  author={Yang, Yu and Liang, Alan and Mei, Jianbiao and Ma, Yukai and Liu, Yong and Lee, Gim Hee},
  journal={arXiv preprint arXiv:2506.13558},
  year={2025}
}

@article{mei2024dreamforge,
  title={Dreamforge: Motion-aware autoregressive video generation for multi-view driving scenes},
  author={Mei, Jianbiao and Hu, Tao and Yang, Xuemeng and Wen, Licheng and Yang, Yu and Wei, Tiantian and Ma, Yukai and Dou, Min and Shi, Botian and Liu, Yong},
  journal={arXiv preprint arXiv:2409.04003},
  year={2024}
}

@misc{hailuo,
  author = {MiniMax},
  title = {Hailuo {AI}},
  year = {2024},
  howpublished = {\url{https://hailuoai.video}},
}

@article{alonso2024diffusion,
  title={Diffusion for world modeling: Visual details matter in atari},
  author={Alonso, Eloi and Jelley, Adam and Micheli, Vincent and Kanervisto, Anssi and Storkey, Amos J and Pearce, Tim and Fleuret, Fran{\c{c}}ois},
  journal={Advances in Neural Information Processing Systems},
  volume={37},
  pages={58757--58791},
  year={2024}
}

@article{tang2025hunyuan,
  title={Hunyuan-GameCraft-2: Instruction-following Interactive Game World Model},
  author={Tang, Junshu and Liu, Jiacheng and Li, Jiaqi and Wu, Longhuang and Yang, Haoyu and Zhao, Penghao and Gong, Siruis and Yuan, Xiang and Shao, Shuai and Lu, Qinglin},
  journal={arXiv preprint arXiv:2511.23429},
  year={2025}
}

@article{agarwal2025cosmos,
  title={Cosmos world foundation model platform for physical ai},
  author={Agarwal, Niket and Ali, Arslan and Bala, Maciej and Balaji, Yogesh and Barker, Erik and Cai, Tiffany and Chattopadhyay, Prithvijit and Chen, Yongxin and Cui, Yin and Ding, Yifan and others},
  journal={arXiv preprint arXiv:2501.03575},
  year={2025}
}

@article{zhang2025world,
  title={World-in-world: World models in a closed-loop world},
  author={Zhang, Jiahan and Jiang, Muqing and Dai, Nanru and Lu, Taiming and Uzunoglu, Arda and Zhang, Shunchi and Wei, Yana and Wang, Jiahao and Patel, Vishal M and Liang, Paul Pu and others},
  journal={arXiv preprint arXiv:2510.18135},
  year={2025}
}

@article{yang2024cogvideox,
  title={Cogvideox: Text-to-video diffusion models with an expert transformer},
  author={Yang, Zhuoyi and Teng, Jiayan and Zheng, Wendi and Ding, Ming and Huang, Shiyu and Xu, Jiazheng and Yang, Yuanming and Hong, Wenyi and Zhang, Xiaohan and Feng, Guanyu and others},
  journal={arXiv preprint arXiv:2408.06072},
  year={2024}
}

@article{wu2025video,
  title={Video World Models with Long-term Spatial Memory},
  author={Wu, Tong and Yang, Shuai and Po, Ryan and Xu, Yinghao and Liu, Ziwei and Lin, Dahua and Wetzstein, Gordon},
  journal={arXiv preprint arXiv:2506.05284},
  year={2025}
}

@inproceedings{henschel2025streamingt2v,
  title={Streamingt2v: Consistent, dynamic, and extendable long video generation from text},
  author={Henschel, Roberto and Khachatryan, Levon and Poghosyan, Hayk and Hayrapetyan, Daniil and Tadevosyan, Vahram and Wang, Zhangyang and Navasardyan, Shant and Shi, Humphrey},
  booktitle={Proceedings of the Computer Vision and Pattern Recognition Conference},
  pages={2568--2577},
  year={2025}
}

@article{chen2025sana,
  title={Sana-video: Efficient video generation with block linear diffusion transformer},
  author={Chen, Junsong and Zhao, Yuyang and Yu, Jincheng and Chu, Ruihang and Chen, Junyu and Yang, Shuai and Wang, Xianbang and Pan, Yicheng and Zhou, Daquan and Ling, Huan and others},
  journal={arXiv preprint arXiv:2509.24695},
  year={2025}
}

@article{zeng2025coagent,
  title={CoAgent: Collaborative Planning and Consistency Agent for Coherent Video Generation},
  author={Zeng, Qinglin and Cai, Kaitong and Chen, Ruiqi and Lv, Qinhan and Wang, Keze},
  journal={arXiv preprint arXiv:2512.22536},
  year={2025}
}

@article{li2025editthinker,
  title={Editthinker: Unlocking iterative reasoning for any image editor},
  author={Li, Hongyu and Zhang, Manyuan and Zheng, Dian and Guo, Ziyu and Jia, Yimeng and Feng, Kaituo and Yu, Hao and Liu, Yexin and Feng, Yan and Pei, Peng and others},
  journal={arXiv preprint arXiv:2512.05965},
  year={2025}
}

@article{guo2025thinking,
  title={Thinking-while-Generating: Interleaving Textual Reasoning throughout Visual Generation},
  author={Guo, Ziyu and Zhang, Renrui and Li, Hongyu and Zhang, Manyuan and Chen, Xinyan and Wang, Sifan and Feng, Yan and Pei, Peng and Heng, Pheng-Ann},
  journal={arXiv preprint arXiv:2511.16671},
  year={2025}
}

@article{liao2025imagegen,
  title={Imagegen-cot: Enhancing text-to-image in-context learning with chain-of-thought reasoning},
  author={Liao, Jiaqi and Yang, Zhengyuan and Li, Linjie and Li, Dianqi and Lin, Kevin and Cheng, Yu and Wang, Lijuan},
  journal={arXiv preprint arXiv:2503.19312},
  year={2025}
}

@inproceedings{zhuo2025reflection,
  title={From reflection to perfection: Scaling inference-time optimization for text-to-image diffusion models via reflection tuning},
  author={Zhuo, Le and Zhao, Liangbing and Paul, Sayak and Liao, Yue and Zhang, Renrui and Xin, Yi and Gao, Peng and Elhoseiny, Mohamed and Li, Hongsheng},
  booktitle={Proceedings of the IEEE/CVF International Conference on Computer Vision},
  pages={15329--15339},
  year={2025}
}

@article{guo2025can,
  title={Can We Generate Images with CoT? Let's Verify and Reinforce Image Generation Step by Step},
  author={Guo, Ziyu and Zhang, Renrui and Tong, Chengzhuo and Zhao, Zhizheng and Huang, Rui and Zhang, Haoquan and Zhang, Manyuan and Liu, Jiaming and Zhang, Shanghang and Gao, Peng and others},
  journal={arXiv preprint arXiv:2501.13926},
  year={2025}
}

@article{VisionReward,
  title={Visionreward: Fine-grained multi-dimensional human preference learning for image and video generation},
  author={Xu, Jiazheng and Huang, Yu and Cheng, Jiale and Yang, Yuanming and Xu, Jiajun and Wang, Yuan and Duan, Wenbo and Yang, Shen and Jin, Qunlin and Li, Shurun and others},
  journal={arXiv preprint arXiv:2412.21059},
  year={2024}
}

@inproceedings{han2025video,
  title={Video-Bench: Human-Aligned Video Generation Benchmark},
  author={Han, Hui and Li, Siyuan and Chen, Jiaqi and Yuan, Yiwen and Wu, Yuling and Deng, Yufan and Leong, Chak Tou and Du, Hanwen and Fu, Junchen and Li, Youhua and others},
  booktitle={Proceedings of the Computer Vision and Pattern Recognition Conference},
  pages={18858--18868},
  year={2025}
}

@article{kong2024hunyuanvideo,
  title={Hunyuanvideo: A systematic framework for large video generative models},
  author={Kong, Weijie and Tian, Qi and Zhang, Zijian and Min, Rox and Dai, Zuozhuo and Zhou, Jin and Xiong, Jiangfeng and Li, Xin and Wu, Bo and Zhang, Jianwei and others},
  journal={arXiv preprint arXiv:2412.03603},
  year={2024}
}

@article{peng2025opensora2,
  title={Open-sora 2.0: Training a commercial-level video generation model in \$200 k},
  author={Peng, Xiangyu and Zheng, Zangwei and Shen, Chenhui and Young, Tom and Guo, Xinying and Wang, Binluo and Xu, Hang and Liu, Hongxin and Jiang, Mingyan and Li, Wenjun and others},
  journal={arXiv preprint arXiv:2503.09642},
  year={2025}
}

@article{xue2025dancegrpo,
  title={DanceGRPO: Unleashing GRPO on Visual Generation},
  author={Xue, Zeyue and Wu, Jie and Gao, Yu and Kong, Fangyuan and Zhu, Lingting and Chen, Mengzhao and Liu, Zhiheng and Liu, Wei and Guo, Qiushan and Huang, Weilin and others},
  journal={arXiv preprint arXiv:2505.07818},
  year={2025}
}

@article{park2025deepvideo,
  title={Deepvideo-r1: Video reinforcement fine-tuning via difficulty-aware regressive grpo},
  author={Park, Jinyoung and Na, Jeehye and Kim, Jinyoung and Kim, Hyunwoo J},
  journal={arXiv preprint arXiv:2506.07464},
  year={2025}
}

@article{chen2025exploring,
  title={Exploring the effect of reinforcement learning on video understanding: Insights from seed-bench-r1},
  author={Chen, Yi and Ge, Yuying and Wang, Rui and Ge, Yixiao and Qiu, Lu and Shan, Ying and Liu, Xihui},
  journal={arXiv preprint arXiv:2503.24376},
  year={2025}
}

@article{chen2025grpo,
  title={Grpo-care: Consistency-aware reinforcement learning for multimodal reasoning},
  author={Chen, Yi and Ge, Yuying and Wang, Rui and Ge, Yixiao and Cheng, Junhao and Shan, Ying and Liu, Xihui},
  journal={arXiv preprint arXiv:2506.16141},
  year={2025}
}

@article{ge2023making,
  title={Making llama see and draw with seed tokenizer},
  author={Ge, Yuying and Zhao, Sijie and Zeng, Ziyun and Ge, Yixiao and Li, Chen and Wang, Xintao and Shan, Ying},
  journal={arXiv preprint arXiv:2310.01218},
  year={2023}
}

@inproceedings{zhang2023videollama,
  title={Video-llama: An instruction-tuned audio-visual language model for video understanding},
  author={Zhang, Hang and Li, Xin and Bing, Lidong},
  booktitle={Proceedings of the 2023 conference on empirical methods in natural language processing: system demonstrations},
  pages={543--553},
  year={2023}
}

@inproceedings{he2024videoscore,
  title={Videoscore: Building automatic metrics to simulate fine-grained human feedback for video generation},
  author={He, Xuan and Jiang, Dongfu and Zhang, Ge and Ku, Max and Soni, Achint and Siu, Sherman and Chen, Haonan and Chandra, Abhranil and Jiang, Ziyan and Arulraj, Aaran and others},
  booktitle={Proceedings of the 2024 Conference on Empirical Methods in Natural Language Processing},
  pages={2105--2123},
  year={2024}
}

@article{he2025videoscore2,
  title={Videoscore2: Think before you score in generative video evaluation},
  author={He, Xuan and Jiang, Dongfu and Nie, Ping and Liu, Minghao and Jiang, Zhengxuan and Su, Mingyi and Ma, Wentao and Lin, Junru and Ye, Chun and Lu, Yi and others},
  journal={arXiv preprint arXiv:2509.22799},
  year={2025}
}

@article{wang2024lift,
  title={Lift: Leveraging human feedback for text-to-video model alignment},
  author={Wang, Yibin and Tan, Zhiyu and Wang, Junyan and Yang, Xiaomeng and Jin, Cheng and Li, Hao},
  journal={arXiv preprint arXiv:2412.04814},
  year={2024}
}

@article{wang2025unifiedreward,
  title={Unified reward model for multimodal understanding and generation},
  author={Wang, Yibin and Zang, Yuhang and Li, Hao and Jin, Cheng and Wang, Jiaqi},
  journal={arXiv preprint arXiv:2503.05236},
  year={2025}
}

@article{wu2023qalign,
  title={Q-align: Teaching lmms for visual scoring via discrete text-defined levels},
  author={Wu, Haoning and Zhang, Zicheng and Zhang, Weixia and Chen, Chaofeng and Liao, Liang and Li, Chunyi and Gao, Yixuan and Wang, Annan and Zhang, Erli and Sun, Wenxiu and others},
  journal={arXiv preprint arXiv:2312.17090},
  year={2023}
}

@article{liu2025aigve,
  title={AIGVE-MACS: Unified Multi-Aspect Commenting and Scoring Model for AI-Generated Video Evaluation},
  author={Liu, Xiao and Zhang, Jiawei},
  journal={arXiv preprint arXiv:2507.01255},
  year={2025}
}

@article{bansal2025videophy,
  title={Videophy-2: A challenging action-centric physical commonsense evaluation in video generation},
  author={Bansal, Hritik and Peng, Clark and Bitton, Yonatan and Goldenberg, Roman and Grover, Aditya and Chang, Kai-Wei},
  journal={arXiv preprint arXiv:2503.06800},
  year={2025}
}

@inproceedings{wu2023dover,
  title={Exploring video quality assessment on user generated contents from aesthetic and technical perspectives},
  author={Wu, Haoning and Zhang, Erli and Liao, Liang and Chen, Chaofeng and Hou, Jingwen and Wang, Annan and Sun, Wenxiu and Yan, Qiong and Lin, Weisi},
  booktitle={Proceedings of the IEEE/CVF international conference on computer vision},
  pages={20144--20154},
  year={2023}
}

@article{wu2025qsave,
  title={Q-Save: Towards Scoring and Attribution for Generated Video Evaluation},
  author={Wu, Xiele and Zhang, Zicheng and Chen, Mingtao and Liu, Yixian and Liu, Yiming and Wang, Shushi and Hu, Zhichao and Liu, Yuhong and Zhai, Guangtao and Liu, Xiaohong},
  journal={arXiv preprint arXiv:2511.18825},
  year={2025}
}

@article{wan2025wan,
  title={Wan: Open and advanced large-scale video generative models},
  author={Wan, Team and Wang, Ang and Ai, Baole and Wen, Bin and Mao, Chaojie and Xie, Chen-Wei and Chen, Di and Yu, Feiwu and Zhao, Haiming and Yang, Jianxiao and others},
  journal={arXiv preprint arXiv:2503.20314},
  year={2025}
}

@article{seedance2026seedance,
  title={Seedance 2.0: Advancing Video Generation for World Complexity},
  author={Seedance, Team and Chen, De and Chen, Liyang and Chen, Xin and Chen, Ying and Chen, Zhuo and Chen, Zhuowei and Cheng, Feng and Cheng, Tianheng and Cheng, Yufeng and others},
  journal={arXiv preprint arXiv:2604.14148},
  year={2026}
}

@inproceedings{chi2025empowering,
  title={Empowering world models with reflection for embodied video prediction},
  author={Chi, Xiaowei and Fan, Chun-Kai and Zhang, Hengyuan and Qi, Xingqun and Zhang, Rongyu and Chen, Anthony and Chan, Chi-Min and Xue, Wei and Liu, Qifeng and Zhang, Shanghang and others},
  booktitle={Forty-second International Conference on Machine Learning},
  year={2025}
}

@inproceedings{geng2025motion,
  title={Motion prompting: Controlling video generation with motion trajectories},
  author={Geng, Daniel and Herrmann, Charles and Hur, Junhwa and Cole, Forrester and Zhang, Serena and Pfaff, Tobias and Lopez-Guevara, Tatiana and Aytar, Yusuf and Rubinstein, Michael and Sun, Chen and others},
  booktitle={Proceedings of the Computer Vision and Pattern Recognition Conference},
  pages={1--12},
  year={2025}
}

@inproceedings{akkerman2025interdyn,
  title={Interdyn: Controllable interactive dynamics with video diffusion models},
  author={Akkerman, Rick and Feng, Haiwen and Black, Michael J and Tzionas, Dimitrios and Abrevaya, Victoria Fern{\'a}ndez},
  booktitle={Proceedings of the IEEE/CVF Conference on Computer Vision and Pattern Recognition},
  pages={12467--12479},
  year={2025}
}

@inproceedings{li2025wonderplay,
  title={Wonderplay: Dynamic 3d scene generation from a single image and actions},
  author={Li, Zizhang and Yu, Hong-Xing and Liu, Wei and Yang, Yin and Herrmann, Charles and Wetzstein, Gordon and Wu, Jiajun},
  booktitle={Proceedings of the IEEE/CVF International Conference on Computer Vision},
  pages={9080--9090},
  year={2025}
}

@article{zhan2026perpetualwonder,
  title={PerpetualWonder: Long-Horizon Action-Conditioned 4D Scene Generation},
  author={Zhan, Jiahao and Li, Zizhang and Yu, Hong-Xing and Wu, Jiajun},
  journal={arXiv preprint arXiv:2602.04876},
  year={2026}
}

@article{liu2026realwonder,
  title={RealWonder: Real-Time Physical Action-Conditioned Video Generation},
  author={Liu, Wei and Chen, Ziyu and Li, Zizhang and Wang, Yue and Yu, Hong-Xing and Wu, Jiajun},
  journal={arXiv preprint arXiv:2603.05449},
  year={2026}
}

@article{chen2025learning,
  title={Learning world models for interactive video generation},
  author={Chen, Taiye and Hu, Xun and Ding, Zihan and Jin, Chi},
  journal={arXiv preprint arXiv:2505.21996},
  year={2025}
}

@article{kong20253d,
  title={3D and 4D world modeling: A survey},
  author={Kong, Lingdong and Yang, Wesley and Mei, Jianbiao and Liu, Youquan and Liang, Ao and Zhu, Dekai and Lu, Dongyue and Yin, Wei and Hu, Xiaotao and Jia, Mingkai and others},
  journal={arXiv preprint arXiv:2509.07996},
  year={2025}
}

@article{liang2025worldlens,
  title={WorldLens: Full-spectrum evaluations of driving world models in real world},
  author={Liang, Ao and Kong, Lingdong and Yan, Tianyi and Liu, Hongsi and Yang, Wesley and Huang, Ziqi and Yin, Wei and Zuo, Jialong and Hu, Yixuan and Zhu, Dekai and others},
  journal={arXiv preprint arXiv:2512.10958},
  year={2025}
}

@article{chu2026agentic,
  title={Agentic World Modeling: Foundations, Capabilities, Laws, and Beyond},
  author={Chu, Meng and Zhang, Xuan Billy and Lin, Kevin Qinghong and Kong, Lingdong and Zhang, Jize and Tu, Teng and Ma, Weijian and Huang, Ziqi and Yang, Senqiao and Huang, Wei and others},
  journal={arXiv preprint arXiv:2604.22748},
  year={2026}
}

@article{wu2026visual,
  title={Visual Generation in the New Era: An Evolution from Atomic Mapping to Agentic World Modeling},
  author={Wu, Keming and Yang, Zuhao and Zhang, Kaichen and Wang, Shizun and Zhu, Haowei and Leng, Sicong and Yang, Zhongyu and Wang, Qijie and Wang, Sudong and Wang, Ziting and others},
  journal={arXiv preprint arXiv:2604.28185},
  year={2026}
}

@article{he2026gems,
  title={GEMS: Agent-Native Multimodal Generation with Memory and Skills},
  author={He, Zefeng and Huang, Siyuan and Qu, Xiaoye and Li, Yafu and Zhu, Tong and Cheng, Yu and Yang, Yang},
  journal={arXiv preprint arXiv:2603.28088},
  year={2026}
}

@article{feng2026gen,
  title={Gen-Searcher: Reinforcing Agentic Search for Image Generation},
  author={Feng, Kaituo and Zhang, Manyuan and Chen, Shuang and Lin, Yunlong and Fan, Kaixuan and Jiang, Yilei and Li, Hongyu and Zheng, Dian and Wang, Chenyang and Yue, Xiangyu},
  journal={arXiv preprint arXiv:2603.28767},
  year={2026}
}

@article{lin2025jarvisart,
  title={Jarvisart: Liberating human artistic creativity via an intelligent photo retouching agent},
  author={Lin, Yunlong and Lin, Zixu and Lin, Kunjie and Bai, Jinbin and Pan, Panwang and Li, Chenxin and Chen, Haoyu and Wang, Zhongdao and Ding, Xinghao and Li, Wenbo and others},
  journal={arXiv preprint arXiv:2506.17612},
  year={2025}
}

@inproceedings{zhao2025cot,
  title={Cot-vla: Visual chain-of-thought reasoning for vision-language-action models},
  author={Zhao, Qingqing and Lu, Yao and Kim, Moo Jin and Fu, Zipeng and Zhang, Zhuoyang and Wu, Yecheng and Li, Zhaoshuo and Ma, Qianli and Han, Song and Finn, Chelsea and others},
  booktitle={Proceedings of the Computer Vision and Pattern Recognition Conference},
  pages={1702--1713},
  year={2025}
}

@article{du2023learning,
  title={Learning universal policies via text-guided video generation},
  author={Du, Yilun and Yang, Sherry and Dai, Bo and Dai, Hanjun and Nachum, Ofir and Tenenbaum, Josh and Schuurmans, Dale and Abbeel, Pieter},
  journal={Advances in neural information processing systems},
  volume={36},
  pages={9156--9172},
  year={2023}
}

@article{gao2025seedance,
  title={Seedance 1.0: Exploring the boundaries of video generation models},
  author={Gao, Yu and Guo, Haoyuan and Hoang, Tuyen and Huang, Weilin and Jiang, Lu and Kong, Fangyuan and Li, Huixia and Li, Jiashi and Li, Liang and Li, Xiaojie and others},
  journal={arXiv preprint arXiv:2506.09113},
  year={2025}
}

@article{long20262,
  title={{A\textsuperscript{2}RD}: Agentic Autoregressive Diffusion for Long Video Consistency},
  author={Long, Do Xuan and Song, Yale and Kan, Min-Yen and Pfister, Tomas and Le, Long T},
  journal={arXiv preprint arXiv:2605.06924},
  year={2026}
}
}

\end{document}